\documentclass[pdflatex,sn-mathphys-num]{sn-jnl}

\usepackage{graphicx}%
\usepackage{multirow}%
\usepackage{amsmath,amssymb,amsfonts}%
\usepackage{amsthm}%
\usepackage{mathrsfs}%
\usepackage[title]{appendix}%
\usepackage{xcolor}%
\usepackage{textcomp}%
\usepackage{manyfoot}%
\usepackage{booktabs}%
\usepackage{algorithm}%
\usepackage{algorithmicx}%
\usepackage{algpseudocode}%
\usepackage{listings}%
\usepackage[table]{xcolor}
\usepackage{booktabs}
\usepackage{multirow}
\usepackage{adjustbox}
\usepackage{geometry}
\usepackage{subcaption} 
\usepackage{placeins}
\usepackage{grffile} 
\usepackage{hyperref} 
\usepackage{cleveref} 

\theoremstyle{thmstyleone}%
\theoremstyle{thmstyletwo}%

\theoremstyle{thmstylethree}%

\begin{document}

\title[Counting the Cost of War Under Satellite Embargo]{Counting the Cost of War Under Satellite Embargo: Zero-Shot Estimation of Impacted Infrastructure}


\author*[1]{\fnm{Saleh Sakib} \sur{Ahmed}}\email{salehsakibahmed@gmail.com}


\author*[1]{\fnm{M. Sohel} \sur{Rahman}}\email{sohel.kcl@gmail.com}

\affil[1]{\orgdiv{Department of Computer Science and Engineering}, \orgname{Bangladesh University of Engineering and Technology}, \orgaddress{\street{Palashi}, \city{Dhaka}, \postcode{1000}, \country{Bangladesh}}}


\abstract{
Rapid estimation of impacted structures—critical for conflict-zone humanitarian response—is frequently hindered by post-strike satellite data embargoes and imagery blackouts. We bypass this operational bottleneck by reframing impacted building mapping as a zero-shot geometric projection task on archival, pre-strike maps. Using coordinate and incident text from LiveUAMap and ArcGIS, Large Language Models extract weapon payloads ($W$) to project kinetic blast perimeters via Hopkinson--Cranz scaling ($R_{\text{base}} = Z W^{1/3}$). To count exposed structures within these zones without post-strike imagery, we introduce two technical innovations: \textit{Adaptive Field-of-View} to eliminate resolution (zoom) bias in 2D segmentation (SAMGeo), and \textit{2.5D pseudo-height depth maps} combined with segmentation masks to help Large Vision-Language Models (LVLMs) resolve overlapping, dense rooftops. Evaluated on 2026 Middle East conflict data, depth-augmented LVLMs dramatically outperform traditional segmentation in congested urban centers. This establishes a powerful hybrid paradigm for zero-shot crisis mapping: ultra-fast 2D segmentation for sparse rural zones, and depth-augmented LVLMs for dense urban environments.}


\maketitle
\section{Introduction}

During urban crises and geopolitical conflicts, rapidly quantifying structural exposure is critical across multiple domains. In the vital hours following a kinetic strike, turning fragmented text reports into actionable impact maps is essential for emergency response, strategic planning, and civil accountability. For Humanitarian Aid and Disaster Relief (HADR) dispatchers, rapidly identifying impacted structures from initial text alerts informs whether to deploy heavy search-and-rescue machinery to a 50-building impact zone or dispatch localized ambulances. For defense and Geospatial Intelligence (GEOINT) analysts, rapid pre-strike Collateral Damage Estimation (CDE) and immediate exposure modeling are required without re-tasking constrained satellite orbits or risking aerial reconnaissance over contested airspace. Simultaneously, Open-Source Intelligence (OSINT) collectives and legal observers rely on independent structural estimates to pierce state information blockades and audit civilian impacts.

Although modern satellite constellations can technically supply high-resolution tracking, real-world access to post-strike imagery is systematically delayed. For civilian and OSINT sectors, critical information blackouts are routinely created by commercial data embargoes; during the early stages of the 2026 Middle-East crisis, for example, major commercial providers imposed an initial 96-hour image blackout that was subsequently extended to 14 days~\cite{aljazeera2026planet}. Even for advanced defense entities, continuous post-event observation is hindered by orbital revisit windows and cloud cover~\cite{esa2020earthobservation,esa2023eoguide,usgs2023clouds}, while active counter-space capabilities---including directed energy and electronic jamming---further restrict post-strike data acquisition~\cite{spaceforce2024directedenergy,spaceforce2024spacethreats}. Mitigating these blackouts traditionally requires manually cross-referencing archival maps with empirical blast charts---a laborious process fundamentally incompatible with time-sensitive emergency response. This highlights an urgent need for automated vision systems capable of mapping impacted structures immediately on pre-strike imagery, bypassing post-event satellite delays entirely.

To address this challenge, this paper presents an automated, zero-shot framework for estimating impacted structures by counting exposed buildings within a predicted kinetic blast perimeter using only pre-strike static satellite maps and text reports. In this framework, infrastructure impact estimation is formulated as a predictive geometric task that estimates built-environment vulnerability prior to the availability of post-strike imagery. Evaluated on empirical data from the 2026 Middle-East conflict, we demonstrate the pipeline's robustness across dense architecture, low visual contrast, and severe commercial embargoes. To evaluate performance under varying degrees of spatial uncertainty, the framework ingests geographic coordinates and incident text from two distinct platform types: rapid but spatially volatile crowdsourced feeds (LiveUAMap~\cite{liveuamap_iran}) and precise spatial registries (ArcGIS StoryMaps~\cite{mealie2026interactive}). Comparing these data streams allows us to benchmark zero-shot impact mapping performance across both early, unverified field alerts and high-fidelity geospatial intelligence.

Our pipeline processes crisis metadata through a structured, sequential workflow. First, reported strike coordinates retrieve pre-strike static satellite imagery from standard mapping services (\textit{e.g.}, ESRI, Google). Simultaneously, a Large Language Model (LLM) analyzes unstructured text alerts to extract the specific munition payload mass ($W$). Using this payload value, the system applies the Hopkinson--Cranz scaling law ($R_{\text{base}} = Z W^{1/3}$) to project a baseline physical hazard boundary. This formulation enables rapid, zero-shot estimation without requiring computationally prohibitive 3D blast simulations that model urban canyoning and wave reflection. The physical justifications for selecting representative payload mass ($W$) and scaling coefficient ($Z$) are detailed in Sec.~\ref{sec:radius_justification} of the Supplementary Material. Finally, the pipeline projects this reference perimeter directly onto pre-strike imagery, framing the identification of impacted structures as a clean spatial counting task: quantifying standing buildings subjected to potential blast effects within the perimeter. This pre-strike inventory serves as our quantitative baseline for structural exposure, establishing built-environment vulnerability prior to post-strike visual confirmation.

To automate building counting within the projected perimeter, we evaluate two computational paradigms: programmatic 2D segmentation and Large Vision-Language Model (LVLM) inference. For our programmatic baseline, we evaluate SAMGeo~\cite{wu2023samgeo}, built on Meta's Segment Anything Model (SAM) framework~\cite{kirillov2023segment,ravi2025sam,carion2025sam}. However, experimentation reveals that SAMGeo is highly sensitive to input resolution and field-of-view: zooming out causes adjacent rooftops to merge into continuous pixel masses, whereas zooming in truncates contextual boundaries. To standardize segmentation across diverse map sources, we introduce an \textit{Adaptive Field-of-View} routine that dynamically calibrates zoom levels and spatial padding.

For the LVLM pathway, direct spatial object counting remains a challenging task due to weak visual grounding~\cite{xu2025llm,wang2024language}. While recent approaches employ 2D visual prompts---such as Set-of-Mark (SoM) overlays~\cite{yang2023set} or pre-segmented map layers~\cite{qharabagh2024lvlm,zou2023segment}---flat 2D representations degrade in dense urban cores where touching rooftops lack distinct horizontal contrast. To resolve this, we leverage the spatial regularizing properties of monocular depth estimation~\cite{ranftl2020towards,yang2024depth,lin2025depth,bochkovskiy2025depth}. Although developed primarily for ground-level scenes, monocular depth models can be effectively applied to top-down satellite rasters to extract relative elevation gradients without stereoscopic input~\cite{gultekin2025fusing,li2025multistage}. These 2.5D pseudo-height maps provide vertical visual cues that allow vision-language models to visually decouple adjoining structures. We systematically benchmark four visual grounding configurations---raw imagery, location labels, 2D segmentation masks, and 2.5D relative depth overlays---across dense and Mixture-of-Experts (MoE) LVLM architectures.

Our results reveal a clear performance trade-off governed by structural density. Programmatic 2D segmentation (SAMGeo) achieves near-perfect accuracy with minimal computational overhead in sparse environments ($0$--$5$ structures). However, in dense urban settings ($16+$ structures), 2D segmentation fails to resolve adjacent rooftops, leading to severe undercounting. In contrast, depth-augmented LVLMs successfully isolate clustered roofs using relative elevation cues, maintaining high accuracy in dense urban cores across both crowdsourced and curated text feeds. Because LVLMs require higher computational resources and exhibit non-deterministic variance, these findings establish a practical density-routed paradigm for zero-shot crisis mapping: deploy lightweight 2D segmentation for sparse rural terrain, and dynamically route dense urban zones to depth-augmented LVLMs.

By eliminating the wait for delayed post-strike imagery, this framework delivers the immediate spatial data required by humanitarian, defense, and legal teams. Beyond its operational utility, it offers significant value to AI research by reframing standard object counting into a constrained spatial reasoning task. Models must demonstrate precise spatial awareness---counting structures strictly within the predicted blast zone while ignoring complex city blocks directly outside the boundary. This introduces a challenging, real-world benchmark to evaluate how effectively vision models follow spatial rules and utilize depth cues. To support future research, our georeferenced dataset, codebase, model prompts, and map layers are publicly available.

\begin{figure*}[!htbp]
    \centering
    \includegraphics[width=\linewidth]{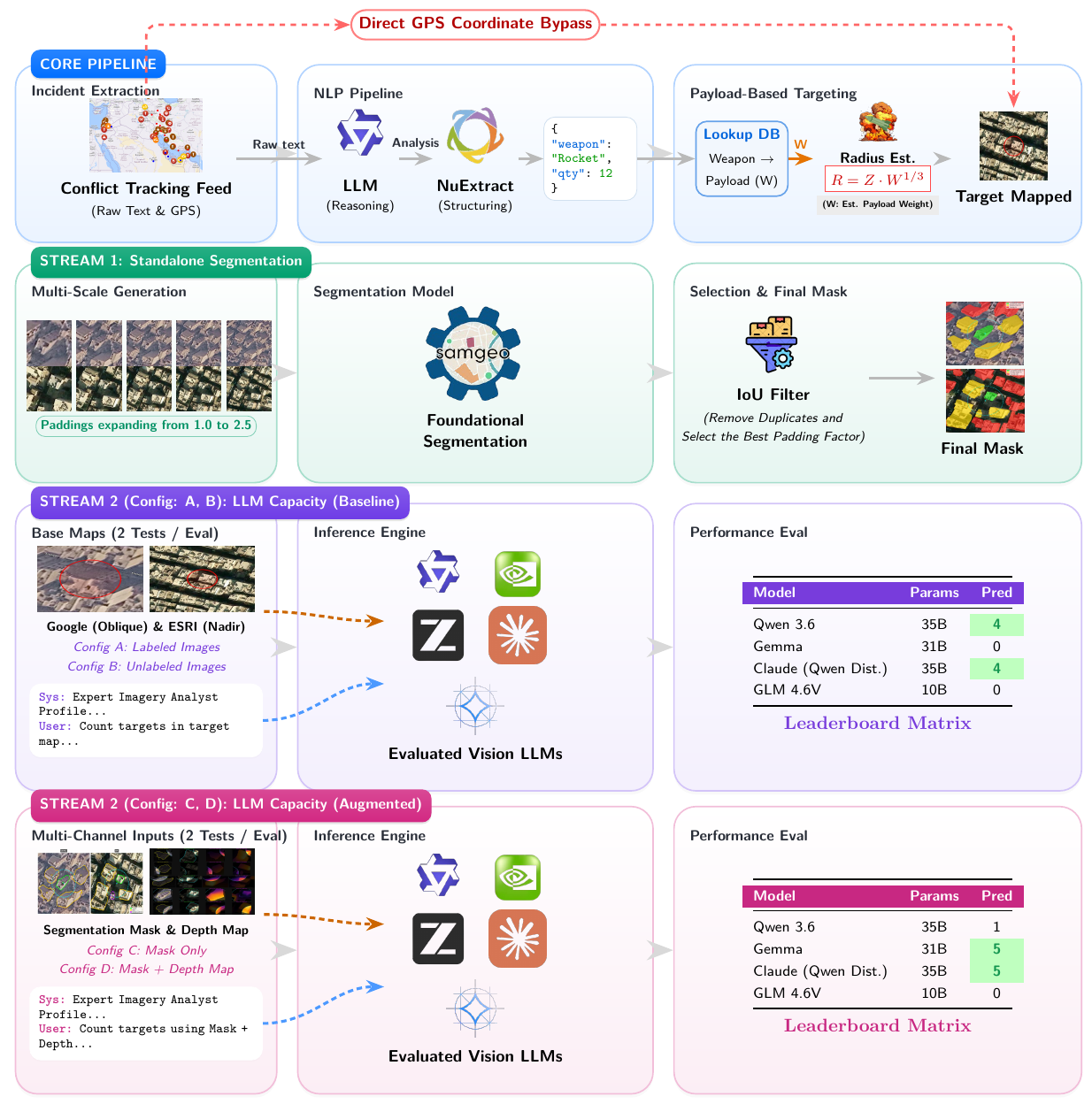}
    \caption{\textbf{Unified Framework: From Raw Text Alerts to Blast Mapping and Building Counting.} The end-to-end framework bypasses satellite blackouts by converting unstructured open-source text alerts into a deterministic building-counting task. 
    \textbf{Core Pipeline:} The framework ingests conflict data to extract spatial location and weapon profiles; pre-validated data bypasses this stage via the \textit{Direct GPS Coordinate Bypass}. Extracted coordinates automatically trigger the retrieval of archival, pre-strike satellite imagery. Simultaneously, an LLM cascade processes text descriptions to isolate the weapon type and map it to a representative payload mass ($W$). The system then applies the Hopkinson-Cranz scaling law ($R_{\text{base}} = ZW^{1/3}$) to calculate a physics-enforced blast radius ($R_{\text{base}}$). Projecting this boundary onto pre-strike imagery frames the exposure assessment as estimating the total standing structures within the blast perimeter. 
    \textbf{Evaluation Streams:} Two operational paradigms are evaluated to execute this counting task: 
    (1) \textbf{Stream 1 (Programmatic Segmentation):} Structures are segmented using an \textit{Adaptive Field-of-View} routine to calibrate sensor fields-of-view, a foundational segmentation engine (SAMGeo), and an \textit{IoU Filter} to eliminate duplicate shapes. 
    (2) \textbf{Stream 2 (Vision-Language Evaluation):} Open-weight multimodal LLMs are benchmarked across four prompt configurations: \textbf{Config A} (location labeled satellite imagery), \textbf{Config B} (unlabeled raw satellite imagery baseline), \textbf{Config C} (augmented with 2D segmentation masks only), and \textbf{Config D} (augmented with both segmentation masks and relative depth maps). Performance metrics are compiled in the \textit{Leaderboard Matrix} to measure absolute prediction accuracy.}
    \label{fig:full_pipeline}
\end{figure*}

\begin{figure*}[!htbp]
    \centering
    \includegraphics[width=\linewidth]{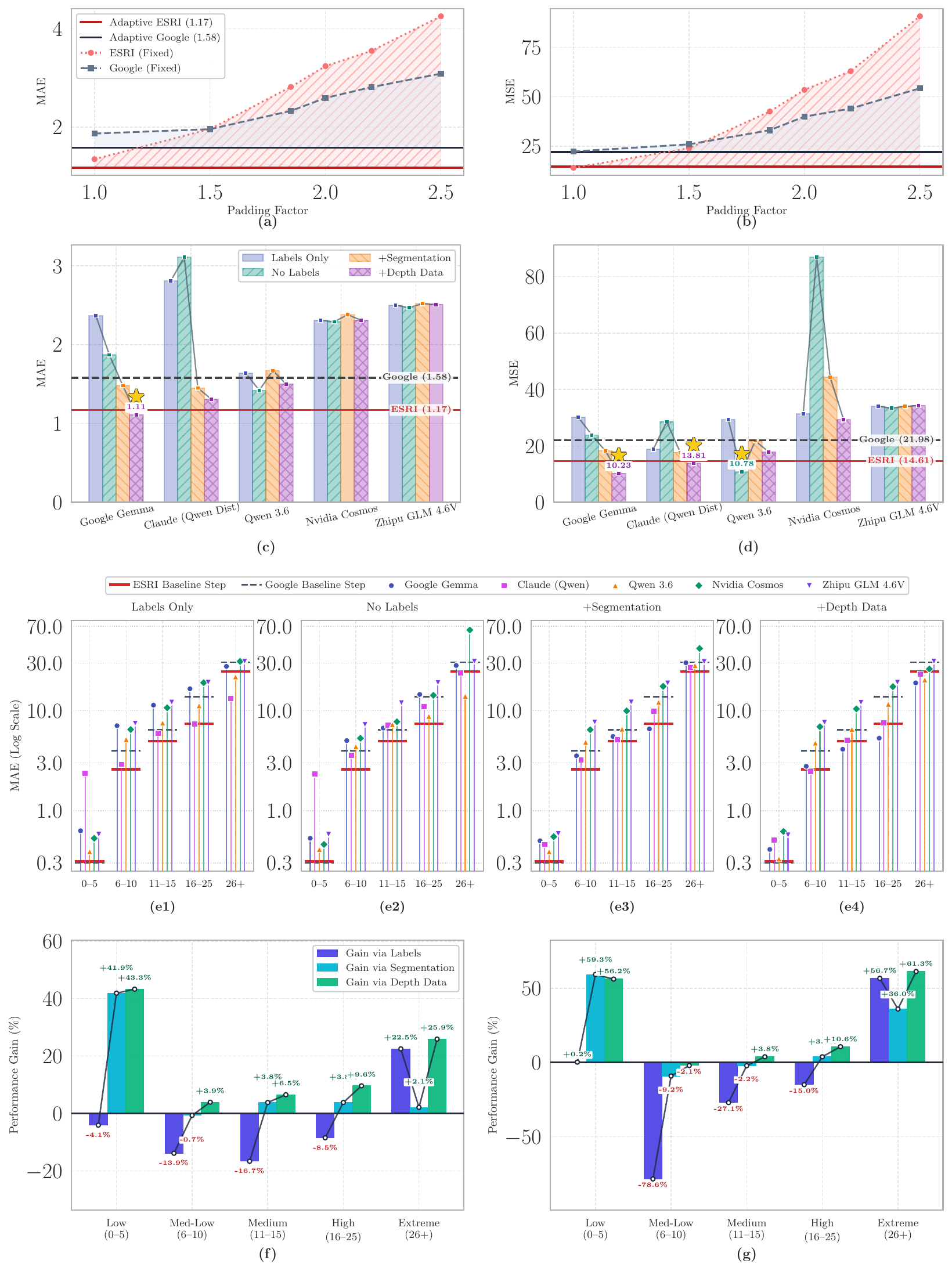}
    \caption{\textbf{Performance Evaluation and Ablation Analysis of the LVLM Architecture on LiveUAMap.} 
    \textbf{(a)~\&~(b)} Error metrics (MAE and MSE) across spatial padding factors, contrasting the adaptive strategy against fixed baselines. 
    \textbf{(c)~\&~(d)} Benchmarking of five vision-language models across four visual configurations against Google and ESRI references, with stars highlighting optimal configurations. 
    \textbf{(e1--e4)} Localized MAE breakdown across five distinct urban density tiers. 
    \textbf{(f)~\&~(g)} Relative performance shift (\%) demonstrating the sequential marginal impact of text labels, segmentation masks, and relative depth cues.}
    \label{fig:vlm_spatial_density_dashboard}
\end{figure*}

\section{Results}
\label{sec:results}

\subsection*{System Overview and Core Workflow}
\label{sec:system_overview}

As shown in Fig.~\ref{fig:full_pipeline}, our framework estimates building damage directly from unstructured web alerts. The approach rests on a simple premise: once strike coordinates and the impact radius are known, damage assessment reduces to a geometric building-counting task on pre-strike satellite imagery. This strategy avoids real-time data blackouts by relying entirely on archival imagery and open-source intelligence. The processing workflow runs across three main stages.

\paragraph{Core Pipeline: Damage radius estimation and image retrieval.} First, the system collects text alerts from LiveUAMap and ArcGIS StoryMaps. It cleans this data by keeping only kinetic attack reports from LiveUAMap and exact geographic coordinates from ArcGIS. A two-step language model then reads the text: the first model estimates the weapon type ($w_{\text{class}}$) and attack severity ($\Delta_{\text{modifier}}$), while a second model (NuExtract~\cite{numind_nuextract_2_0_8b}) organizes these details into a clean JSON format.

To calculate the blast zone, the system converts the weapon type ($w_{\text{class}}$) into an estimated explosive weight ($W$). It then uses standard blast scaling laws ($R_{\text{base}} = ZW^{1/3}$) to figure out a starting radius ($R_{\text{base}}$). Adjusting this radius with the severity score ($\Delta_{\text{modifier}}$) gives the final damage radius ($R_{\text{moderate}}$), which accounts for real-world factors like secondary explosions and defines the exact area to check for building damage.

Finally, the system downloads satellite images taken before the strike from ESRI and Google Maps. It expands the image view area using a padding factor ($\alpha$, where $R_{\text{padded}} = \alpha \cdot R_{\text{moderate}}$) so the models can see the surrounding environment for better visual context. This parameter regulates the spatial extent and field-of-view (FOV) presented to the downstream models. Crucially, while the model observes the broader contextual frame, structural exposure is strictly evaluated by counting only those buildings residing within the unpadded target radius ($R_{\text{moderate}}$).

\paragraph{Programmatic building segmentation (Stream 1).} Stream 1 uses SAMGeo to outline building footprints in the damage zone. Because building detection accuracy depends heavily on the visible surrounding area and spatial contect, our \textit{Adaptive Field-of-View} routine tests six different padding factors ($\alpha_j \in \Omega = \{1.0, \dots, 2.5\}$) to find the optimal spatial extent. 

At each zoom level, SAMGeo is queried using an ensemble of synonymous text prompts (e.g., ``houses,'' ``small buildings'') to cast a wide net for structures. The system then cleans these raw detections in two steps: first, it removes nested parts (like a balcony outlined inside a house) by discarding smaller shapes that fall mostly inside larger ones; second, it merges redundant, overlapping boundaries to ensure each building is only counted once. This leaves a refined set of unique building footprints ($\mathcal{M}_{\text{final}}$).

To determine the optimal padding factor ($\alpha^*$) without ground-truth labels, the system evaluates candidate spatial extents by scoring segmented structures relative to the target blast circle ($R_{\text{moderate}}$) across both ESRI and Google basemaps. Each detected building is assigned a spatial inclusion score: structures completely inside $R_{\text{moderate}}$ receive a weight of $1.0$, partially overlapping structures receive $0.5$, and external structures receive $0.0$. The system selects the padding factor ($\alpha^*$) that maximizes this cumulative inclusion score, favoring the wider spatial extent in the event of a tie. Using this optimal view, Stream 1 calculates the final structural exposure count as the sum of all fully and partially intersecting building segments.

\paragraph{Vision-language reasoning (Stream 2)}
Stream 2 evaluates zero-shot spatial counting across open-weight Large Vision-Language Models (LVLMs) within damage footprints using four visual input configurations:

\textbf{Standard inputs.} Config A provides location-labeled imagery with native map annotations, whereas Config B uses clean, unlabeled satellite imagery to measure baseline performance. Location labels were added to Config A to test whether explicit visual annotations improve spatial grounding by requiring models to list locations within the damage radius (results detailed in Supplementary Section~\ref{sec:text_extraction}).

\textbf{Spatial augmentations.} Designed to systematically resolve structural ambiguity in congested urban environments, two spatial augmentation strategies are evaluated: Config C overlays 2D programmatic segmentation masks generated by Stream 1 (e.g., SAMGeo), supplying explicit geometric contours to delineate individual building footprints. Config D incorporates both segmentation maps and monocular depth maps to provide 2.5D pseudo-height cues and relative elevation gradients. While 2D segmentation frequently fails when adjacent rooftops share identical spectral signatures or colors, Config D introduces vertical spatial relief. This enables vision-language models to distinguish touching structures by height variance, directly mitigating boundary collapse in extreme density tiers.
    
To eliminate sampling variance, each location is queried five times ($N=5$). A parsing engine running NuExtract converts raw text outputs into structured numbers, taking the median count as the canonical prediction.

\begin{table}[!htbp]
  \centering
  \caption{Comparison of Total building segmentation performance between adaptive and fixed padding strategies across the Liveuamap and ArcGIS datasets. The best result per column is highlighted in \textbf{bold}, and the second best is \underline{underlined}. Lower is better.}
  \label{tab:combined_total_main}
  
  \begin{tabular}{l cccc cccc}
    \toprule
    \multirow{3}{*}{Padding Strategy} & \multicolumn{4}{c}{Liveuamap Dataset} & \multicolumn{4}{c}{ArcGIS StoryMap Dataset} \\
    \cmidrule(lr){2-5} \cmidrule(lr){6-9}
    & \multicolumn{2}{c}{Google Imagery} & \multicolumn{2}{c}{ESRI Imagery} & \multicolumn{2}{c}{Google Imagery} & \multicolumn{2}{c}{ESRI Imagery} \\
    \cmidrule(lr){2-3} \cmidrule(lr){4-5} \cmidrule(lr){6-7} \cmidrule(lr){8-9}
    & MAE & MSE & MAE & MSE & MAE & MSE & MAE & MSE \\
    \midrule
    Fixed 1.0  & \underline{1.87} & \underline{22.33} & \underline{1.35} & \textbf{14.06} & \underline{3.01} & \underline{25.91} & \underline{2.79} & \underline{22.23} \\
    Fixed 1.5  & 1.96 & 25.94 & 1.96 & 23.96 & 3.02 & 26.79 & 3.14 & 25.55 \\
    Fixed 1.85 & 2.33 & 33.08 & 2.81 & 42.52 & 3.81 & 38.98 & 4.11 & 45.32 \\
    Fixed 2.0  & 2.59 & 39.93 & 3.24 & 53.47 & 4.01 & 44.61 & 4.54 & 55.29 \\
    Fixed 2.2  & 2.81 & 44.00 & 3.55 & 62.87 & 4.70 & 56.04 & 5.24 & 72.65 \\
    Fixed 2.5  & 3.09 & 54.25 & 4.25 & 90.63 & 5.56 & 74.16 & 6.37 & 105.02 \\
    \midrule
    Adaptive   & \textbf{1.58} & \textbf{21.98} & \textbf{1.17} & \underline{14.61} & \textbf{2.07} & \textbf{19.26} & \textbf{2.42} & \textbf{19.29} \\
    \bottomrule
  \end{tabular}
  
\end{table}

\subsection{Results of Stream 1: Building Segmentation}
\label{subsec:stream1_results}

To evaluate segmentation performance, we benchmarked the dynamic Adaptive Field-of-View against rigid padding baselines ($1.0\times$ to $2.5\times$). As shown in Table~\ref{tab:combined_total_main} (supported by Figure~\ref{fig:vlm_spatial_density_dashboard}a--b and Supplementary Figure~\ref{fig:arcgis_results_llm}a--b), the tightest crop ($1.0\times$) provides the strongest fixed baseline. Expanding the spatial extent degrades performance monotonically across all datasets; for example, increasing fixed padding to $2.5\times$ on LiveUAMap ESRI imagery spikes Mean Squared Error (MSE) by over $540\%$ (from $14.06$ to $90.63$) and more than triples Mean Absolute Error (MAE) from $1.35$ to $4.25$.

The Adaptive FOV consistently outperforms fixed cropping by dynamically picking the best zoom level per scene. On the high-precision ArcGIS dataset, the adaptive module slashes Google Maps MAE by $31.2\%$ (from $3.01$ to $2.07$) and MSE by $25.7\%$ (from $25.91$ to $19.26$), while reducing ESRI MAE by $13.3\%$ (from $2.79$ to $2.42$). Similarly, on LiveUAMap Google imagery, it reduces total MAE by $15.5\%$ (from $1.87$ to $1.58$) and achieves the lowest overall dataset error on LiveUAMap ESRI with an MAE of $1.17$.

A granular category breakdown (Supplementary Tables~\ref{tab:supp_complete} and \ref{tab:supp_partial}) illustrates the exact mechanics driving the performance gains of the adaptive approach:

\noindent\textbf{Structures fully enclosed within the perimeter.} For completely contained buildings, a rigid $1.0\times$ crop performs competitively because targets remain centrally positioned within the evaluation frame, maintaining a minor advantage on the LiveUAMap dataset (e.g., Google imagery MAE $0.77$ vs. $0.81$ for adaptive; ESRI imagery MAE $0.60$ vs. $0.64$; Table~\ref{tab:supp_complete}). However, on the more structurally complex ArcGIS dataset, the adaptive field-of-view (FOV) strategy achieves the best overall performance, reducing complete-structure MAE by $6.8\%$ on Google imagery ($1.18 \to 1.10$; MSE $7.44 \to 7.01$) and outperforming all fixed crops on ESRI imagery ($1.13 \to 1.12$; MSE $6.50 \to 6.38$).

\noindent\textbf{Structures partially bisecting the perimeter.} For structures truncated by the evaluation boundary, rigid $1.0\times$ crops sever critical visual features along image edges due to insufficient peripheral context. The adaptive FOV mechanism directly resolves these edge-clipping artifacts, achieving the best performance across all metrics, datasets, and imagery providers (Table~\ref{tab:supp_partial}). On LiveUAMap, adaptive FOV slashes partial-structure MAE by $25.4\%$ on Google imagery ($1.18 \to 0.88$; MSE $6.97 \to 5.39$) and $23.1\%$ on ESRI imagery ($0.91 \to 0.70$; MSE $4.42 \to 3.55$). This contextual advantage is even more pronounced on the ArcGIS dataset, where adaptive padding reduces partial MAE by $37.3\%$ on Google imagery ($2.01 \to 1.26$; MSE $8.93 \to 5.07$) and $19.9\%$ on ESRI imagery ($1.91 \to 1.53$; MSE $8.01 \to 5.99$).

\begin{table*}[t]
\centering
\small
\caption{\textbf{Multimodal Ablation Performance across LiveUAMap and ArcGIS Datasets.} Comprehensive evaluation of MAE and MSE metrics across nested visual feature configurations (explicit text \textit{Labels}, unassisted \textit{No Lbls}, 2D segmentation overlays \textit{+Seg}, and composite depth maps \textit{+Depth}). Note that for notation brevity in headers, the \textit{+Depth} designation denotes a composite input combining relative depth maps alongside 2D segmentation overlays rather than standalone depth. Panel~A details results on the crowdsourced LiveUAMap dataset; Panel~B details results on the high-precision ArcGIS dataset. Bold formatting indicates the best result and underlining indicates second-best within each model's configuration row or cohort. Asterisks ($^*$) denote LVLM configurations that outperform the highest-performing deterministic SAMGeo baseline on the corresponding dataset.}
\label{tab:combined_ablation_performance}

\begin{tabular}{llcccccccc}
\toprule
\multicolumn{10}{l}{\textbf{Panel A: LiveUAMap Dataset (Crowdsourced Coordinate Stream)}} \\
\midrule
\textbf{Model Name} & \textbf{Config} & \multicolumn{4}{c}{\textbf{Mean Absolute Error (MAE) $\downarrow$}} & \multicolumn{4}{c}{\textbf{Mean Squared Error (MSE) $\downarrow$}} \\
\cmidrule(lr){3-6} \cmidrule(lr){7-10}
 & & \textbf{Labels} & \textbf{No Lbls} & \textbf{+Seg} & \textbf{+Depth} & \textbf{Labels} & \textbf{No Lbls} & \textbf{+Seg} & \textbf{+Depth} \\
\midrule
Claude (Qwen Dist) & \shortstack{35B \\ \scriptsize \textit{MoE}} & $2.81$ & $3.11$ & $\underline{1.45}$ & $\mathbf{1.31}$ & $18.79$ & $28.51$ & $\underline{17.67}$ & $\mathbf{13.81}^{*}$ \\[0.5ex]
Qwen 3.6 Baseline  & \shortstack{35B \\ \scriptsize \textit{MoE}} & $1.64$ & $\mathbf{1.42}$ & $1.67$ & $\underline{1.50}$ & $29.30$ & $\mathbf{10.78}^{*}$ & $21.80$ & $\underline{17.88}$ \\[0.5ex]
Google Gemma 4     & \shortstack{31B \\ \scriptsize \textit{Dense}} & $2.37$ & $1.87$ & $\underline{1.48}$ & $\mathbf{1.11}^{*}$ & $30.15$ & $23.85$ & $\underline{18.23}$ & $\mathbf{10.23}^{*}$ \\[0.5ex]
Nvidia Cosmos      & \shortstack{32B \\ \scriptsize \textit{Dense}} & $\underline{2.31}$ & $\mathbf{2.29}$ & $2.38$ & $2.31$ & $\underline{31.41}$ & $86.91$ & $44.36$ & $\mathbf{29.29}$ \\[0.5ex]
Zhipu GLM 4.6V     & \shortstack{10B \\ \scriptsize \textit{Dense}} & $\underline{2.50}$ & $\mathbf{2.47}$ & $2.52$ & $2.51$ & $34.04$ & $\mathbf{33.42}$ & $\underline{34.00}$ & $34.31$ \\[0.5ex]
\midrule
SAMGeo Google Reference & 0.85B & \multicolumn{4}{c}{\small MAE: 1.58} & \multicolumn{4}{c}{\small MSE: 21.98} \\
SAMGeo ESRI Reference   & 0.85B & \multicolumn{4}{c}{\small MAE: 1.17} & \multicolumn{4}{c}{\small MSE: 14.61} \\
\bottomrule
\end{tabular}

\vspace{1.2em}

\begin{tabular}{llcccccc}
\toprule
\multicolumn{8}{l}{\textbf{Panel B: ArcGIS Dataset (High-Precision Coordinate Stream)}} \\
\midrule
\textbf{Model Name} & \textbf{Config} & \multicolumn{3}{c}{\textbf{Mean Absolute Error (MAE) $\downarrow$}} & \multicolumn{3}{c}{\textbf{Mean Squared Error (MSE) $\downarrow$}} \\
\cmidrule(lr){3-5} \cmidrule(lr){6-8}
 & & \textbf{No Lbls} & \textbf{+Seg} & \textbf{+Depth} & \textbf{No Lbls} & \textbf{+Seg} & \textbf{+Depth} \\
\midrule
Claude (Qwen Dist) & \shortstack{35B \\ \scriptsize \textit{MoE}} & $\underline{2.61}$ & $2.69$ & $\mathbf{2.23}$ & $\mathbf{18.26}^{*}$ & $22.91$ & $\underline{18.35}^{*}$ \\[0.5ex]
Qwen 3.6 Baseline  & \shortstack{35B \\ \scriptsize \textit{MoE}} & $\underline{2.90}$ & $3.03$ & $\mathbf{2.70}$ & $\underline{28.49}$ & $29.25$ & $\mathbf{23.92}$ \\[0.5ex]
Google Gemma 4     & \shortstack{31B \\ \scriptsize \textit{Dense}} & $4.19$ & $\underline{2.69}$ & $\mathbf{2.08}$ & $44.59$ & $\underline{21.39}$ & $\mathbf{14.28}^{*}$ \\[0.5ex]
\midrule
SAMGeo Google Reference & 0.85B & \multicolumn{3}{c}{\small MAE: 2.07} & \multicolumn{3}{c}{\small MSE: 19.26} \\
SAMGeo ESRI Reference   & 0.85B & \multicolumn{3}{c}{\small MAE: 2.42} & \multicolumn{3}{c}{\small MSE: 19.29} \\
\bottomrule
\end{tabular}

\end{table*}
\subsection{Results of Stream 2: LVLM Building Counting}
\label{subsec:results_stream2}

This subsection presents the empirical evaluation of Large Vision-Language Models (LVLMs) for zero-shot building counting within projected blast perimeters. Models are benchmarked across both the crowdsourced LiveUAMap and high-precision ArcGIS datasets under four multimodal configurations, isolating the impact of 2D segmentation overlays (\textit{+Seg}) and relative depth maps (\textit{+Depth}). To optimize spacing across tables and figures, shortened labels are used throughout; specifically, \textit{+Depth} denotes integrating relative depth gradients alongside 2D segmentation overlays, rather than a standalone depth layer.

\subsubsection{Global Model-Level Performance and Feature Ablation}
\label{subsubsec:global_ablation}

Adding composite relative depth maps (\textit{+Depth}) significantly reduces building counting errors across all evaluated datasets. Rather than requiring complex 3D geometry parsing, relative depth overlays provide straightforward structural height cues that allow Large Vision-Language Models (LVLMs) to resolve connected, flat roofs into distinct building boundaries. This effectively mitigates severe under-counting in high-density urban areas (Table~\ref{tab:combined_ablation_performance}, Figure~\ref{fig:vlm_spatial_density_dashboard}c--d).

In contrast, explicitly burning text-based location prompts (\textit{Labels}) into the image canvas consistently underperformed compared to unassisted baseline images (\textit{No Lbls}) and structural visual prompts. On LiveUAMap (Panel~A), adding text labels introduced visual occlusion and prompt clutter, degrading counting accuracy for top performers like \textbf{Google Gemma 4 (31B)} (MAE increased from $1.87$ to $2.37$) and \textbf{Qwen 3.6 (35B)} (MAE increased from $1.42$ to $1.64$). Because location labels proved counterproductive and failed to yield spatial benefits, the \textit{Labels} configuration was entirely omitted from the high-precision ArcGIS benchmark (Panel~B) to optimize computational throughput and focus on geometric visual features.

\textbf{Google Gemma 4 (31B Dense)} demonstrated the strongest overall performance when combined with depth overlays. On LiveUAMap, incorporating depth reduced its unassisted baseline Mean Absolute Error (MAE) by $40.6\%$ (from $1.87$ to $1.11$) and Mean Squared Error (MSE) by $57.1\%$ (from $23.85$ to $10.23$). This trend extended to ArcGIS, where depth reduced Gemma 4's unassisted MAE by $50.4\%$ (from $4.19$ to $2.08$) and slashed its MSE by $67.9\%$ (from $44.59$ to $14.28$).

Among Mixture-of-Experts (MoE) architectures, \textbf{Claude (Qwen Dist, 35B)} showed severe dependency on spatial visual aids. On LiveUAMap, adding 2D segmentation overlays (\textit{+Seg}) dropped its MAE by $53.4\%$ (from $3.11$ to $1.45$), while depth (\textit{+Depth}) reduced its MSE by $51.6\%$ (from $28.51$ to $13.81$). On ArcGIS, depth yielded its optimal performance at $2.23$ MAE and $18.35$ MSE. Conversely, \textbf{Qwen 3.6 Baseline (35B)} achieved strong performance on LiveUAMap using bare unassisted imagery ($1.42$ MAE, $10.78$ MSE). However, on the fine-grained ArcGIS dataset, it required depth maps to constrain error variance, reaching $2.70$ MAE and $23.92$ MSE.

Conversely, \textbf{Nvidia Cosmos (32B Dense)} and \textbf{Zhipu GLM 4.6V (10B Dense)} failed to effectively leverage both 2D segmentation and depth prompts, plateauing within a high error regime ($2.29$ to $2.52$ MAE) on LiveUAMap. Owing to this inability to parse spatial context, both models were excluded from subsequent ArcGIS evaluations to conserve computational resources.

Ultimately, integrating depth maps allowed leading LVLMs to surpass standard deterministic baselines. On LiveUAMap, the Gemma 4 \textit{+Depth} configuration outperformed SAMGeo Google (MAE $1.11$ vs. $1.58$; MSE $10.23$ vs. $21.98$) and SAMGeo ESRI (MAE $1.11$ vs. $1.17$; MSE $10.23$ vs. $14.61$). On ArcGIS, Gemma 4 \textit{+Depth} decisively beat SAMGeo ESRI (MAE $2.08$ vs. $2.42$) and matched SAMGeo Google's pure counting precision (MAE $2.08$ vs. $2.07$) while offering far greater stability against severe outlier errors (MSE $14.28$ vs. $19.26$).


\begin{figure}[!htbp]
    \centering
    \includegraphics[width=\linewidth]{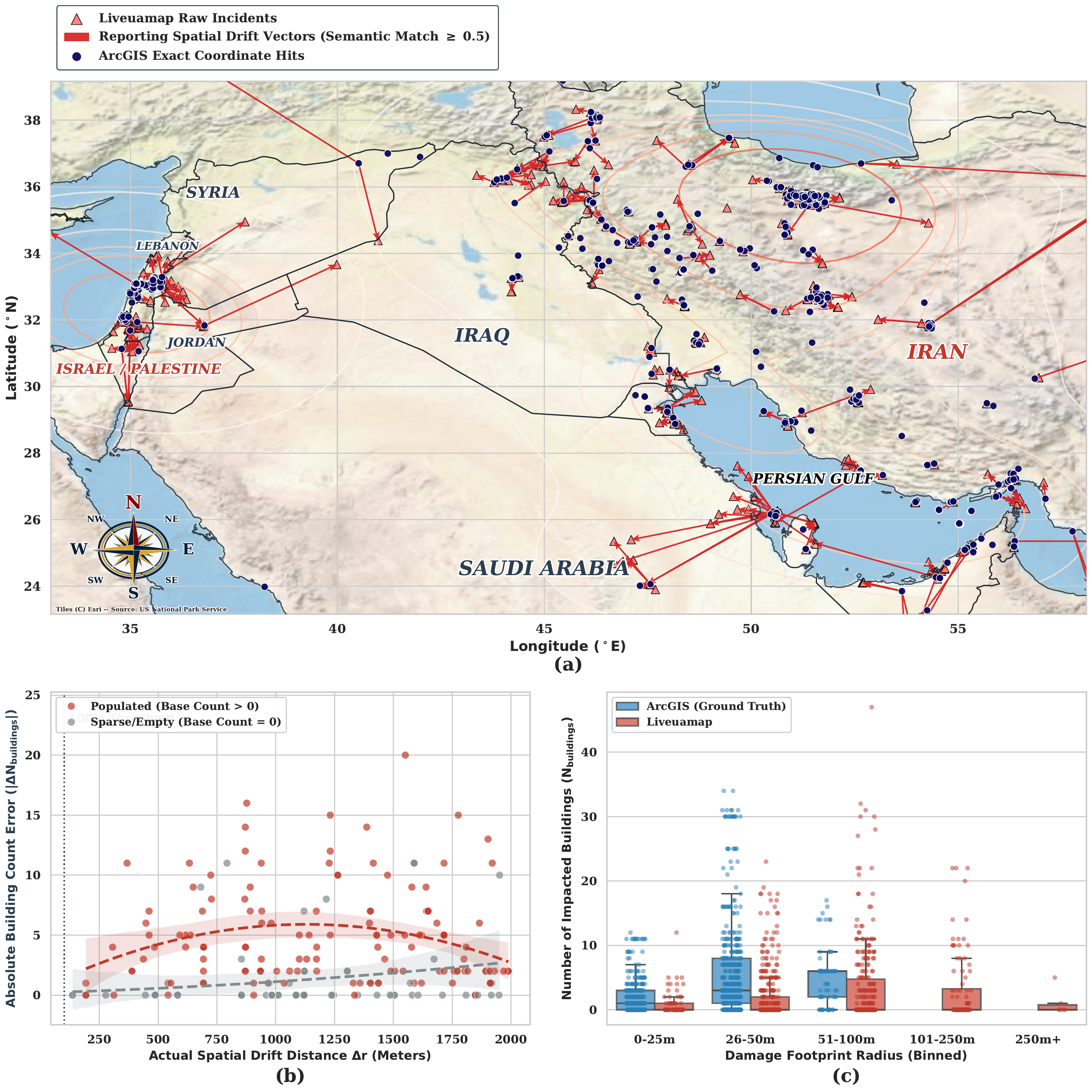}
    \caption{\textbf{Geospatial Distribution, Incident Drift, and Building Exposure Error Analysis (Liveuamap vs. ArcGIS).} 
    \textbf{(a)} Spatial mapping of event coordinates across the Middle East. Blue markers indicate exact ArcGIS ground truth coordinates, while red triangles denote raw Liveuamap incident reports. Red arrows represent spatial drift vectors linking semantically matched events (semantic match $\geq$ 0.5). 
    \textbf{(b) Sensitivity to Drift:} Correlation between actual spatial drift distance ($\Delta r$ in meters) and the absolute error in estimated building exposure counts. Data is stratified by populated areas (red) versus sparse/empty areas (grey), overlaid with polynomial regression trend lines. 
    \textbf{(c) Sensitivity to the Damage Footprint:} Box and strip plots comparing the distribution of impacted building counts between the ArcGIS (blue) and Liveuamap (red) datasets, categorized by binned damage footprint radii.}
    \label{fig:spatial_distribution_arcgis_vs_liveuamap}
\end{figure}
\begin{figure}[!htbp]
    \centering
    \includegraphics[width=\linewidth]{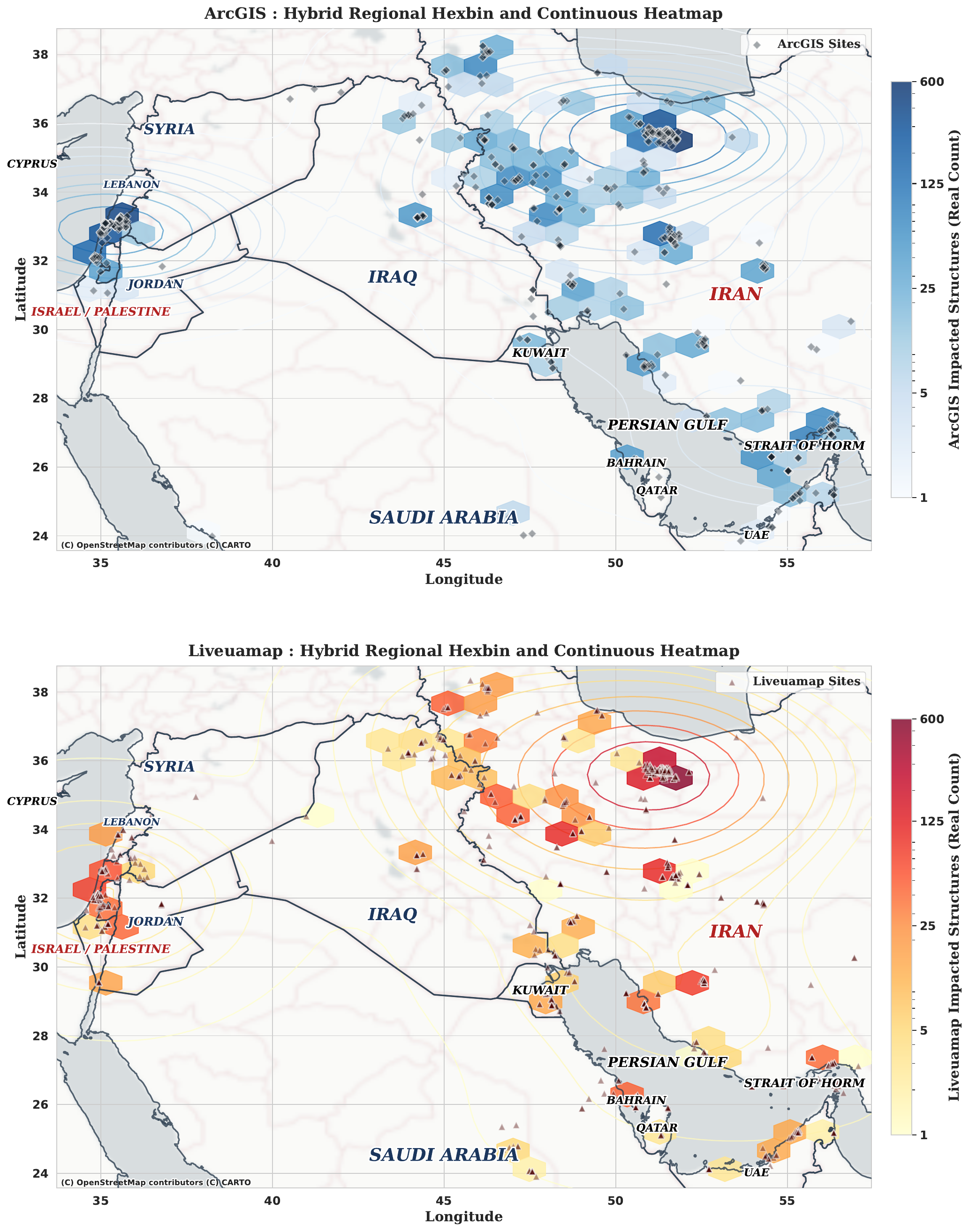}
    \caption{\textbf{Comparison of Structural Loss Heatmaps.} Regional loss aggregation generated from ArcGIS (\textbf{top}) and LiveUAMap (\textbf{bottom}) data. Hexagonal binning tracks cumulative structural impact using a logarithmic density scale, while overlaid contours delineate high-density exposure hotspots.}
    \label{fig:hybrid_damage_heatmaps}
\end{figure}

\subsubsection{Density-Stratified Regional Performance Profiles}
\label{subsubsec:density_stratification}

\paragraph{Aggregated Configuration Trends Across Density Tiers}

Across both datasets and all building density levels, adding relative depth maps systematically reduces counting errors by resolving spatial ambiguity. As detailed in Tables~\ref{tab:liveuamap_stacked_summary} and \ref{tab:arcgis_stacked_summary} and visualized in Figs.~\ref{fig:vlm_spatial_density_dashboard}f--g and \ref{fig:arcgis_results_llm}f--g (Supplementary Materials) across five density tiers (low, $0$--$5$, to extreme, $26+$ structures), depth integration consistently improves performance. On LiveUAMap, depth cues deliver major error reductions at both ends of the density spectrum: in low-density scenes ($0$--$5$), depth cuts baseline MAE by 43.3\% ($0.86 \to 0.49$) and MSE by 56.2\% ($3.90 \to 1.71$), while in extreme crowds ($26+$), depth lowers baseline MAE by 25.9\% ($32.38 \to 24.00$) and slashes MSE by 61.3\% ($1830.98 \to 708.08$). The ArcGIS dataset shows an even stronger improvement across every single tier: low-density MAE drops by 22.8\% ($1.54 \to 1.19$), medium-low ($6$--$10$) MAE drops by 35.9\% ($5.33 \to 3.42$; MSE $34.92 \to 18.50$), medium ($11$--$15$) MAE drops by 33.7\% ($9.59 \to 6.36$; MSE $104.12 \to 58.81$), and extreme-density MSE falls by 42.5\% ($786.62 \to 452.10$).

\paragraph{Detailed Region-wise Analysis of Each Model}
As visualized in Figs.~\ref{fig:vlm_spatial_density_dashboard}e1--e3 and \ref{fig:arcgis_results_llm}e1--e3 (Supplementary Materials) and
Granular model-level breakdowns in Supplementary Tables~\ref{tab:liveuamap_detailed_mae}--\ref{tab:arcgis_detailed_mse} reveal distinct architectural responses to multimodal feature integration:

\noindent\textbf{High-density robustness (Gemma 31B).} Dense architectures handle crowded scenes best because they reliably process and integrate 3D depth tokens. On LiveUAMap, adding depth consistently improves Gemma's performance as density grows, cutting MSE by 63.1\% in medium-low scenes ($35.77 \to 13.19$), 64.2\% in medium scenes ($64.12 \to 22.94$), and 82.8\% in high-density scenes ($258.00 \to 44.50$). In extreme ArcGIS scenes ($26+$ buildings), depth-augmented Gemma cuts MAE by 52.6\% ($33.14 \to 15.71$) and MSE by 72.4\% ($1131.14 \to 312.00$). Crucially, Gemma’s extreme-tier MSE ($312.00$) beats standard baselines: it outperforms Reference Google ($517.00$) by 39.6\% and Reference ESRI ($\text{MAE } 20.14$ vs. Gemma's $15.71$) by 22.0\%.

\noindent\textbf{Mid-tier optimization vs. extreme collapse.} Mixture-of-Experts (MoE) models excel in moderate crowds, but break down when scenes get too congested. Claude (Qwen Dist) leads in medium density, cutting LiveUAMap medium-tier MSE by 44.4\% ($62.53 \to 34.76$), high-tier MSE by 54.6\% ($163.77 \to 74.32$), and setting a record-low ArcGIS medium-low MSE of $11.82$ (beating Google's $13.61$ and ESRI's $27.50$). However, in extreme density ($26+$ buildings), extra depth visual data overloads MoE models. Claude's extreme ArcGIS MSE worsens by 46.6\% ($398.14 \to 583.86$), and Qwen 3.6 suffers a 101.3\% MSE penalty on LiveUAMap ($280.80 \to 565.20$). Interestingly, basic Qwen 3.6 without depth handles extreme crowds surprisingly well on LiveUAMap ($\text{MAE } 14.00, \text{MSE } 280.80$), showing that plain models can outperform MoEs before extra visual data causes a collapse.

\noindent\textbf{Sparse-zone sensitivity and invariance.} Adding depth map overlays does not always help, as extra visual data can confuse models in empty areas or be completely ignored. In sparse areas ($0$--$5$ buildings), depth overlays add unnecessary visual noise. This increases low-density MSE for Qwen 3.6 on ArcGIS by 24.1\% ($2.82 \to 3.50$) and for Nvidia Cosmos on LiveUAMap by 67.0\% ($1.24 \to 2.06$). Meanwhile, Zhipu GLM 4.6V ignores depth prompts entirely, with performance barely changing (between $-15.0\%$ and $+0.0\%$) across all LiveUAMap tiers, proving it cannot process depth visual inputs effectively.

\noindent\textbf{Polarized baseline dynamics.} Standard 2D segmentation models work great in simple scenes, but completely fall apart in dense urban areas where 3D depth is needed. In sparse scenes, standard 2D tools set the best baseline (LiveUAMap Google MAE $0.30$, MSE $0.92$; ArcGIS MAE $0.82$, MSE $1.80$). However, as scenes get crowded, touching roofs with similar colors blend together, causing severe undercounting. On LiveUAMap, SAMGeo Google’s MAE jumps from $0.30$ in low density to $13.86$ in high density, reaching a worst-case MAE of $30.60$ and MSE of $972.60$ in extreme crowds. Similarly, SAMGeo ESRI on ArcGIS degrades from a low-density MSE of $3.15$ to $180.45$ (high) and $492.43$ (extreme). In contrast, depth-augmented Gemma ($312.00$ MSE) reduces extreme-density error variance by 39.6\% compared to SAMGeo Google and 36.6\% compared to SAMGeo ESRI.

\begin{figure*}[!htbp]
    \centering
    \includegraphics[width=\linewidth]{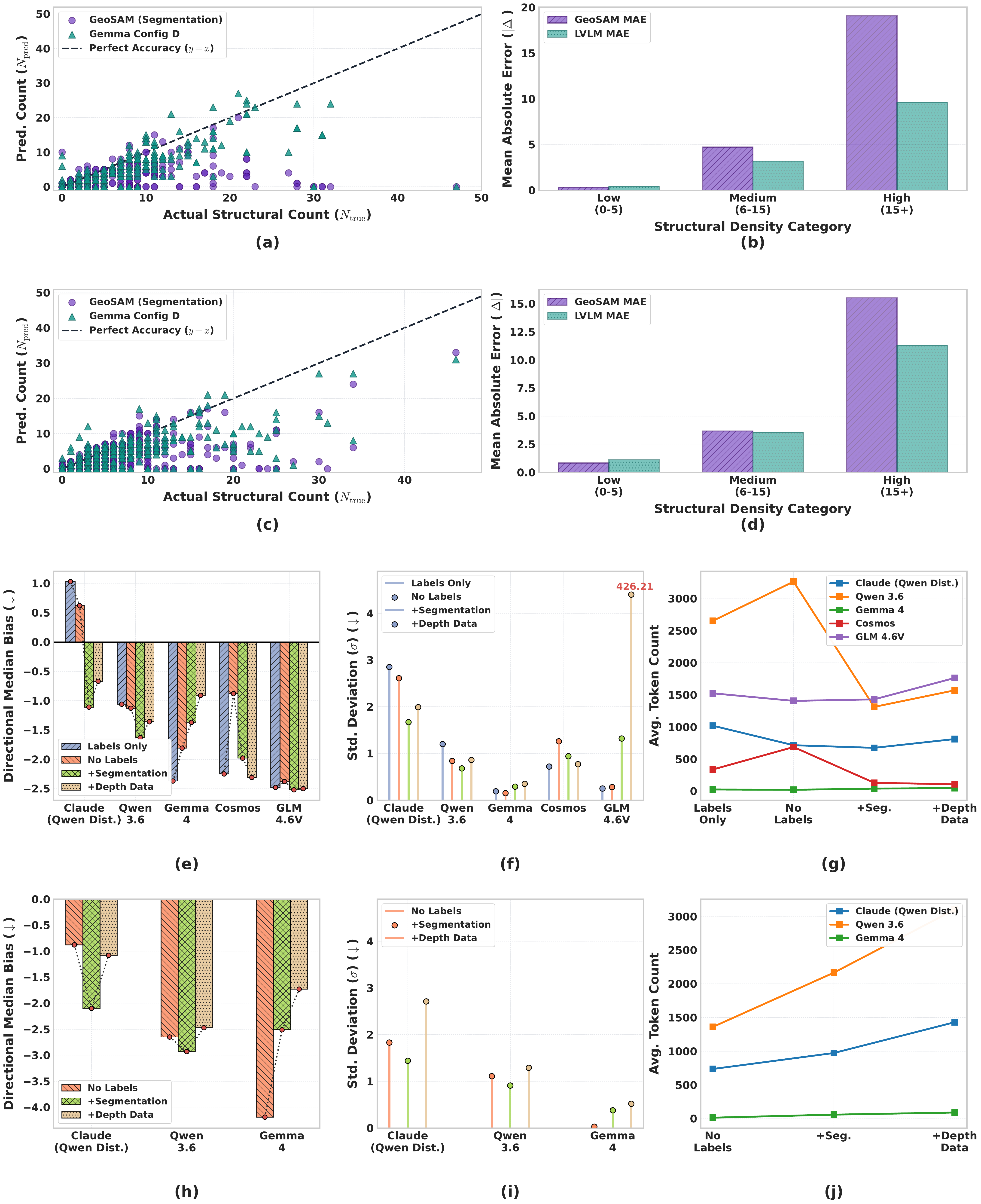}
    
    \caption{\textbf{Comprehensive Structural Evaluation Dashboard (LiveUAMap and ArcGIS).} 
    \textbf{(a--d) Best-Performing LVLM vs. Deterministic Baseline:} Scatter plots \textbf{(a, c)} compare predicted ($N_{\text{pred}}$) versus ground-truth ($N_{\text{true}}$) structural counts for the deterministic baseline (\textit{GeoSAM}) and our best-performing multimodal model (\textit{Gemma Config D}) against ideal parity ($y=x$). Error profiles \textbf{(b, d)} illustrate Mean Absolute Error (MAE) across low ($0\text{--}5$), medium ($6\text{--}15$), and high ($15+$) structural density tiers for the LiveUAMap and ArcGIS datasets, respectively. 
    \textbf{(e--j) System Bias, Variance, and Token Overhead:} Evaluation of model stability across five independent inferences of the same query. These panels detail the directional system bias ($\downarrow$) \textbf{(e, h)}, inference variance consistency ($\sigma \downarrow$) \textbf{(f, i)}, and average token generation overhead \textbf{(g, j)} for each evaluated model across different prompting configurations.}
    \label{fig:comprehensive_evaluation}
\end{figure*}

\paragraph{Best Performing Model Comparison (Gemma Config D vs. GeoSAM)}
Visual diagnostic distributions (Figures~\ref{fig:comprehensive_evaluation}a--d) across both benchmarks illustrate the mechanical divergence between deterministic 2D segmentation (GeoSAM) and depth-augmented LVLMs (Gemma Config D). Scatter plots demonstrate that while GeoSAM predictions align closely with the $y=x$ ideal line in sparse scenes, they horizontally collapse toward severe under-predictions as ground-truth counts exceed 15 structures. This density degradation is reflected in the Mean Absolute Error (MAE) profiles: while both models perform comparably in Low ($0$--$5$) density scenes ($\text{MAE} < 1.2$), GeoSAM's MAE surges in High ($15+$) density contexts to $19.1$ on LiveUAMap and $15.5$ on ArcGIS, whereas Gemma Config D mitigates MAE down to $9.6$ and $11.2$, respectively.

Global residual Kernel Density Estimation (KDE) plots (Figures~\ref{fig:supplementary_diagnostics}-left column) confirm that our top multimodal setup maintains a sharper, zero-centered residual peak ($\Delta = N_{\text{pred}} - N_{\text{true}}$), whereas pure 2D segmentation develops a heavy, negatively skewed tail extending past $\Delta = -40$. Density-stratified box plots (Figures~\ref{fig:supplementary_diagnostics}-right column) map this skewed tail directly to the High ($15+$) density category, where GeoSAM’s interquartile range (IQR) plunges deep into the negative domain with median errors reaching $-18$ on LiveUAMap and $-15$ on ArcGIS, accompanied by catastrophic outliers down to $-47$. By leveraging relative depth gradients as spatial regularizers, Gemma Config D truncates this negative residual tail, elevates high-density median errors closer to zero, and eliminates the chronic under-counting that fundamentally limits 2D segmentation tools in congested urban footprints.


\begin{figure}[!htbp]
    \centering
    \includegraphics[width=\linewidth]{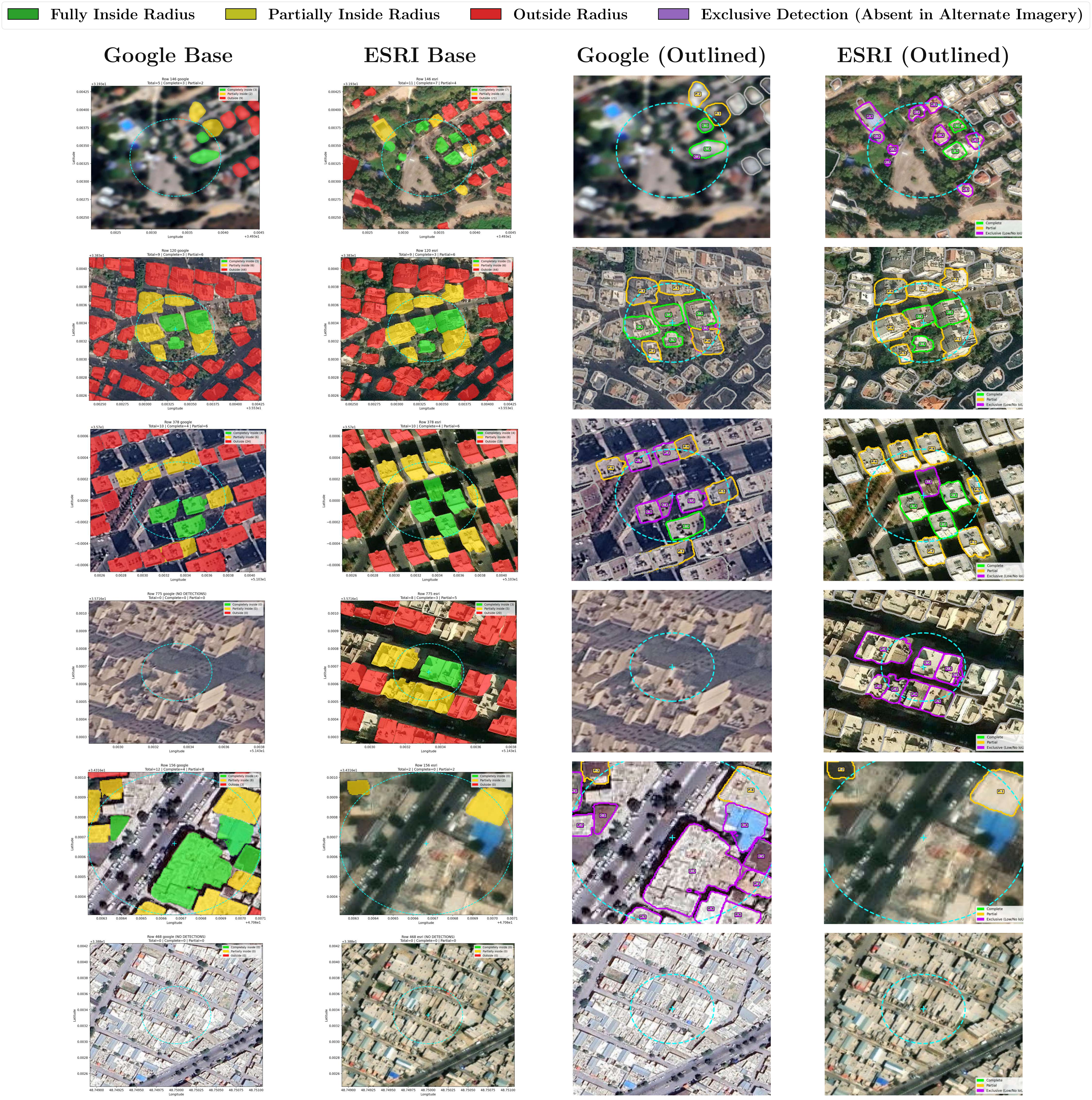}
    \caption{\textbf{Qualitative Segmentation Performance and Vulnerability Analysis of SAMGeo.} Building footprint extractions across Google and ESRI basemaps are categorized by spatial-boundary intersections: full containment within the target radius (green), partial intersection (yellow), total exclusion (red), and baseline-exclusive footprints missing from the opposing platform (purple). Rows 1--3 demonstrate robust, high-fidelity segmentation in high-contrast urban settings. Rows 4--5 highlight single-platform failures induced by localized imagery artifacts, including localized motion blurring on Google imagery (Row 4) and poor illumination contrast on ESRI imagery (Row 5). Row 6 demonstrates a systemic omission error on both platforms caused by severe off-nadir perspective skew and building height displacement, defining the operational boundary limits of the baseline segmentation architecture.}
\label{fig:geo_sam_performance_vis}
\end{figure}

\begin{figure}[!htbp]
    \centering
    \includegraphics[width=\linewidth]{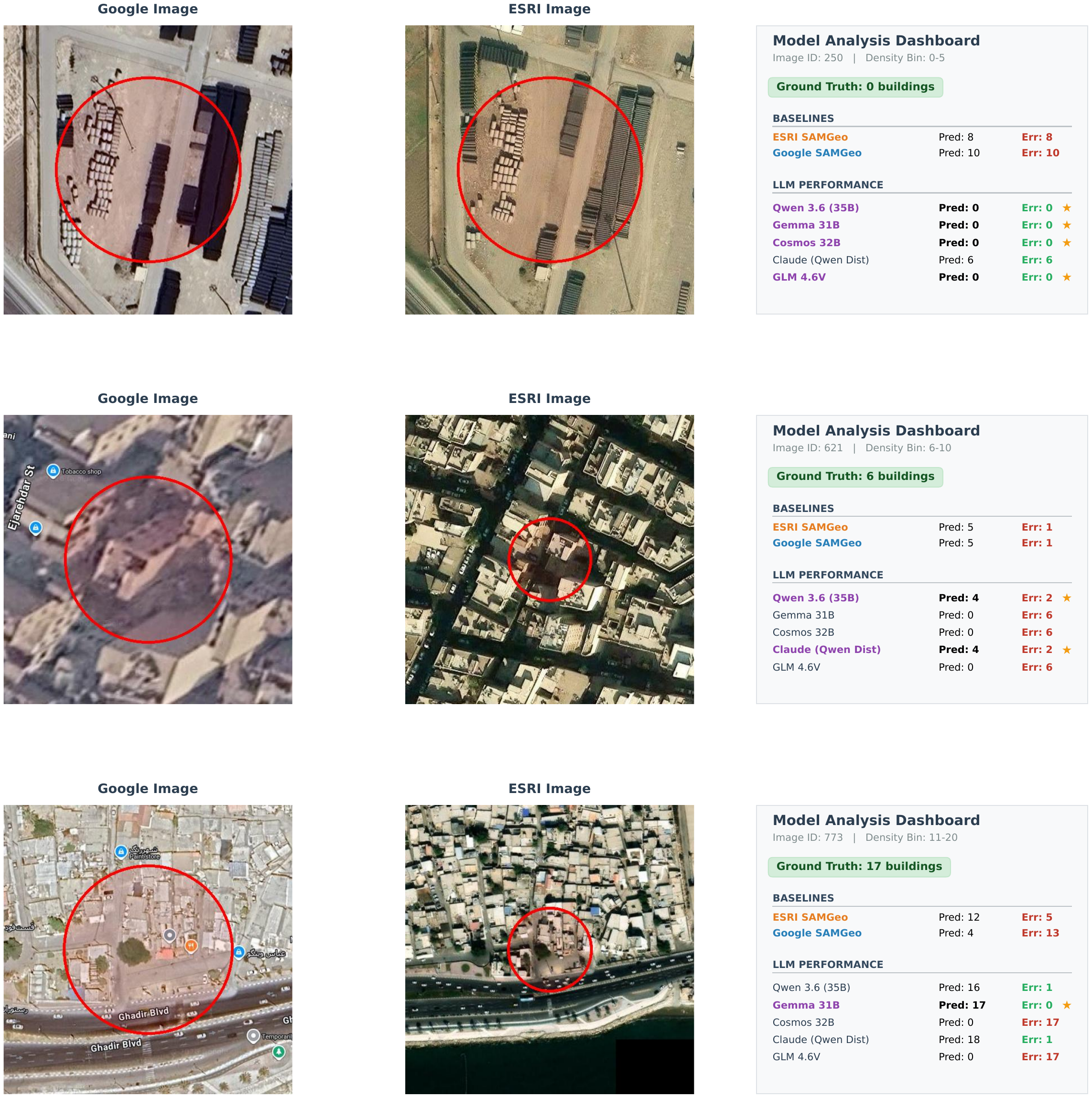}
    \caption{\textbf{Comparative Model Analysis Visualization for Unassisted LVLMs (Track-A, Config B).} Performance evaluation of baseline segmentation models (ESRI and Google SAMGeo) versus Track-A LVLM pipelines across three distinct structural density bins (Low: $0\text{--}5$, Moderate: $6\text{--}10$, and High: $11\text{--}20$). For each scene, input Google and ESRI satellite imagery are presented alongside a quantitative dashboard detailing predicted building counts and absolute errors relative to ground truth. Best-performing models achieving the lowest error in each scenario are denoted with a star ($\star$), highlighting the corrective capacity of vision-language models against the over-prediction and under-prediction tendencies of traditional baselines.}
\label{fig:llm_performance_vis}
\end{figure}

\begin{figure}[!htbp]
    \centering
    \includegraphics[width=\linewidth]{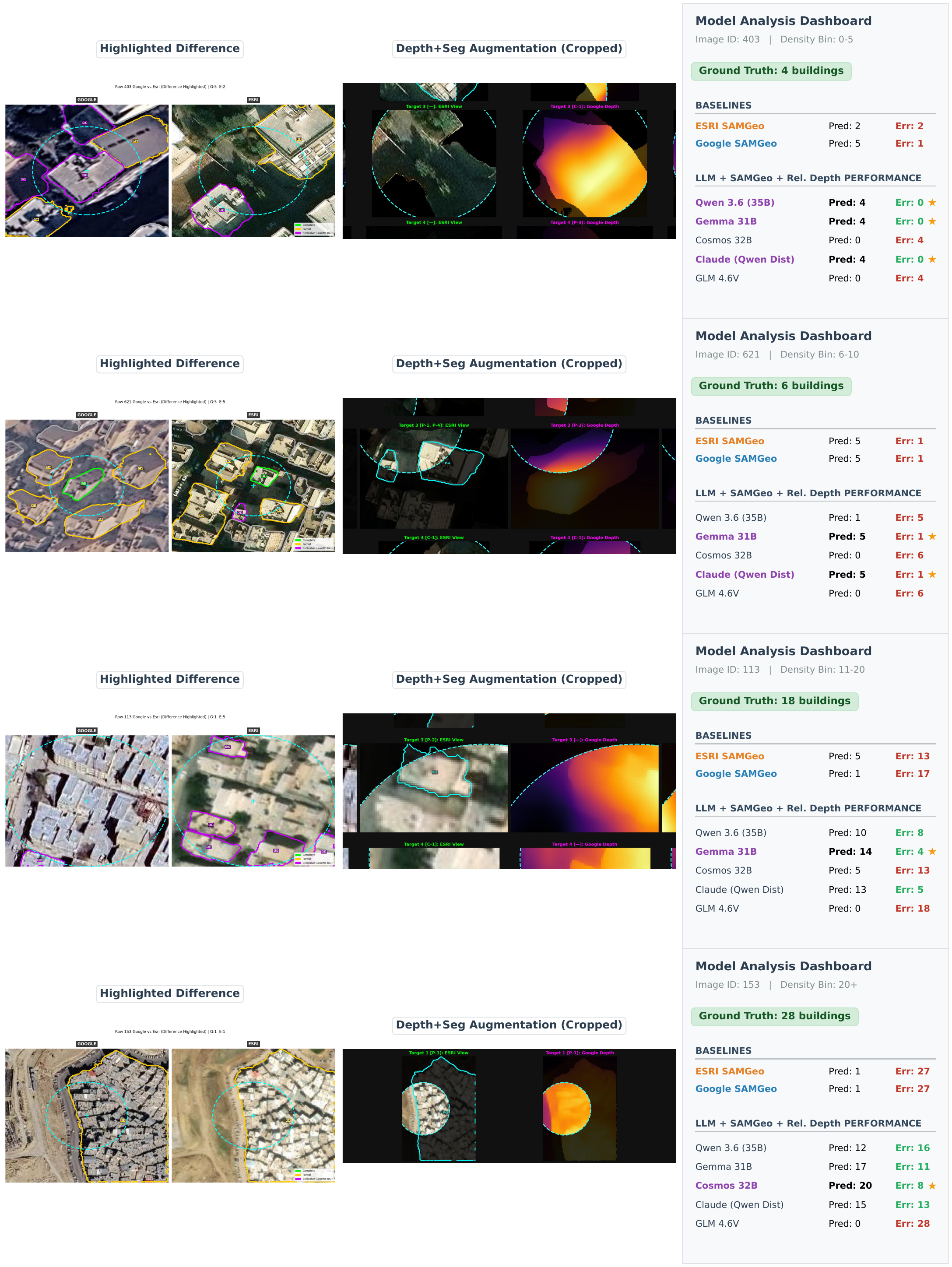}
    \caption{\textbf{Multimodal Performance Visualization for Track-B (Config D) Across Density Regimes.} Comprehensive qualitative and quantitative evaluation across four structural density bins: low ($0\text{--}5$), moderate ($6\text{--}10$), high ($11\text{--}20$), and ultra-high ($20+$). Each tier illustrates the structural input features provided to the networks via highlighted boundaries (left) and cropped relative depth-plus-segmentation maps (center), alongside a metrics dashboard (right) detailing predicted building counts and absolute error margins. Best-performing models within each cohort are indicated by a star ($\star$). The visualization highlights the efficacy of multimodal visual augmentations in mitigating baseline under-segmentation in highly clustered or occluded scenes, where architectures like Gemma~31B and Cosmos~32B recover substantial structural counts missed by traditional ESRI and Google SAMGeo baselines.}
\label{fig:llm_augmentation_performance_vis}
\end{figure}
\subsubsection{Generation Length, Output Variance, and Directional Bias}
\label{subsubsec:token_overhead_variance}

Evaluating model performance across five independent inference passes reveals distinct behaviors in how multimodal inputs impact text generation volume, repeatability, and directional error (Supplementary Tables~\ref{tab:bias_std_analysis}--\ref{tab:arcgis_word_count_analysis}, Figure~\ref{fig:comprehensive_evaluation}e--j).

\paragraph{Word Generation Bounds and Token Overhead}
Adding structural spatial modalities significantly alters output verbosity, exposing fundamental architectural differences in spatial prompt processing (Supplementary Tables~\ref{tab:word_count_analysis} and \ref{tab:arcgis_word_count_analysis}, Figures~\ref{fig:comprehensive_evaluation}g and \ref{fig:comprehensive_evaluation}j). Google Gemma 4 (31B Dense) demonstrates exceptional computational efficiency, maintaining highly concise responses under $100$ words across all prompting configurations ($13.41$ to $90.06$ words on ArcGIS; $21.28$ to $48.36$ words on LiveUAMap). Conversely, Mixture-of-Experts (MoE) models display extreme generation volatility. Qwen 3.6 (35B) undergoes a massive token expansion on ArcGIS, surging from $1360.30$ words at baseline to $3102.60$ words when incorporating continuous depth maps (\textit{+Depth}). On LiveUAMap, however, spatial overlays act as a prompt-constraining visual anchor, halving Qwen's $3265.18$-word baseline down to $1312.98$ words under 2D segmentation (\textit{+Seg}). Meanwhile, Zhipu GLM 4.6V operates within a rigid, heavy footprint ($1406.93$ to $1764.56$ words) regardless of the visual modality provided.

\paragraph{Output Variance and Inference Repeatability}
Measuring output variance across five independent queries per location evaluates each model's deterministic stability (Supplementary Tables~\ref{tab:bias_std_analysis} and \ref{tab:arcgis_bias_std_analysis}, Figures~\ref{fig:comprehensive_evaluation}f and \ref{fig:comprehensive_evaluation}i). Google Gemma 4 achieves near-perfect output reproducibility in unassisted visual settings ($\sigma = 0.03$ on ArcGIS; $\sigma = 0.15$ on LiveUAMap). For MoE architectures, 2D segmentation overlays (\textit{+Seg}) serve as the most effective variance stabilizers, constraining repeated counting fluctuations for Claude ($\sigma = 1.67$ on LiveUAMap; $\sigma = 1.44$ on ArcGIS) and Qwen 3.6 ($\sigma = 0.68$ on LiveUAMap; $\sigma = 0.91$ on ArcGIS). In contrast, Zhipu GLM 4.6V experiences a catastrophic architectural failure under continuous depth rendering (\textit{+Depth}): its output standard deviation explodes to an unmanageable $\sigma = 426.21$ on LiveUAMap, demonstrating an inability to consistently process dense spatial depth encodings without destabilizing its autoregressive sampling.

\paragraph{Directional Median Bias and Error Correction}
Tracking system bias magnitude isolates whether models systematically over-count or under-count impacted structures relative to ground truth (Supplementary Tables~\ref{tab:bias_std_analysis} and \ref{tab:arcgis_bias_std_analysis}, Figures~\ref{fig:comprehensive_evaluation}e and \ref{fig:comprehensive_evaluation}h). Unassisted visual baselines exhibit a pervasive negative directional bias across nearly all models, confirming a widespread tendency toward under-counting in complex satellite imagery. Claude (Qwen Distill) presents a notable exception on LiveUAMap, showing slight positive over-counting at baseline ($+1.03$ with labels; $+0.62$ without labels; Figure~\ref{fig:comprehensive_evaluation}e) before shifting to minor under-counting when visual overlays are introduced ($-1.11$ for \textit{+Seg}; $-0.67$ for \textit{+Depth}). Crucially, explicit spatial prompting acts as a powerful bias corrector. Google Gemma 4 suffers severe baseline under-counting on ArcGIS ($-4.19$; Figure~\ref{fig:comprehensive_evaluation}h), but adding continuous depth maps (\textit{+Depth}) drastically recovers lost detections, reducing bias to a cohort-best $-1.73$ on ArcGIS and $-0.91$ on LiveUAMap.
\begin{table}[!htbp]
\centering
\small

\caption{\textbf{Comprehensive Structural Assessment Matrix.} (Top) One-Sided Paired Wilcoxon Signed-Rank Significance Matrix ($p$-values) evaluating directional reductions in the Absolute Error (calculated using the median of 5 independent inference runs per datapoint). Bold formatting with asterisks denotes that the augmented configuration statistically outperforms the preceding setup ($p < 0.05$). (Bottom) Comparative trade-off analysis and density-stratified deployment paradigm across distinct urban density distributions.}
\label{tab:comprehensive_analysis}

\begin{tabular}{llccc}
\toprule
\multicolumn{5}{l}{\textbf{Panel A: Statistical Significance ($p$-values) of Absolute Error Reduction ($\downarrow$)}} \\
\midrule

\textbf{Dataset} & \textbf{Model Variant} & \textbf{Baseline $\rightarrow$ +Seg} & \textbf{+Seg $\rightarrow$ +Depth} & \textbf{Baseline $\rightarrow$ +Depth} \\ 
\midrule
\textit{ArcGIS}  & Qwen 3.6 (35B)         & 0.9993                 & $\boldsymbol{< 0.0001^{***}}$ & 0.1162                       \\
                 & Gemma 31B              & $\boldsymbol{< 0.0001^{***}}$ & $\boldsymbol{< 0.0001^{***}}$ & $\boldsymbol{< 0.0001^{***}}$    \\
                 & Claude (Qwen Dist)     & 0.6744                 & $\boldsymbol{< 0.0001^{***}}$ & $\boldsymbol{< 0.0001^{***}}$    \\ 
\midrule
\textit{LiveUAMap}& Qwen 3.6 (35B)        & 0.7954                 & $\mathbf{0.0305^{*}}$  & 0.3198                       \\
                 & Gemma 31B              & $\boldsymbol{< 0.0001^{***}}$ & $\boldsymbol{< 0.0001^{***}}$ & $\boldsymbol{< 0.0001^{***}}$    \\
                 & Cosmos 32B             & 1.0000                 & 0.9809                    & 1.0000                       \\
                 & Claude (Qwen Dist)     & $\boldsymbol{< 0.0001^{***}}$ & 0.1038                    & $\boldsymbol{< 0.0001^{***}}$    \\
                 & GLM 4.6V               & 0.9959                 & 0.1852                    & 0.9774                       \\ 
\bottomrule
\end{tabular}

\vspace{0.6cm} 

\begin{tabular}{@{} p{3.2cm} p{5.5cm} p{5.5cm} @{}}
\toprule
\multicolumn{3}{l}{\textbf{Panel B: Trade-off Analysis \& Density-Stratified Deployment Paradigm}} \\
\midrule
\textbf{Parameter} & \textbf{Programmatic Segmentation (SAMGeo)} & \textbf{Depth-Augmented LVLM} \\ 
\midrule
\textbf{Counting Accuracy} & High & Highest \\
\textbf{Computational Cost} & Lower & Higher \\
\textbf{Runtime} & Moderate (constrained by single-image processing; lacks batching). & Model-dependent (Fast for Gemma due to batching; slower for MoE). \\ 
\midrule
\multicolumn{3}{l}{\textbf{Density-Stratified Deployment Recommendation}} \\ 
\midrule
\textbf{Low-Density \newline (0--5 structures)} & \textbf{Recommended.} Most computationally efficient. Establishes minimum error floors while avoiding latency. & \textbf{Base Configs Sufficient.} Depth augmentation introduces variance and is computationally wasteful. \\
\textbf{Moderate-Density \newline (6--15 structures)} & Viable baseline, but begins to degrade as structural occlusion and grouping increase. & \textbf{Tipping Point.} Depth augmentation emerges as highly effective, drastically reducing MSE over base models. \\
\textbf{Extreme-Density \newline (16+ structures)} & Fails due to catastrophic under-segmentation boundaries and severe visual occlusion. & \textbf{Strictly Required.} Synergistic augmentation leverages 2.5D Z-axis data to suppress exponential error scaling. \\ 
\bottomrule
\end{tabular}
\end{table}
\subsubsection{Statistical Validation of Contextual Augmentations}
\label{subsubsec:statistical_results}

Table~\ref{tab:comprehensive_analysis} (Panel A) summarizes the resulting $p$-values from the Wilcoxon signed-rank tests across both the ArcGIS and LiveUAMap evaluation streams. Globally, depth-augmented architectures demonstrate a statistically significant reduction in counting error ($p < 0.05$) compared to their respective visual baselines, mathematically confirming that 2.5D pseudo-height cues provide robust spatial regularization. 

However, the trajectory and significance of these improvements vary considerably across model families. Notably, Gemma 31B exhibits a universal and highly significant error reduction ($p < 0.0001$) across every progressive augmentation step on both datasets. Similarly, Claude (Qwen Dist) demonstrates strong end-to-end significance (Baseline vs. +Depth, $p < 0.0001$), despite dataset-dependent sensitivity to intermediate 2D segmentation. In contrast, architectures such as Cosmos 32B and GLM 4.6V fail to extract statistically meaningful spatial reasoning improvements from the visual augmentations ($p > 0.10$), underscoring a stark divergence in cross-modal alignment and spatial grounding capabilities across base foundation models.

\subsection{Qualitative Analysis of Model}
\label{subsec:qualitative_analysis}

\subsubsection{Segmentation Vulnerabilities (SAMGeo)}
Visualizations of SAMGeo footprint extraction (Figure~\ref{fig:geo_sam_performance_vis}) reveal critical environmental vulnerabilities. While the model accurately segments structures in high-contrast imagery in the first three rows, it is highly sensitive to basemap quality. Localized blurring on Google basemaps in Row 4 leads directly to missed target detections (showing no detections), whereas it successfully captures structures on the ESRI base. Conversely, Row 5 demonstrates the model successfully capturing footprints via Google imagery while struggling with the ESRI basemap, where low contrast causes it to miss several structures.

\subsubsection{LVLM Contextual Filtering and Depth Recovery}
\label{subsubsec:qualitative_case_studies}
Comparisons of unassisted LVLMs (Figure~\ref{fig:llm_performance_vis}) and depth-augmented configurations (Figure~\ref{fig:llm_augmentation_performance_vis}) demonstrate how vision-language scaling resolves deterministic blind spots.

\textbf{Noise Filtering in Sparse Scenes.} In zero-building environments (Figure~\ref{fig:llm_performance_vis}, Row 1; Ground Truth: 0), SAMGeo models hallucinate structures—predicting 8 (ESRI) and 10 (Google)—by misinterpreting background noise. Conversely, unassisted LVLMs (Qwen 3.6, Gemma 31B, Cosmos 32B) correctly output zero. This reliability is maintained with depth augmentation (Figure~\ref{fig:llm_augmentation_performance_vis}, Row 1; Ground Truth: 4), where depth cues cleanly guide Qwen 3.6, Gemma 31B, and Claude to perfect predictions of 4 structures, while SAMGeo models under-predict (2) and over-predict (5).

\textbf{Resolving Occlusion in Extreme Density.} In congested environments (Figure~\ref{fig:llm_performance_vis}, Row 3; Ground Truth: 17), SAMGeo suffers under-segmentation collapse, predicting only 12 (ESRI) and 4 (Google) structures. Unassisted Gemma 31B perfectly matches the ground truth here (17), and Qwen 3.6 predicts 16. 

The distinct advantage of depth cues emerges in ultra-high density scenarios. In a highly occluded scene (Figure~\ref{fig:llm_augmentation_performance_vis}, Row 3; Ground Truth: 18), ESRI and Google SAMGeo baselines collapse to just 5 and 1 detections, respectively. In contrast, Gemma 31B uses depth layers to retrieve 14 structures. This trend peaks in extreme occlusion (Figure~\ref{fig:llm_augmentation_performance_vis}, Row 4; Ground Truth: 28), where both SAMGeo models detect a single building. Here, depth-augmented Cosmos 32B successfully leverages vertical topography to separate merged rooftops, leading the cohort by recovering 20 structures.
\subsection{Distribution of Conflict Incidents}
\label{subsec:conflict_distribution}

\subsubsection{Geospatial Drift and Coordinate Precision}
Figure~\ref{fig:spatial_distribution_arcgis_vs_liveuamap}a and Supplementary Figure~\ref{fig:geospatial_drift_dashboard} map the spatial drift of crowdsourced LiveUAMap alerts compared to precise ArcGIS entries. By combining nearest-neighbor matching with a 50\% semantic similarity threshold, we successfully paired 714 of 890 reported events. Urban corridors like Central Iran (Tehran and Isfahan) show strong geographic alignment, but distinct spatial divergences occur inland and across the Persian Gulf.

Across the 714 paired events, mean spatial drift is 34.87 km, but the median is only 4.07 km. This indicates that while most incidents align closely, extreme outliers (up to 3930.85 km) create a heavy long-tail distribution.

Reporting precision varies significantly by region. Spatial instability is highest in the Mesopotamian Dispersion ($n=51$; median 14.37 km, mean 40.35 km) and the Gulf Maritime Chokepoint ($n=54$; mean 56.12 km). Conversely, Central and Northern Iran ($n=282$, 39.5\% of matches) shows the tightest alignment with a median drift of just 2.75 km and a mean of 14.07 km. The Levant Convergence Zone ($n=111$) and Gulf Countries ($n=124$) exhibit moderate median drifts of 5.89 km and 4.30 km.

Directionally, reporting bias shows a slight aggregate drift toward the South-East (-0.0336$^\circ$ latitude, 0.0636$^\circ$ longitude). However, dispersion across all quadrants confirms these inaccuracies are multidirectional noise rather than a uniform offset.

\subsubsection{Sensitivity to Location Drift and Exposure Radius}
To evaluate how location inaccuracies affect building exposure estimates, we examined building count errors ($|\Delta N_{\text{buildings}}|$) relative to location drift distance ($\Delta r$) and analyzed exposed structural counts grouped by reported exposure radii. Detailed diagnostic metrics are provided in Supplementary Table~\ref{tab:sensitivity_analysis}.

\paragraph{Sensitivity to Spatial Drift}
Figure~\ref{fig:spatial_distribution_arcgis_vs_liveuamap}b shows how location displacement affects building count errors. In populated areas ($N_{\text{ArcGIS}} > 0$), minor location errors below 250\,m cause minimal discrepancy, averaging just 0.33 miscounted buildings (Supplementary Table~\ref{tab:sensitivity_analysis}). Beyond 250\,m, however, errors increase sharply to an average of 4.40 to 5.56 buildings per event—representing a 140\% to 185\% distortion over baseline averages—with extreme errors reaching 10 to 20 buildings. In sparse or rural areas ($N_{\text{ArcGIS}} = 0$), typical errors remain near zero (median of 0.0 buildings). Nevertheless, Figure~\ref{fig:spatial_distribution_arcgis_vs_liveuamap}b highlights notable exceptions where errors spike up to 8, 10, and 11 buildings at drift distances beyond 750\,m. These spikes occur when open-source reports record structural exposure where ground-truth records show none ($|\Delta N_{\text{buildings}}| = N_{\text{Liveuamap}}$). 

These findings establish 250\,m as a practical quality-filtering threshold for automated data processing, demonstrating that open-source reports with location errors exceeding 250\,m introduce severe count distortions ($>100\%$) and high error variation. Furthermore, the occurrence of multi-building errors in rural areas shows that low building density does not prevent false positives; uncorrected location shifts can still incorrectly attribute exposure to nearby structures.

\paragraph{Sensitivity to Exposure Footprint Radius}
Figure~\ref{fig:spatial_distribution_arcgis_vs_liveuamap}c compares building counts ($N_{\text{buildings}}$) across reported exposure radius categories ($0\text{--}25\,\text{m}$, $26\text{--}50\,\text{m}$, $51\text{--}100\,\text{m}$, $101\text{--}250\,\text{m}$, and $>250\,\text{m}$). In the ground-truth baseline (ArcGIS), recorded exposure is heavily concentrated within the $26\text{--}50\,\text{m}$ radius category (averaging 5.88 buildings per event), with no ground-truth events recorded above 100\,m in this sample (Supplementary Table~\ref{tab:sensitivity_analysis}). In contrast, open-source reports (Liveuamap) show a wider spread across radius categories, reflecting differences in how open-source platforms and official sources record exposure boundaries.

This comparison provides a practical benchmark for setting default search radii around incident points when specific weapon or payload data is unavailable. Specifically, a $26\text{--}50\,\text{m}$ search radius aligns best with validated exposure footprints for standard kinetic incidents, whereas expanding search radii beyond 100\,m risks overestimating exposure by including unimpacted surrounding structures in the count.
\subsubsection{Structural Exposure Analysis}
\label{sssec:building_count_analysis}

Reported incidents span critical geopolitical theaters, including the Levant (Israel, Lebanon, Syria), Central Israel, maritime chokepoints (Arabian Peninsula, Strait of Hormuz), and Iranian logistical hubs like Isfahan (Figure~\ref{fig:hybrid_damage_heatmaps}).

Country-wise aggregation (Supplementary Table~\ref{tab:country_damage}) reveals heavy spatial concentration. Iran (2,647 ArcGIS vs. 1,762 LiveUAMap structures) and the Israel/Palestine theater (954 vs. 240) absorb the vast majority of documented exposure. Secondary exposures occur in the UAE (64 vs. 33), Bahrain (43 vs. 45), Kuwait (30 vs. 21), Qatar (0 vs. 3), Iraq (9 vs. 4), and Saudi Arabia (5 vs. 0).

At the micro-level, Tehran is the most concentrated urban center (1,159 ArcGIS; 940 LiveUAMap). Other localized hotspots include Buqei'a in northern Israel (954 ArcGIS) and Sardasht in Iran (253 LiveUAMap).

An empirical distribution analysis of building exposure counts ($N$) (Supplementary Figure~\ref{fig:groundtruth_distribution}) shows right-skewed distributions for both sources. The ArcGIS mean ($\mu \approx 4.2$) tracks higher than LiveUAMap's ($\mu \approx 2.5$). Boxplots and violin profiles reveal wider dispersion and long upper tails (extending toward $N=50$) in ArcGIS. Cumulative probability functions confirm LiveUAMap saturates at lower building counts significantly faster than ArcGIS.

Importantly, these counts reflect theoretical exposure, not verified physical destruction. The metrics assume exact coordinates (rather than regional centroids) and ground-level kinetic impacts. In reality, mid-air interceptions or drifts into unpopulated terrain would significantly reduce or eliminate actual physical damage.
\section{Discussion}
\label{sec:discussion}

\subsection{Role of Context in Spatial Extent}
\label{subsec:discussion_zoom}

Satellite image segmentation requires balancing spatial context: over-constricting crops building edges, while over-expanding adds disruptive background noise. 

Large, fixed spatial extent (padding factor) degrades performance by injecting background clutter. For instance, increasing fixed padding from 1.0x to 2.5x (Liveuamap ESRI) spikes Mean Squared Error (MSE) from 14.06 to 90.63 and triples the Mean Absolute Error (MAE) (Table~\ref{tab:combined_total_main}). Conversely, a tight 1.0x crop excels on perfectly centered buildings---achieving a 0.77 MAE versus the Adaptive Selector's 0.81 (Table~\ref{tab:supp_complete})---but fails completely in partially inside structures (Table~\ref{tab:supp_partial}).

The Adaptive Selector resolves this trade-off by dynamically scaling the field of view based on model confidence. It retains accuracy on centered targets while correcting boundary errors, reducing MAE on partial structures by 25.4\% (Liveuamap Google) and 37.3\% (ArcGIS Google) compared to the rigid 1.0x crop. Across all evaluated metrics, dynamically tuning spatial context consistently outperforms static windows.

\subsection{LVLM Sensitivity and Token Footprints}

Mixture-of-Experts (MoE) models demonstrate volatile responses to spatial overlays. The base Qwen 3.6 (35B) achieves its lowest error natively; adding 2D masks or depth data degrades its accuracy. This supports the routing distraction hypothesis~\cite{xu2026seeing}, where dense visual overlays misdirect tokens away from logic modules into less effective visual layers. Consequently, Qwen 3.6 exhibits erratic generation lengths, exploding to over 3,100 words on depth-augmented ArcGIS, yet compressing by 50\% on Liveuamap. Conversely, the fine-tuned Claude-distilled MoE avoids this routing vulnerability, effectively leveraging relative depth to halve its Liveuamap MSE.

Dense architectures exhibit similar variability. While Nvidia Cosmos (32B) stagnates across all spatial configurations (bounded at an MAE of 2.29--2.38), Google Gemma 4 (31B) excels. Operating with extreme token efficiency (under 100 words), Gemma directly translates spatial inputs into structural corrections. This spatial integration shifts Gemma's severe ArcGIS underestimation bias from $-4.19$ to a cohort-best $-1.73$ and slashes extreme-density Liveuamap MSE by 82.8\%.

\subsection{Relative Depth Cues for Mitigating Extreme Occlusion}
In highly congested scenes (26+ structures), purely programmatic 2D visual boundaries degrade. Tight structural layouts and overlapping spectral signatures cause independent buildings to merge visually into single contiguous pixel clusters. This extreme spatial occlusion forces standard segmentation models into severe under-counting, a failure mode that cascades into the unassisted LVLMs as high negative system biases.

The empirical data demonstrates that relative depth maps systematically mitigate this specific limitation. By providing a 2.5D topographic representation, depth cues supply vertical anchor points that allow the generative models to separate overlapping objects based on physical proximity rather than visible 2D contours. This stabilization mechanism is directly reflected in the aggregated Liveuamap cohort data, where the \textit{+Depth} configuration plunges the baseline extreme-density MSE by 61.3\% (dropping from 1830.98 down to 708.08).

\subsection{Empty Scene Over-Analysis}
Conversely, in sparse environments (0--5 structures), spatial overlays disrupt structural detection. Absent dense target matrices, LVLMs over-analyze background noise and minor topographic variations, hallucinating them as physical structures. Consequently, adding depth induces a 24.1\% MSE degradation for Qwen 3.6 (ArcGIS) and a 67.0\% MSE degradation for Nvidia Cosmos (Liveuamap). In these low-density scenarios, traditional deterministic segmentation remains superior; the SAMGeo Google Reference establishes the lowest error threshold, outperforming all baseline and depth-augmented LVLM configurations.

\subsection{Mechanisms of spatial integration}
The empirical significance patterns (Table~\ref{tab:comprehensive_analysis}. Panel-A). reveal critical architectural dynamics regarding how LVLMs process spatial data. For several dominant architectures (Qwen 3.6, GLM 4.6V, and Cosmos 32B), the introduction of flat, programmatic 2D segments without depth cues yields exceptionally high $p$-values ($p > 0.99$). In a one-sided test maximizing error reduction, this mathematically demonstrates that the 2D segmentation stream actively disrupts counting capabilities for these models, confirming that flat, overlapping visual boundaries induce severe over-segmentation confusion. 

Conversely, the addition of relative depth maps triggers an immense, statistically definitive error reduction. On the precision-verified ArcGIS dataset, the transition to depth augmentation achieves overwhelming significance ($p < 0.0001$) across all evaluated models. By regularizing flat 2D scenes with pseudo-height variations and roof-pitches, the model transitions from arbitrary visual guesswork to structural topography, validating depth-augmentation as a crucial geometric regularizer.

\subsection{Density-stratified deployment paradigm}
Given that structural counting performance shifts dramatically based on localized scene congestion, deploying a single monolithic configuration is highly inefficient. Instead, the aggregated data supports a dynamic, density-stratified application paradigm (Table~\ref{tab:comprehensive_analysis}. Panel-B). 

When inference speed is prioritized and local building density is low (0--5 structures), traditional deterministic segmentation frameworks (SAMGeo) remain the most computationally efficient choice, establishing absolute minimum error floors (MAE $< 1.0$) while avoiding latency. Moderate-density scenes (6--15 structures) represent the critical tipping point for multimodal augmentation. While baseline segmentations remain viable, deploying depth-augmented frameworks introduces massive performance leaps, effectively resolving emerging visual occlusion. Finally, when maximizing precision across extreme-density footprints (16+ structures), depth augmentation transitions from beneficial to strictly essential. The Gemma architecture coupled with relative depth maps stands as the most robust tracking solution, utilizing 2.5D spatial context to completely bypass the catastrophic under-segmentation boundaries that cause traditional baselines to fail.

\subsection{Geospatial Analysis of Platform-Reported Incidence}

Comparing event coordinates and structural damage counts between Liveuamap (crowdsourced) and ArcGIS (systematic mapping) reveals how military interception capabilities and media reporting dynamics introduce systematic variance into open-source conflict data. To ensure analytical rigor, this framework applies strict inclusion constraints, yielding conservative damage estimates. The model processes only events with exact coordinate matches, excluding geographically ambiguous entries. Furthermore, multi-weapon logs are simplified to account only for the single weapon with the largest damage radius. Consequently, the resulting structural counts represent a lower-bound baseline rather than an exhaustive census of destruction.

\subsubsection*{Unintercepted Impacts and Structural Scaling in Iran}
In regions with limited multi-layered air defense infrastructure, higher ground-impact rates result in precise incident coordinates correlating closely with physical damage. The automated framework aggregates a national count of 2,723 affected structures in ArcGIS and 1,837 in Liveuamap. Both datasets indicate spatial clustering in major urban centers and border regions; Tehran accounts for 1,159 structures in ArcGIS and 940 in Liveuamap (Fig.~\ref{fig:hybrid_damage_heatmaps}), alongside notable impacts in Isfahan, Harsin, and Sardasht.

While these raw structural counts are conservative, they can be contextualized alongside state-reported figures when adjusted for local urban density. Official Iranian state sources reported approximately 81,000 damaged residential units within the first month~\cite{IranGov2026,NRC2026Iran}. To evaluate how baseline structural counts map onto unit-level estimates, we apply a representative urban density model for high-density Iranian civilian housing (10 stories at 4 units per floor):

\begin{equation}
\text{Civilian Units} = \text{Building Count} \times (10 \text{ floors} \times 4 \text{ units/floor})
\end{equation}

Applying this scaling factor of 40 units per building to our aggregated national counts (1,762 to 2,647 structures; see Section~\ref{sssec:building_count_analysis} and Table~\ref{tab:country_damage}) yields a projected range of 70,480 to 105,880 residential units. This bracket encompasses the reported 81,000-unit figure, demonstrating that the automated framework's structural detection aligns realistically with high-density architectural baselines.

\subsection{Limitations and Future Work}
The primary limitation of the proposed framework lies in its structural dependence on the fidelity of its initial ingestion streams. The alignment between the framework's exposure estimations and actual ground truth relies strictly on the precision of harvested geographical coordinates and the accuracy of weapon profiles extracted from unstructured text descriptions. Within these variables, coordinate precision is paramount; if the initial geolocations contain significant spatial errors, downstream algorithmic adjustments and vision-language enhancements cannot retroactively correct the underlying geometric displacement. 

Furthermore, coordinate uncertainty manifests systemically through several distinct real-world vectors that limit model reliability. First, because this framework processes crowdsourced feeds like LiveUAMap, it is inherently vulnerable to observer reporting bias, where ground users frequently log coordinates from where an explosion was witnessed or heard rather than its true point of impact. Second, active conflict environments are frequently subjected to intense electronic warfare, where widespread GPS/GNSS jamming and spoofing systematically corrupt the localized metadata captured by consumer devices. Finally, these spatial inaccuracies induce severe perimeter edge-effects; in highly dense, heterogeneous urban topographies, a minor coordinate shift of merely 10 to 30 meters completely alters the projected boundary circle, causing entire clusters of high-density structures to be erroneously included or excluded. Therefore, this framework must be treated as a predictive architecture for automating exposure estimation and humanitarian triage, rather than a definitive, post-hoc census of physical destruction.

To address spatial and source-fidelity challenges, future work should develop a dedicated ground reporting application that pairs precision location tagging with a verified ingestion backend. Unlike bloated social media platforms that strip metadata, this tool features a map-centric interface with mandatory live, in-app photo capture. Disabling gallery uploads prevents users from submitting recycled or synthetic media. Upon capture, a zero-trust verification pipeline extracts raw EXIF metadata, hardware sensor telemetry, and tamper-proof timestamps, cryptographically signing the payload directly on the device. To maintain operation during telecommunications blackouts, the system uses decentralized peer-to-peer (P2P) mesh networking and edge-computed perceptual hashing to locally synchronize and deduplicate uploads offline. These cryptographic and architectural safeguards isolate monitoring pipelines from unverified media, ensuring downstream automated damage models ingest reliable geospatial inputs.

Moreover, our pipeline utilizes standard explosive payload masses to systematically map reported weapon classes to their corresponding damage radii. This configuration allows for precise calibration should highly specific or localized payload metrics be preferred by the user. Because the structural reference counts are inherently defined by these spatial bounds, any adjustment to the baseline payload parameters naturally scales the evaluation area, which may slightly alter the resulting damage count. For this reason, the ground truth within this benchmark is best understood as a parameterized reference baseline calibrated to specific physical inputs, rather than a static absolute value. Crucially, however, while updating these payloads recalibrates the baseline building counts, our core technical findings—specifically the performance advantages, error-reduction dynamics, and structural behaviors of the depth-augmented models—remain completely invariant. The underlying algorithmic conclusions hold true across varying boundary definitions, as the comparative framework evaluates how effectively these models reason through spatial data within a given geographic constraint.

Finally, continuous advancements in computer vision segmentation architectures and large vision-language models (LVLMs) will likely enhance automated object counting and contextual extraction from text profiles. Beyond purely structural metrics, pairing these building-level damage exposure estimations with localized population density maps presents a promising avenue for future research to dynamically estimate potential civilian casualty rates and optimize automated humanitarian response logistics during active crises.

\subsection{Conclusion}
\label{sec:conclusion}
This paper presents an automated framework for rapid conflict damage assessment that bypasses commercial data blackouts by fusing open-source intelligence with physics-based blast radius modeling. Through systematic benchmarking of semantic segmentation engines and Large Vision-Language Models (LVLMs), we identify key mechanisms for optimizing geospatial context: an adaptive field-of-view routine stabilizes target counting across varying spatial extent, while relative depth integration provides vital $Z$-axis cues that resolve visual occlusion in dense urban environments (reducing error variance by up to 82.8\%). Conversely, our findings reveal that complex models tend to over-parameterize sparse rural scenes, where lightweight segmenters remain more reliable and cost-effective. Ultimately, operational efficiency requires a density-stratified hybrid deployment paradigm—routing sparse scenes to fast 2D segmentation baselines like SAMGeo while reserving depth-augmented LVLMs for complex, highly occluded urban footprints.
\section{Methodology}

\subsection{Damage radius estimation and pre-strike image collection}
\label{sec:method_radius_estimation}

This module translates qualitative open-source incident reports into quantitative spatial bounds for satellite imagery extraction. The pipeline ingests geolocated text alerts, calculates a conservative physical damage radius, and retrieves the corresponding imagery using an expanded contextual bounding box.

\subsubsection{Data acquisition and multi-source ingestion}
\label{sssec:incident_ingestion}

Data streams are collected from two open-source intelligence platforms: Liveuamap and curated ArcGIS StoryMaps. For the Liveuamap platform, each record contains a coordinate vector $\mathbf{x} = (\phi_{\text{lat}}, \lambda_{\text{lon}})$ and a short text title $T$. A rule-based keyword filter is applied to identify kinetic events, specifically targeting terms such as shelling, Shahed, SAM, rocket, drone, explosion, and missile. An event is retained if $T$ contains at least one of these keywords, while non-kinetic entries are discarded. In contrast, entries in the ArcGIS StoryMap platform include structured metadata such as actor, target, and spatial precision flags alongside coordinates and brief text descriptions. Only records explicitly marked with exact coordinates are retained from this source.

Because Liveuamap coordinates lack explicit spatial precision flags, they are treated as an unverified, spatially volatile baseline for early estimation testing, whereas the curated ArcGIS StoryMap records serve as the spatially precise data source. Following the application of the rule-based filters and precision sorting pipelines across both platforms, the ingestion engine successfully isolated a final corpus of 890 precise-coordinate events from the ArcGIS StoryMaps stream and 858 kinetic events from the Liveuamap platform.

\subsubsection{Weapon classification and language parsing}
\label{sssec:language_parsing}

For each retained incident, a two-stage language modeling pipeline parses the unstructured text description ($T$) into verified operational fields. This two-stage process separates reasoning from formatting. First, a reasoning-focused language model extracts semantic attributes from $T$ into an intermediate textual summary using chain-of-thought generation. The model isolates the weapon classification by identifying all referenced weapon types and quantities—selecting the class with the largest explosive baseline to ensure full spatial recall—and assesses the qualitative level of destruction to define a dynamic spatial modifier ($\Delta_{\text{modifier}}$). Second, a specialized parsing model ingests this intermediate reasoning output and maps it directly into a rigid JSON schema via a structured formatting template. This template enforces uniform formatting for the weapon classes ($w_{\text{class}}$) and structural impact labels, guaranteeing syntactical stability for the subsequent radius calculation steps.

\subsubsection{Physics-grounded damage radius estimation}
\label{sssec:radius_calculation}

The final spatial footprint is computed in two sequential steps. First, each weapon class is assigned a representative TNT-equivalent mass ($W$). The baseline operational radius ($R_{\text{base}}$) is computed using the Hopkinson--Cranz cube-root scaling law~\cite{hopkinson1915,cranz1926},
\begin{equation}
R_{\text{base}} = ZW^{1/3},
\end{equation}
where $W$ is the TNT-equivalent explosive mass and $Z=5\,\mathrm{m/kg^{1/3}}$ is the reference scaled distance. The selected value lies within the far-field blast regime ($Z>4\,\mathrm{m/kg^{1/3}}$), providing a physically motivated benchmark across threat categories. This yields baseline radii spanning, for example, $5\,\mathrm{m}$ for small kamikaze drones ($W=1\,\mathrm{kg}$), $20\,\mathrm{m}$ for heavy artillery ($W=64\,\mathrm{kg}$), and up to $60\,\mathrm{m}$ for ballistic missiles ($W=1{,}728\,\mathrm{kg}$). A complete parameter breakdown for all weapon classes and payloads (Table~\ref{tab:munition_payload_scaling}), along with their detailed justifications (Section~\ref{sec:radius_justification}), is provided in the Supplementary Materials.

Second, the baseline radius is dynamically adjusted using textual reports to account for situational variables, such as partial detonations, secondary explosions, or structural resistance:
\begin{equation}
R_{\text{moderate}} = \left\lfloor R_{\text{base}} \times \max(0.2, 1.0 + \Delta_{\text{modifier}}) \right\rfloor,
\end{equation}
where $\Delta_{\text{modifier}} \in \{-0.5, 0, +0.5\}$ quantifies the reported severity of the strike. This scales the effective damage boundary accordingly—for instance, adjusting the $20\,\mathrm{m}$ heavy artillery baseline across an operational window of $10\,\mathrm{m}$ ($\Delta_{\text{modifier}}=-0.5$) to $30\,\mathrm{m}$ ($\Delta_{\text{modifier}}=+0.5$). Because explosive blast volumes scale cubically ($V \propto R^3$), a simple linear adjustment to the radius mathematically reflects higher-order shifts in the destructive footprint. Rather than implying a literal increase in the weapon's physical payload, scaling the radius by a factor of 1.5 serves as a heuristic upper bound to model catastrophic secondary effects—such as the ignition of stored fuel or sympathetic detonation—that drastically expand the effective damage zone. The $0.2$ lower bound ensures the spatial footprint remains strictly positive, preventing computational errors during downstream segmentation.

\subsubsection{Pre-event satellite image retrieval}
\label{sssec:image_retrieval}

Using the coordinates $\mathbf{x}$ collected from the reporting platforms and the adjusted radius $R_{\text{moderate}}$, satellite imagery is retrieved via the ESRI World Imagery and Google Maps APIs, prioritizing the temporally closest available pre-event imagery. To incorporate surrounding contextual elements beyond the immediate damage boundary, a structural padding multiplier $\alpha$ is applied to expand the operational acquisition radius:
\begin{equation}
R_{\text{padded}} = \alpha \cdot R_{\text{moderate}}.
\label{eq:radius_pad}
\end{equation}

This modified radius, $R_{\text{padded}}$, defines the geographic bounding box for image extraction, determining both the final image dimensions and the optimal map zoom level. A higher $\alpha$ expands the bounding box and decreases the zoom level, providing a wider field of view for surrounding context, whereas a lower $\alpha$ constricts the bounding box, yielding a high-resolution focus on structural details. Based on this bounding box, map tiles are retrieved, stitched, and cropped exactly to the $R_{\text{padded}}$ boundary. While $\alpha$ adjusts the visual context and overall image size, the physical damage boundary remains strictly fixed at $R_{\text{moderate}}$ to evaluate the target area and render the downstream spatial overlays.

\subsection{Automated building counting framework}
\label{sec:automated_counting}

To count the structures within the estimated damage zone, the framework evaluates two distinct computational streams: an automated geometric segmentation stream and a multimodal vision-language modeling stream.

\subsubsection{Stream 1: Building segmentation and Adaptive Field-of-View selection}
\label{sssec:segmentation_stream}

To accurately process varying urban densities, this stream dynamically adjusts the map zoom level and spatial context rather than relying on a static viewpoint. Empirical testing indicates that building detection accuracy depends heavily on the visible geographic area surrounding the immediate damage zone. To determine the optimal spatial balance, the system evaluates six discrete scaling values ($\Omega = \{1.0, 1.5, 1.85, 2.0, 2.2, 2.5\}$), each of which directly dictates the operational zoom level and the extent of surrounding geographic context supplied to the model.

For each candidate scale factor $\alpha_j \in \Omega$, the satellite image footprint is processed by a geospatial segmentation model (SAMGeo) to detect building outlines. These raw outlines are optimized through three sequential cleanup steps. First, the segmentation model is queried using an ensemble of synonymous structural prompts (e.g., small buildings, houses, residential buildings) to generate a comprehensive pool of candidate footprints ($\mathcal{M}_{\text{raw}}$). Second, to eliminate internal nesting from semantically similar queries, the pipeline filters out auxiliary structural outlines ($M_s$) if more than $30\%$ of their area falls inside a primary outline ($M_b$), expressed as $|M_s \cap M_b| / |M_s| > 0.3$. Surviving auxiliary shapes are merged to form a cleaned candidate pool. Finally, an area-indexed Non-Maximum Suppression (NMS) routine is applied to collapse redundant boundaries into a single unique boundary by discarding any smaller outline demonstrating an Intersection-over-Union (IoU) $\geq 0.5$ ($50\%$) with a larger baseline outline.

Once the polygons are refined across all candidate padding factors ($\alpha_j$), they are evaluated by scoring building footprint alignment relative to $R_{\text{moderate}}$. Each detected building $M_k$ is assigned a spatial tracking weight ($\omega(M_k)$): $1.0$ if fully inside the circle (allowing for a $3\%$ boundary noise tolerance), $0.5$ if intersecting the perimeter, and $0.0$ if completely outside. To identify the most effective zoom level, the system computes an aggregate fitness score $\mathcal{S}(\alpha_j)$ for each candidate scale by summing the spatial weights of all buildings detected across both the Google and ESRI imagery. The framework selects the adaptive field-of-view scale $\alpha^*$ that maximizes this overall confidence score. If multiple padding factors yield identical scores, the pipeline defaults to the wider view ($\max \alpha$) to ensure downstream vision-language models receive maximum geographic context.

\subsubsection{Stream 2: Large vision-language model evaluation configurations}
\label{sssec:lvlm_stream}

The second stream evaluates the spatial reasoning and structure-counting capabilities of open-weight Large Vision-Language Models (LVLMs) within the estimated damage footprint. Rather than applying programmatic geometric filters, this stream assesses how a model visually interprets urban layouts when presented with different prompt variations. The models are benchmarked across four progressive configurations to systematically evaluate how the addition of textual, vector, and depth layers enhances counting accuracy.

In Configuration A (Raw Multimodal Baseline), the model receives raw, unannotated satellite imagery from Google Maps and ESRI side-by-side, guided only by a static red circle marking the target damage zone. Configuration B (Textual Label Anchoring) retains the red boundary circle but includes all geographic text tags natively generated by the map engines, explicitly prompting the model to transcribe these textual landmarks to anchor itself visually before counting. Configuration C (Multi-Source Mask Fusion) injects the clean building outlines generated by the Stream 1 pipeline directly onto the imagery as colored overlays. These shapes are color-coded based on their spatial alignment between datasets: green for verified structures fully inside the zone, yellow for structures crossing the boundary, and magenta for dissimilar structures detected on a single source. Finally, Configuration D (Topographic Height Stack) pairs the horizontal outlines from Configuration C with monocular vertical depth profiles generated via a depth-estimation model (Depth Anything). Mapped to a high-contrast Inferno colormap and combined with a spatial focus mask that dims neighboring context by $70\%$, this arrangement provides the model with explicit vertical pseudo-height cues to distinguish tightly packed structures.

\subsubsection{Data extraction and multi-run median filtering}
\label{sssec:consensus_protocol}

To convert the unstructured natural language outputs generated by the LVLMs into standardized quantitative data, the pipeline integrates the NuExtract-2.0-8B model. Because LVLMs frequently produce verbose chain-of-thought text, traditional programmatic parsing is highly prone to failure. Instead, NuExtract is utilized to map the raw descriptive text against a predefined template, generating a clean, strictly typed JSON object that reliably isolates the number of structures fully inside, partially inside, and the total impacted count. 

To mitigate stochastic generation variance and occasional model hallucinations, the framework executes a consensus protocol. For each target location, the LVLM (running on the vLLM inference engine at a temperature of $\tau = 0.3$) is queried $N = 5$ times independently. The final structural count ($Y^*$) is defined as the median value of these trials, $Y^* = \text{median}(\{y_1, y_2, y_3, y_4, y_5\})$. This statistical filter automatically eliminates anomalous outliers and spurious model behaviors.

\subsubsection{Ground Truth Annotation and Consensus Protocol}
As mentioned previously, location and pre-strike images were collected from the ArcGIS StoryMaps and Liveuamap platforms. For each benchmark location, the target radius was first computed using the blast-radius estimation procedure described in Sec.~\ref{sssec:radius_calculation}, defining the explicit spatial boundaries for the structural counting task. Human annotators then established the reference building counts within these zones using a collaborative panel of five independent researchers. 

The panel utilized a multi-view verification framework, utilizing the overhead nadir satellite imagery with ground-level perspective data. For each geographic scene, annotators cross-referenced the nadir footprint boundaries with street-level imagery obtained from both Google Street View and Mapillary, depending on localized platform availability. This dual-perspective approach allowed the panel to verify structural boundaries, resolve overhead occlusion artifacts, and accurately categorize fully versus partially impacted structures within the computed radius.

To maximize reliability and mitigate individual visual or cognitive bias, the panel evaluated the target locations collectively rather than in isolation. When discrepancies or boundary ambiguities arose regarding whether a building layout fell within the calculated spatial threshold, the panel engaged in a structured, deliberative discussion. These analytical debates involved a collaborative re-examination of the multi-view imagery streams—tracking physical indicators such as roof-line shadows, architectural facades, and visible structural debris. Final ground truth reference counts were recorded only once the group achieved unanimous consensus across all five panel members, establishing a highly reliable and verified ground truth dataset without requiring post-hoc statistical smoothing or proxy voting metrics.

\subsection{Statistical Significance Testing}
\label{subsec:statistical_significance}

To rigorously evaluate whether the integration of spatial segmentation and cross-modal relative depth maps fundamentally improves structural counting accuracy, we conduct a query-level inferential analysis. To isolate the effect of the augmentations from the inherent stochasticity of Large Vision-Language Models (LVLMs), we utilize the canonical median predictions ($N=5$) established in the evaluation pipeline. Because absolute counting errors ($\text{AE} = |N_{\text{pred}} - N_{\text{true}}|$) are bounded by zero, strictly non-normal, and heavily right-skewed in dense urban topographies, standard parametric paired $t$-tests are mathematically invalid. We therefore apply a non-parametric, paired, one-sided Wilcoxon signed-rank test across identical scene queries. Our directional alternative hypothesis ($H_1$) asserts that progressive contextual augmentation significantly reduces scene-level absolute counting errors ($\text{AE}_{\text{baseline}} > \text{AE}_{\text{treatment}}$).

\FloatBarrier 
\backmatter

\bmhead{Acknowledgements}
We would like to thank Nahian and Assaduzzan from East Delta University for leading the annotation Team. We would also like to thank Dr. Ubydul Haque from Rutgers University for initiating this project.

\bibliography{sn-bibliography}
\section*{Declarations}

\begin{itemize}
\item \textbf{Funding:} No funding was received for conducting this study.
\item \textbf{Conflict of interest/Competing interests:} The authors declare no conflicts of interest or competing financial interests.
\item \textbf{Ethics approval and consent to participate:} Not applicable.
\item \textbf{Data availability:} The data generated and analyzed during this study are publicly available on Figshare: \url{https://doi.org/10.6084/m9.figshare.33121100}.
\item \textbf{Code availability:} The repository containing the source code for this analysis is available on GitHub: \url{https://github.com/bojack-horseman91/cost_of_war}.
\item \textbf{Author contribution:}

\textbf{Saleh Sakib Ahmed}: Conceptualization, Methodology, Software, Data Curation, Formal Analysis, Visualization, and Writing -- Original Draft.

\textbf{M. Sohel Rahman}: Supervision, Formal Analysis, Writing -- Review \& Editing.
\end{itemize}
\noindent

\bigskip

\fontfamily{cmr}\selectfont 
\setcounter{section}{0}
\setcounter{figure}{0}
\setcounter{table}{0}

\renewcommand{\thesection}{S\arabic{section}}
\renewcommand{\thefigure}{S\arabic{figure}}
\renewcommand{\thetable}{S\arabic{table}}

\begin{table}[htbp]
\centering
\small
\caption{Country-Wise Aggregated Structural Exposure Report across ArcGIS and LiveUAMap Datasets.}
\label{tab:country_damage}
\begin{tabular}{lrr}
\toprule
\textbf{Country / Region} & \textbf{ArcGIS Structural Count} & \textbf{LiveUAMap Structural Count} \\ 
\midrule
Bahrain               & 43    & 45    \\
Iran                  & 2,647 & 1,762 \\
Iraq                  & 9     & 4     \\
Israel / Palestine    & 954   & 240   \\
Jordan                & 0     & 0     \\
Kuwait                & 30    & 21    \\
Qatar                 & 0     & 3     \\
Saudi Arabia          & 5     & 0     \\
United Arab Emirates  & 64    & 33    \\
Other / Unclassified  & 0     & 35    \\ 
\bottomrule
\end{tabular}
\end{table}

\begin{figure}[htbp]
    \centering
    \includegraphics[width=\textwidth]{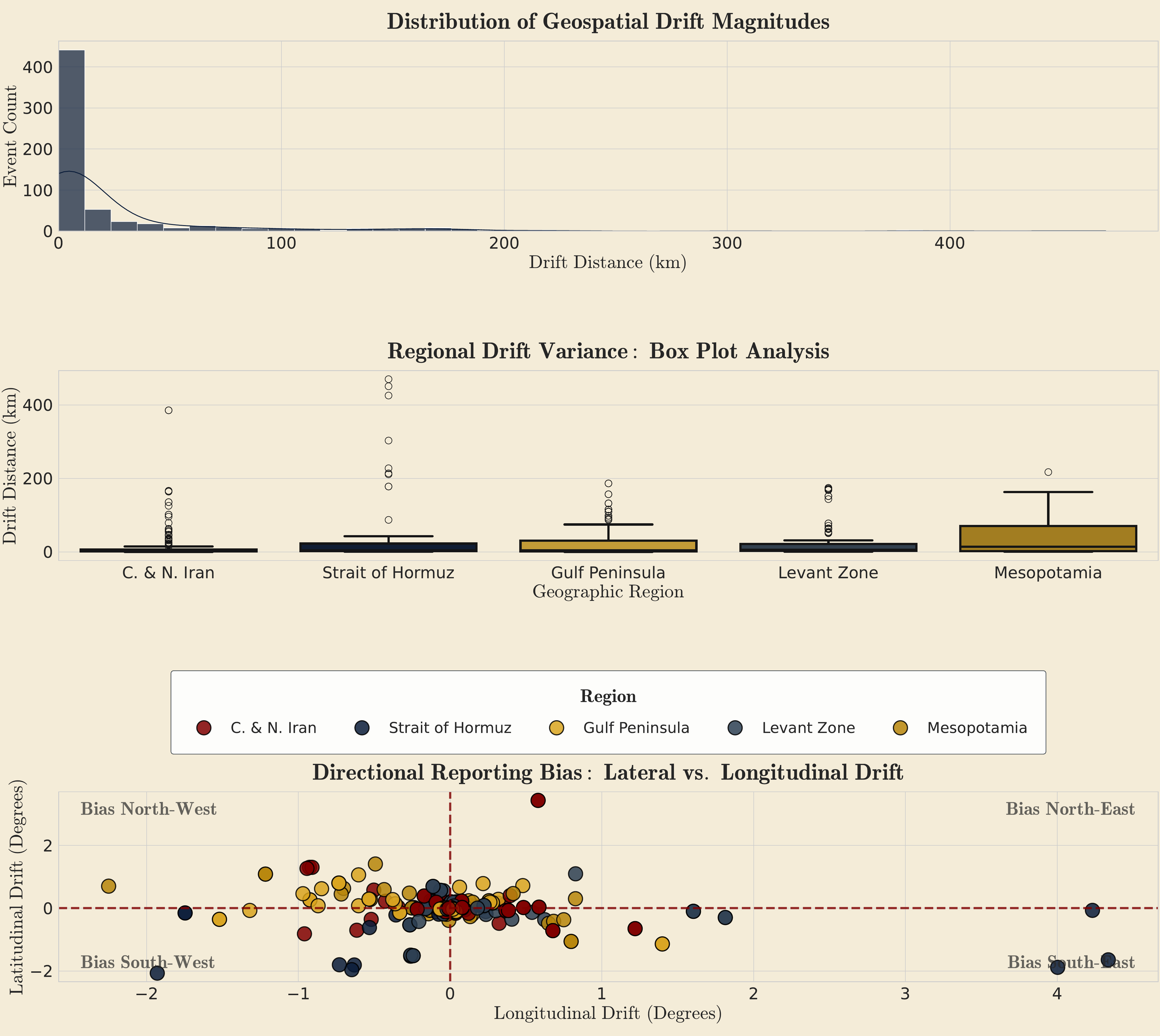}
    \caption{Quantitative analysis of geospatial drift between live reporting and baseline ArcGIS datasets. The top panel displays the overall distribution of drift magnitudes in kilometers. The middle panel highlights the regional variance of these drift distances across five distinct geographic zones. The bottom panel illustrates directional reporting bias by mapping latitudinal versus longitudinal drift, categorized by region.}
    \label{fig:geospatial_drift_dashboard}
\end{figure}
\begin{table}[htbp]
  \centering
  \caption{Comparison of segmentation performance for COMPLETE structures across datasets. The best result per column is \textbf{bold}, and the second best is \underline{underlined}. Lower is better.}
  \label{tab:supp_complete}
  \begin{tabular}{l cccc cccc}
    \toprule
    \multirow{3}{*}{Padding Strategy} & \multicolumn{4}{c}{Liveuamap Dataset} & \multicolumn{4}{c}{ArcGIS StoryMap Dataset} \\
    \cmidrule(lr){2-5} \cmidrule(lr){6-9}
    & \multicolumn{2}{c}{Google Imagery} & \multicolumn{2}{c}{ESRI Imagery} & \multicolumn{2}{c}{Google Imagery} & \multicolumn{2}{c}{ESRI Imagery} \\
    \cmidrule(lr){2-3} \cmidrule(lr){4-5} \cmidrule(lr){6-7} \cmidrule(lr){8-9}
    & MAE & MSE & MAE & MSE & MAE & MSE & MAE & MSE \\
    \midrule
    Fixed 1.0  & \textbf{0.77} & \textbf{6.49} & \textbf{0.60} & \textbf{4.82} & \underline{1.18} & \underline{7.44} & \underline{1.13} & \underline{6.50} \\
    Fixed 1.5  & 0.84 & \underline{7.24} & 0.65 & \underline{5.48} & 1.22 & 9.16 & 1.21 & 8.07 \\
    Fixed 1.85 & 0.85 & 7.36 & 0.71 & 6.00 & 1.26 & 9.80 & 1.25 & 9.05 \\
    Fixed 2.0  & 0.85 & 7.45 & 0.73 & 6.26 & 1.27 & 9.92 & 1.25 & 9.04 \\
    Fixed 2.2  & 0.86 & 7.53 & 0.74 & 6.36 & 1.25 & 9.97 & 1.26 & 9.34 \\
    Fixed 2.5  & 0.88 & 7.75 & 0.78 & 6.80 & 1.29 & 10.17 & 1.27 & 9.72 \\
    \midrule
    Adaptive   & \underline{0.81} & 7.53 & \underline{0.64} & 5.78 & \textbf{1.10} & \textbf{7.01} & \textbf{1.12} & \textbf{6.38} \\
    \bottomrule
  \end{tabular}
  
\end{table}
\begin{table}[htbp]
  \centering
  \caption{Comparison of segmentation performance for PARTIAL structures across datasets. The best result per column is \textbf{bold}, and the second best is \underline{underlined}. Lower is better.}
  \label{tab:supp_partial}
  \begin{tabular}{l cccc cccc}
    \toprule
    \multirow{3}{*}{Padding Strategy} & \multicolumn{4}{c}{Liveuamap Dataset} & \multicolumn{4}{c}{ArcGIS StoryMap Dataset} \\
    \cmidrule(lr){2-5} \cmidrule(lr){6-9}
    & \multicolumn{2}{c}{Google Imagery} & \multicolumn{2}{c}{ESRI Imagery} & \multicolumn{2}{c}{Google Imagery} & \multicolumn{2}{c}{ESRI Imagery} \\
    \cmidrule(lr){2-3} \cmidrule(lr){4-5} \cmidrule(lr){6-7} \cmidrule(lr){8-9}
    & MAE & MSE & MAE & MSE & MAE & MSE & MAE & MSE \\
    \midrule
    Fixed 1.0  & 1.18 & 6.97 & 0.91 & 4.42 & 2.01 & 8.93 & 1.91 & 8.01 \\
    Fixed 1.5  & \underline{1.02} & \underline{6.52} & \underline{0.78} & \underline{4.32} & 1.61 & 7.00 & 1.64 & \underline{6.64} \\
    Fixed 1.85 & \underline{1.02} & 6.87 & 0.84 & 5.07 & 1.53 & 6.80 & \underline{1.61} & 6.80 \\
    Fixed 2.0  & \underline{1.02} & 6.90 & 0.86 & 5.38 & 1.47 & 6.37 & 1.62 & 6.81 \\
    Fixed 2.2  & 1.05 & 7.08 & 0.86 & 5.50 & \underline{1.41} & \underline{6.23} & 1.64 & 7.16 \\
    Fixed 2.5  & 1.09 & 7.35 & 0.93 & 6.16 & 1.44 & 6.65 & 1.64 & 7.17 \\
    \midrule
    Adaptive   & \textbf{0.88} & \textbf{5.39} & \textbf{0.70} & \textbf{3.55} & \textbf{1.26} & \textbf{5.07} & \textbf{1.53} & \textbf{5.99} \\
    \bottomrule
  \end{tabular}
  
\end{table}

\begin{table*}[t]
\centering
\caption{Regional Summary Profile (LiveUAMap): Paired MAE and MSE Performance and Relative Progressions vs. Baseline.}
\label{tab:liveuamap_stacked_summary}
\resizebox{\textwidth}{!}{%
\begin{tabular}{ll ccc ccc c}
\toprule
\textbf{Density Region} & \textbf{Metric} & \textbf{No Labels (Base)} & \textbf{Labels} & \textbf{Impr (Labels)} & \textbf{+Seg} & \textbf{Impr (+Seg)} & \textbf{+Depth} & \textbf{Impr (+Depth)} \\
\midrule
\multirow{2}{*}{Low (0-5)} & MAE  & 0.86  & 0.90  & -4.1\%  & \underline{0.50}  & +41.9\%  & \textbf{0.49}  & +43.3\% \\
                          & MSE  & 3.90  & 3.89  & +0.2\%  & \textbf{1.58}  & +59.3\%  & \underline{1.71}  & +56.2\% \\
\cmidrule(lr){1-9}
\multirow{2}{*}{Med-Low (6-10)} & MAE  & \underline{5.12}  & 5.83  & -13.9\%  & 5.15  & -0.7\%  & \textbf{4.92}  & +3.9\% \\
                          & MSE  & \textbf{33.75}  & 60.29  & -78.6\%  & 36.86  & -9.2\%  & \underline{34.46}  & -2.1\% \\
\cmidrule(lr){1-9}
\multirow{2}{*}{Medium (11-15)} & MAE  & 8.19  & 9.55  & -16.7\%  & \underline{7.88}  & +3.8\%  & \textbf{7.65}  & +6.5\% \\
                          & MSE  & \underline{80.89}  & 102.79  & -27.1\%  & 82.68  & -2.2\%  & \textbf{77.78}  & +3.8\% \\
\cmidrule(lr){1-9}
\multirow{2}{*}{High (16-25)} & MAE  & 13.54  & 14.69  & -8.5\%  & \underline{13.02}  & +3.8\%  & \textbf{12.24}  & +9.6\% \\
                          & MSE  & 227.59  & 261.82  & -15.0\%  & \underline{218.91}  & +3.8\%  & \textbf{203.36}  & +10.6\% \\
\cmidrule(lr){1-9}
\multirow{2}{*}{Extreme (26+)} & MAE  & 32.38  & \underline{25.08}  & +22.5\%  & 31.70  & +2.1\%  & \textbf{24.00}  & +25.9\% \\
                          & MSE  & 1830.98  & \underline{792.08}  & +56.7\%  & 1171.62  & +36.0\%  & \textbf{708.08}  & +61.3\% \\
\bottomrule
\end{tabular}
}
\end{table*}

\begin{table*}[t]
\centering
\small
\caption{Regional Summary Profile (ArcGIS): Paired MAE and MSE Performance and Relative Progressions vs. Baseline.}
\label{tab:arcgis_stacked_summary}
\resizebox{\textwidth}{!}{%
\begin{tabular}{ll ccc cc}
\toprule
\textbf{Density Region} & \textbf{Metric} & \textbf{No Lbls} & \textbf{+Seg} & \textbf{Impr (+Seg)} & \textbf{+Depth} & \textbf{Impr (+Depth)} \\
\midrule
\multirow{2}{*}{Low (0-5)} & MAE  & 1.54  & \underline{1.32}  & +14.4\%  & \textbf{1.19}  & +22.8\% \\
                          & MSE  & 5.43  & \underline{3.93}  & +27.6\%  & \textbf{3.55}  & +34.7\% \\
\cmidrule(lr){1-7}
\multirow{2}{*}{Med-Low (6-10)} & MAE  & 5.33  & \underline{4.86}  & +8.8\%  & \textbf{3.42}  & +35.9\% \\
                          & MSE  & 34.92  & \underline{31.18}  & +10.7\%  & \textbf{18.50}  & +47.0\% \\
\cmidrule(lr){1-7}
\multirow{2}{*}{Medium (11-15)} & MAE  & 9.59  & \underline{7.58}  & +21.0\%  & \textbf{6.36}  & +33.7\% \\
                          & MSE  & 104.12  & \underline{77.51}  & +25.6\%  & \textbf{58.81}  & +43.5\% \\
\cmidrule(lr){1-7}
\multirow{2}{*}{High (16-25)} & MAE  & 15.48  & \underline{14.16}  & +8.5\%  & \textbf{12.32}  & +20.4\% \\
                          & MSE  & 274.30  & \underline{251.69}  & +8.2\%  & \textbf{196.56}  & +28.3\% \\
\cmidrule(lr){1-7}
\multirow{2}{*}{Extreme (26+)} & MAE  & 26.14  & \underline{20.29}  & +22.4\%  & \textbf{19.52}  & +25.3\% \\
                          & MSE  & 786.62  & \underline{507.71}  & +35.5\%  & \textbf{452.10}  & +42.5\% \\
\bottomrule
\end{tabular}
}
\end{table*}
\begin{table*}[t]
\centering
\small
\caption{Detailed Regional Structural Matrix Breakdown (LiveUAMap) — MAE Performance across Nested Configurations.}
\label{tab:liveuamap_detailed_mae}
\resizebox{\textwidth}{!}{%
\begin{tabular}{ll ccccc}
\toprule
\textbf{Model Name} & \textbf{Configuration} & \textbf{Low (0--5)} & \textbf{Med-Low (6--10)} & \textbf{Medium (11--15)} & \textbf{High (16--25)} & \textbf{Extreme (26+)} \\
\midrule
\multicolumn{2}{l}{\textbf{Reference (Google)}} & 0.30 & 3.97 & 6.44 & 13.86 & 30.60 \\
\multicolumn{2}{l}{\textbf{Reference (ESRI)}} & 0.31 & 2.60 & 4.97 & 7.45 & 24.80 \\
\midrule
\multirow{4}{*}{Qwen 3.6 (35B)}
     & Labels Only      & 0.39 & 5.17 & 7.56 & \underline{11.23} & 21.90 \\[0.5ex]
     & No Labels (Base) & 0.41 & \textbf{4.38} & 7.29 & \textbf{8.82} & \textbf{14.00} \\[0.5ex]
     & +Segmentation    & \underline{0.39} & 4.85 & \underline{6.56} & 12.18 & 28.40 \\[0.5ex]
     & +Rel Depth       & \textbf{0.33} (+19.2\%) & \underline{4.77} (-8.8\%) & \textbf{6.53} (+10.5\%) & 11.64 (-32.0\%) & \underline{20.40} (-45.7\%) \\[0.5ex]
\midrule
\multirow{4}{*}{Gemma 31B}
     & Labels Only      & 0.63 & 7.09 & 11.38 & 16.59 & \underline{27.80} \\[0.5ex]
     & No Labels (Base) & 0.53 & 5.03 & 6.71 & 14.55 & 28.40 \\[0.5ex]
     & +Segmentation    & \underline{0.50} & \underline{3.54} & \underline{5.53} & \underline{6.59} & 30.20 \\[0.5ex]
     & +Rel Depth       & \textbf{0.41} (+23.0\%) & \textbf{2.78} (+44.6\%) & \textbf{4.12} (+38.6\%) & \textbf{5.32} (+63.4\%) & \textbf{19.00} (+33.1\%) \\[0.5ex]
\midrule
\multirow{4}{*}{Cosmos 32B}
     & Labels Only      & \underline{0.53} & 6.49 & 10.71 & 19.09 & \underline{31.20} \\[0.5ex]
     & No Labels (Base) & \textbf{0.46} & \textbf{5.31} & \textbf{7.76} & \textbf{14.27} & 64.30 \\[0.5ex]
     & +Segmentation    & 0.55 & \underline{6.46} & \underline{9.97} & 17.55 & 42.00 \\[0.5ex]
     & +Rel Depth       & 0.62 (-33.4\%) & 6.92 (-30.4\%) & 10.44 (-34.5\%) & \underline{17.41} (-22.0\%) & \textbf{26.20} (+59.3\%) \\[0.5ex]
\midrule
\multirow{4}{*}{Claude (Qwen Dist)}
     & Labels Only      & 2.36 & \underline{2.91} & 5.97 & \textbf{7.36} & \textbf{13.30} \\[0.5ex]
     & No Labels (Base) & 2.32 & 3.59 & 7.18 & 10.95 & 24.00 \\[0.5ex]
     & +Segmentation    & \textbf{0.46} & 3.22 & \underline{5.15} & 9.86 & 26.90 \\[0.5ex]
     & +Rel Depth       & \underline{0.51} (+78.1\%) & \textbf{2.46} (+31.4\%) & \textbf{5.06} (+29.5\%) & \underline{7.59} (+30.7\%) & \underline{23.20} (+3.3\%) \\[0.5ex]
\midrule
\multirow{4}{*}{GLM 4.6V}
     & Labels Only      & \underline{0.58} & \underline{7.49} & 12.15 & 19.18 & \underline{31.20} \\[0.5ex]
     & No Labels (Base) & \underline{0.58} & \textbf{7.27} & \textbf{12.00} & \underline{19.09} & \underline{31.20} \\[0.5ex]
     & +Segmentation    & 0.59 & 7.69 & 12.18 & \textbf{18.91} & \textbf{31.00} \\[0.5ex]
     & +Rel Depth       & \textbf{0.57} (+0.5\%) & 7.64 (-5.1\%) & \underline{12.12} (-1.0\%) & 19.23 (-0.7\%) & \underline{31.20} (+0.0\%) \\[0.5ex]
\midrule
\bottomrule
\end{tabular}
}
\end{table*}

\begin{table*}[t]
\centering
\small
\caption{Detailed Regional Structural Matrix Breakdown (LiveUAMap) — MSE Performance across Nested Configurations.}
\label{tab:liveuamap_detailed_mse}
\resizebox{\textwidth}{!}{%
\begin{tabular}{ll ccccc}
\toprule
\textbf{Model Name} & \textbf{Configuration} & \textbf{Low (0--5)} & \textbf{Med-Low (6--10)} & \textbf{Medium (11--15)} & \textbf{High (16--25)} & \textbf{Extreme (26+)} \\
\midrule
\multicolumn{2}{l}{\textbf{Reference (Google)}} & 0.92 & 23.77 & 50.44 & 222.95 & 972.60 \\
\multicolumn{2}{l}{\textbf{Reference (ESRI)}} & 0.95 & 12.96 & 46.97 & 78.73 & 751.80 \\
\midrule
\multirow{4}{*}{Qwen 3.6 (35B)}
     & Labels Only      & \underline{0.92} & 128.29 & 67.15 & \underline{163.95} & 858.30 \\[0.5ex]
     & No Labels (Base) & \textbf{0.89} & \textbf{23.95} & \underline{58.71} & \textbf{88.27} & \textbf{280.80} \\[0.5ex]
     & +Segmentation    & 1.18 & 33.51 & \textbf{56.68} & 202.73 & 886.00 \\[0.5ex]
     & +Rel Depth       & 1.08 (-21.8\%) & \underline{31.69} (-32.3\%) & 62.76 (-6.9\%) & 195.82 (-121.8\%) & \underline{565.20} (-101.3\%) \\[0.5ex]
\midrule
\multirow{4}{*}{Gemma 31B}
     & Labels Only      & 2.05 & 53.96 & 135.38 & 319.14 & \underline{857.40} \\[0.5ex]
     & No Labels (Base) & 1.50 & 35.77 & 64.12 & 258.00 & 874.80 \\[0.5ex]
     & +Segmentation    & \underline{1.48} & \underline{21.36} & \underline{47.71} & \underline{77.41} & 959.60 \\[0.5ex]
     & +Rel Depth       & \textbf{1.20} (+20.4\%) & \textbf{13.19} (+63.1\%) & \textbf{22.94} (+64.2\%) & \textbf{44.50} (+82.8\%) & \textbf{513.20} (+41.3\%) \\[0.5ex]
\midrule
\multirow{4}{*}{Cosmos 32B}
     & Labels Only      & \underline{1.53} & \underline{46.08} & 119.82 & 370.45 & \underline{1003.60} \\[0.5ex]
     & No Labels (Base) & \textbf{1.24} & \textbf{33.56} & \textbf{72.76} & \textbf{257.45} & 6292.70 \\[0.5ex]
     & +Segmentation    & 1.74 & 49.08 & \underline{117.26} & \underline{322.45} & 2191.60 \\[0.5ex]
     & +Rel Depth       & 2.06 (-67.0\%) & 54.51 (-62.4\%) & 119.09 (-63.7\%) & 327.23 (-27.1\%) & \textbf{815.60} (+87.0\%) \\[0.5ex]
\midrule
\multirow{4}{*}{Claude (Qwen Dist)}
     & Labels Only      & 13.17 & \underline{14.24} & 41.85 & \underline{82.18} & \textbf{237.50} \\[0.5ex]
     & No Labels (Base) & 14.24 & 19.72 & 62.53 & 163.77 & 703.00 \\[0.5ex]
     & +Segmentation    & \textbf{1.59} & 19.04 & \underline{41.21} & 130.05 & 828.30 \\[0.5ex]
     & +Rel Depth       & \underline{2.32} (+83.7\%) & \textbf{12.13} (+38.5\%) & \textbf{34.76} (+44.4\%) & \textbf{74.32} (+54.6\%) & \underline{642.80} (+8.6\%) \\[0.5ex]
\midrule
\multirow{4}{*}{GLM 4.6V}
     & Labels Only      & \underline{1.78} & \underline{58.87} & 149.74 & 373.36 & \underline{1003.60} \\[0.5ex]
     & No Labels (Base) & \textbf{1.62} & \textbf{55.76} & \textbf{146.35} & \underline{370.45} & \underline{1003.60} \\[0.5ex]
     & +Segmentation    & 1.94 & 61.31 & 150.53 & \textbf{361.91} & \textbf{992.60} \\[0.5ex]
     & +Rel Depth       & 1.87 (-15.0\%) & 60.77 (-9.0\%) & \underline{149.35} (-2.0\%) & 374.95 (-1.2\%) & \underline{1003.60} (+0.0\%) \\[0.5ex]
\midrule
\bottomrule
\end{tabular}
}
\end{table*}

\begin{table*}[t]
\centering
\small
\caption{Detailed Regional Structural Matrix Breakdown (ArcGIS) — MAE Performance across Nested Configurations.}
\label{tab:arcgis_detailed_mae}
\resizebox{\textwidth}{!}{%
\begin{tabular}{ll ccccc}
\toprule
\textbf{Model Name} & \textbf{Configuration} & \textbf{Low (0--5)} & \textbf{Med-Low (6--10)} & \textbf{Medium (11--15)} & \textbf{High (16--25)} & \textbf{Extreme (26+)} \\
\midrule
\multicolumn{2}{l}{\textbf{Reference (Google)}} & 0.82 & 2.84 & 6.88 & 14.23 & 21.29 \\
\multicolumn{2}{l}{\textbf{Reference (ESRI)}} & 1.12 & 4.49 & 5.83 & 11.29 & 20.14 \\
\midrule
\multirow{3}{*}{Qwen 3.6 (35B)}
     & Only (Base)      & \textbf{1.11} & \underline{5.13} & \underline{9.43} & 16.00 & 26.86 \\[0.5ex]
     & +Seg          & 1.31 & 5.35 & \underline{9.43} & \underline{15.77} & \underline{22.43} \\[0.5ex]
     & +Rel Depth       & \underline{1.19} (-7.4\%) & \textbf{4.62} (+9.9\%) & \textbf{8.35} (+11.4\%) & \textbf{14.00} (+12.5\%) & \textbf{20.14} (+25.0\%) \\[0.5ex]
\midrule
\multirow{3}{*}{Gemma 31B}
     & Only (Base)      & 1.89 & 7.45 & 12.38 & 19.90 & 33.14 \\[0.5ex]
     & +Seg          & \underline{1.37} & \underline{4.66} & \underline{6.47} & \underline{12.42} & \underline{19.00} \\[0.5ex]
     & +Rel Depth       & \textbf{1.11} (+41.2\%) & \textbf{3.05} (+59.1\%) & \textbf{5.50} (+55.6\%) & \textbf{10.29} (+48.3\%) & \textbf{15.71} (+52.6\%) \\[0.5ex]
\midrule
\multirow{3}{*}{Claude (Qwen Dist)}
     & Only (Base)      & 1.61 & \underline{3.42} & 6.97 & \textbf{10.55} & \textbf{18.43} \\[0.5ex]
     & +Seg          & \underline{1.27} & 4.59 & \underline{6.83} & 14.29 & \underline{19.43} \\[0.5ex]
     & +Rel Depth       & \textbf{1.26} (+21.9\%) & \textbf{2.59} (+24.3\%) & \textbf{5.22} (+25.1\%) & \underline{12.68} (-20.2\%) & 22.71 (-23.3\%) \\[0.5ex]
\midrule
\bottomrule
\end{tabular}
}
\end{table*}

\begin{table*}[t]
\centering
\small
\caption{Detailed Regional Structural Matrix Breakdown (ArcGIS) — MSE Performance across Nested Configurations.}
\label{tab:arcgis_detailed_mse}
\resizebox{\textwidth}{!}{%
\begin{tabular}{ll ccccc}
\toprule
\textbf{Model Name} & \textbf{Configuration} & \textbf{Low (0--5)} & \textbf{Med-Low (6--10)} & \textbf{Medium (11--15)} & \textbf{High (16--25)} & \textbf{Extreme (26+)} \\
\midrule
\multicolumn{2}{l}{\textbf{Reference (Google)}} & 1.80 & 13.61 & 62.67 & 249.71 & 517.00 \\
\multicolumn{2}{l}{\textbf{Reference (ESRI)}} & 3.15 & 27.50 & 45.98 & 180.45 & 492.43 \\
\midrule
\multirow{3}{*}{Qwen 3.6 (35B)}
     & Only (Base)      & \textbf{2.82} & \underline{31.93} & \underline{98.12} & \underline{286.13} & 830.57 \\[0.5ex]
     & +Seg          & 4.08 & 35.15 & 108.58 & 293.45 & \underline{645.57} \\[0.5ex]
     & +Rel Depth       & \underline{3.50} (-24.1\%) & \textbf{29.05} (+9.0\%) & \textbf{89.40} (+8.9\%) & \textbf{249.68} (+12.7\%) & \textbf{460.43} (+44.6\%) \\[0.5ex]
\midrule
\multirow{3}{*}{Gemma 31B}
     & Only (Base)      & 6.33 & 57.35 & 155.72 & 406.16 & 1131.14 \\[0.5ex]
     & +Seg          & \underline{3.98} & \underline{29.60} & \underline{58.38} & \underline{209.06} & \underline{437.86} \\[0.5ex]
     & +Rel Depth       & \textbf{2.86} (+54.8\%) & \textbf{14.63} (+74.5\%) & \textbf{45.00} (+71.1\%) & \textbf{148.55} (+63.4\%) & \textbf{312.00} (+72.4\%) \\[0.5ex]
\midrule
\multirow{3}{*}{Claude (Qwen Dist)}
     & Only (Base)      & 7.14 & \underline{15.47} & \underline{58.52} & \textbf{130.61} & \textbf{398.14} \\[0.5ex]
     & +Seg          & \textbf{3.73} & 28.78 & 65.58 & 252.55 & \underline{439.71} \\[0.5ex]
     & +Rel Depth       & \underline{4.28} (+40.0\%) & \textbf{11.82} (+23.6\%) & \textbf{42.02} (+28.2\%) & \underline{191.45} (-46.6\%) & 583.86 (-46.6\%) \\[0.5ex]
\midrule
\bottomrule
\end{tabular}
}
\end{table*}


\begin{table*}[t]
\centering
\small
\caption{Directional Bias and Variance Profile (Liveuamap): Tracking of Structural Output Deviation Fluctuations.}
\label{tab:bias_std_analysis}
\begin{tabular}{llcccccccc}
\toprule
\textbf{Model Name} & \textbf{Config} & \multicolumn{4}{c}{\textbf{System Bias Magnitude $\downarrow$}} & \multicolumn{4}{c}{\textbf{Standard Deviation (Std) $\downarrow$}} \\
\cmidrule(lr){3-6} \cmidrule(lr){7-10}
 & & \textbf{Labels} & \textbf{No Lbls} & \textbf{+Seg} & \textbf{+Depth} & \textbf{Labels} & \textbf{No Lbls} & \textbf{+Seg} & \textbf{+Depth} \\
\midrule
Claude (Qwen Dist) & \shortstack{35B \\ \scriptsize \textit{MoE}} & $1.03$ & $\mathbf{0.62}$ & $-1.11$ & $\underline{-0.67}$ & $2.85$ & $2.61$ & $\mathbf{1.67}$ & $\underline{1.99}$ \\[0.5ex]
Qwen 3.6 Baseline & \shortstack{35B \\ \scriptsize \textit{MoE}} & $\mathbf{-1.06}$ & $\underline{-1.13}$ & $-1.63$ & $-1.36$ & $1.20$ & $\underline{0.84}$ & $\mathbf{0.68}$ & $0.86$ \\[0.5ex]
Google Gemma 4 & \shortstack{31B \\ \scriptsize \textit{Dense}} & $-2.37$ & $-1.81$ & $\underline{-1.37}$ & $\mathbf{-0.91}$ & $\underline{0.19}$ & $\mathbf{0.15}$ & $0.29$ & $0.35$ \\[0.5ex]
Nvidia Cosmos & \shortstack{32B \\ \scriptsize \textit{Dense}} & $-2.25$ & $\mathbf{-0.88}$ & $\underline{-1.98}$ & $-2.31$ & $\mathbf{0.72}$ & $1.26$ & $0.94$ & $\underline{0.77}$ \\[0.5ex]
Zhipu GLM 4.6V & \shortstack{10B \\ \scriptsize \textit{Dense}} & $\underline{-2.48}$ & $\mathbf{-2.38}$ & $-2.52$ & $-2.50$ & $\mathbf{0.25}$ & $\underline{0.28}$ & $1.32$ & $426.21$ \\[0.5ex]
\bottomrule
\end{tabular}
\end{table*}

\begin{table*}[t]
\centering
\small
\caption{Generation Length Profile (Liveuamap): Tracking Average Output Word Bounds Matrix.}
\label{tab:word_count_analysis}
\begin{tabular}{llcccc}
\toprule
\textbf{Model Name} & \textbf{Config} & \textbf{Labels Only} & \textbf{No Labels} & \textbf{+Segmentation} & \textbf{+Depth Data} \\
\midrule
Claude (Qwen Dist) & \shortstack{35B \\ \scriptsize \textit{MoE}} & $1019.90$ & $\underline{715.98}$ & $\mathbf{675.00}$ & $812.36$ \\[0.5ex]
Qwen 3.6 Baseline & \shortstack{35B \\ \scriptsize \textit{MoE}} & $2654.18$ & $3265.18$ & $\mathbf{1312.98}$ & $\underline{1572.02}$ \\[0.5ex]
Google Gemma 4 & \shortstack{31B \\ \scriptsize \textit{Dense}} & $\underline{25.95}$ & $\mathbf{21.28}$ & $39.77$ & $48.36$ \\[0.5ex]
Nvidia Cosmos & \shortstack{32B \\ \scriptsize \textit{Dense}} & $337.72$ & $687.50$ & $\underline{130.31}$ & $\mathbf{106.92}$ \\[0.5ex]
Zhipu GLM 4.6V & \shortstack{10B \\ \scriptsize \textit{Dense}} & $1524.97$ & $\mathbf{1406.93}$ & $\underline{1430.65}$ & $1764.56$ \\[0.5ex]
\bottomrule
\end{tabular}
\end{table*}

\begin{table*}[t]
\centering
\small
\caption{Directional Bias and Variance Profile (ArcGIS StoryMap): Tracking of Structural Output Deviation Fluctuations.}
\label{tab:arcgis_bias_std_analysis}
\begin{tabular}{llcccccc}
\toprule
\textbf{Model Name} & \textbf{Config} & \multicolumn{3}{c}{\textbf{System Bias Magnitude $\downarrow$}} & \multicolumn{3}{c}{\textbf{Standard Deviation (Std) $\downarrow$}} \\
\cmidrule(lr){3-5} \cmidrule(lr){6-8}
 & & \textbf{No Lbls} & \textbf{+Seg} & \textbf{+Depth} & \textbf{No Lbls} & \textbf{+Seg} & \textbf{+Depth} \\
\midrule
Claude (Qwen Dist) & \shortstack{35B \\ \scriptsize \textit{MoE}} & $\mathbf{-0.88}$ & $-2.10$ & $\underline{-1.08}$ & $\underline{1.83}$ & $\mathbf{1.44}$ & $2.71$ \\[0.5ex]
Qwen 3.6 Baseline & \shortstack{35B \\ \scriptsize \textit{MoE}} & $\underline{-2.65}$ & $-2.93$ & $\mathbf{-2.47}$ & $\underline{1.11}$ & $\mathbf{0.91}$ & $1.29$ \\[0.5ex]
Google Gemma 4 & \shortstack{31B \\ \scriptsize \textit{Dense}} & $-4.19$ & $\underline{-2.51}$ & $\mathbf{-1.73}$ & $\mathbf{0.03}$ & $\underline{0.38}$ & $0.52$ \\[0.5ex]
\bottomrule
\end{tabular}
\end{table*}

\begin{table*}[t]
\centering
\small
\caption{Generation Length Profile (ArcGIS StoryMap): Tracking Average Output Word Bounds Matrix.}
\label{tab:arcgis_word_count_analysis}
\begin{tabular}{llccc}
\toprule
\textbf{Model Name} & \textbf{Config} & \textbf{No Labels} & \textbf{+Segmentation} & \textbf{+Depth Data} \\
\midrule
Claude (Qwen Dist) & \shortstack{35B \\ \scriptsize \textit{MoE}} & $\mathbf{736.21}$ & $\underline{973.25}$ & $1430.62$ \\[0.5ex]
Qwen 3.6 Baseline & \shortstack{35B \\ \scriptsize \textit{MoE}} & $\mathbf{1360.30}$ & $\underline{2165.78}$ & $3102.60$ \\[0.5ex]
Google Gemma 4 & \shortstack{31B \\ \scriptsize \textit{Dense}} & $\mathbf{13.41}$ & $\underline{58.45}$ & $90.06$ \\[0.5ex]
\bottomrule
\end{tabular}
\end{table*}

\begin{figure}
    \centering
    \includegraphics[width=\linewidth]{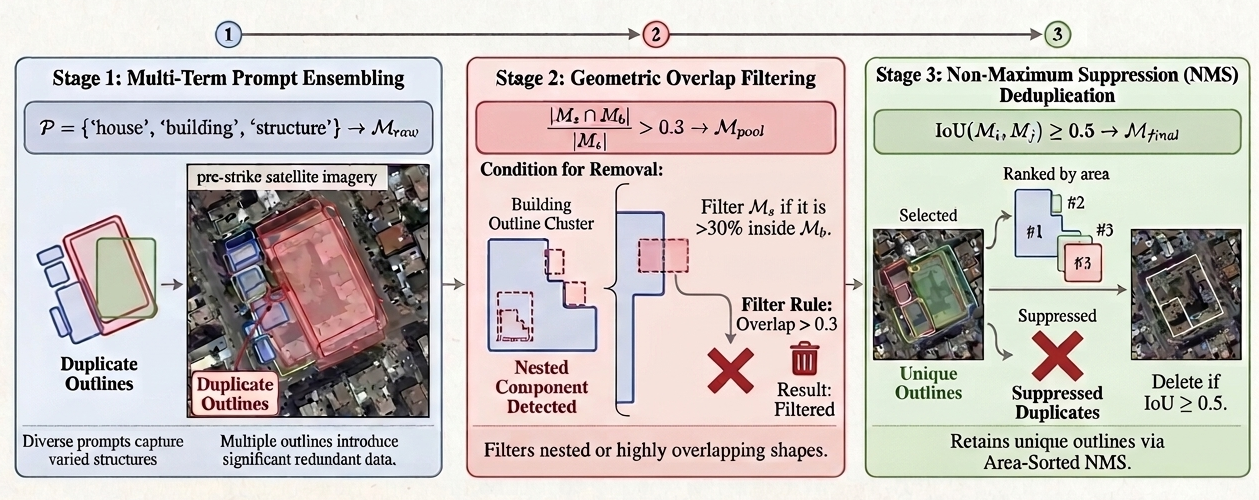}
    \caption{The mathematical pipeline for spatial outline normalization: (1) Stage 1 clusters raw, redundant candidate footprints ($\mathcal{M}_{\text{raw}}$) generated from multi-term prompt ensembling; (2) Stage 2 executes geometric overlap filtering to discard auxiliary nested shapes ($\mathcal{M}_s$) if they are $>30\%$ inside a primary outline ($\mathcal{M}_b$); (3) Stage 3 implements area-indexed Non-Maximum Suppression (NMS) to delete duplicate outlines if $\text{IoU} \ge 0.5$, securing the final consolidated unique targets ($\mathcal{M}_{\text{final}}$).}
\label{fig:spatial_normalization}
\end{figure}
\begin{figure}
    \centering
    \includegraphics[width=\linewidth]{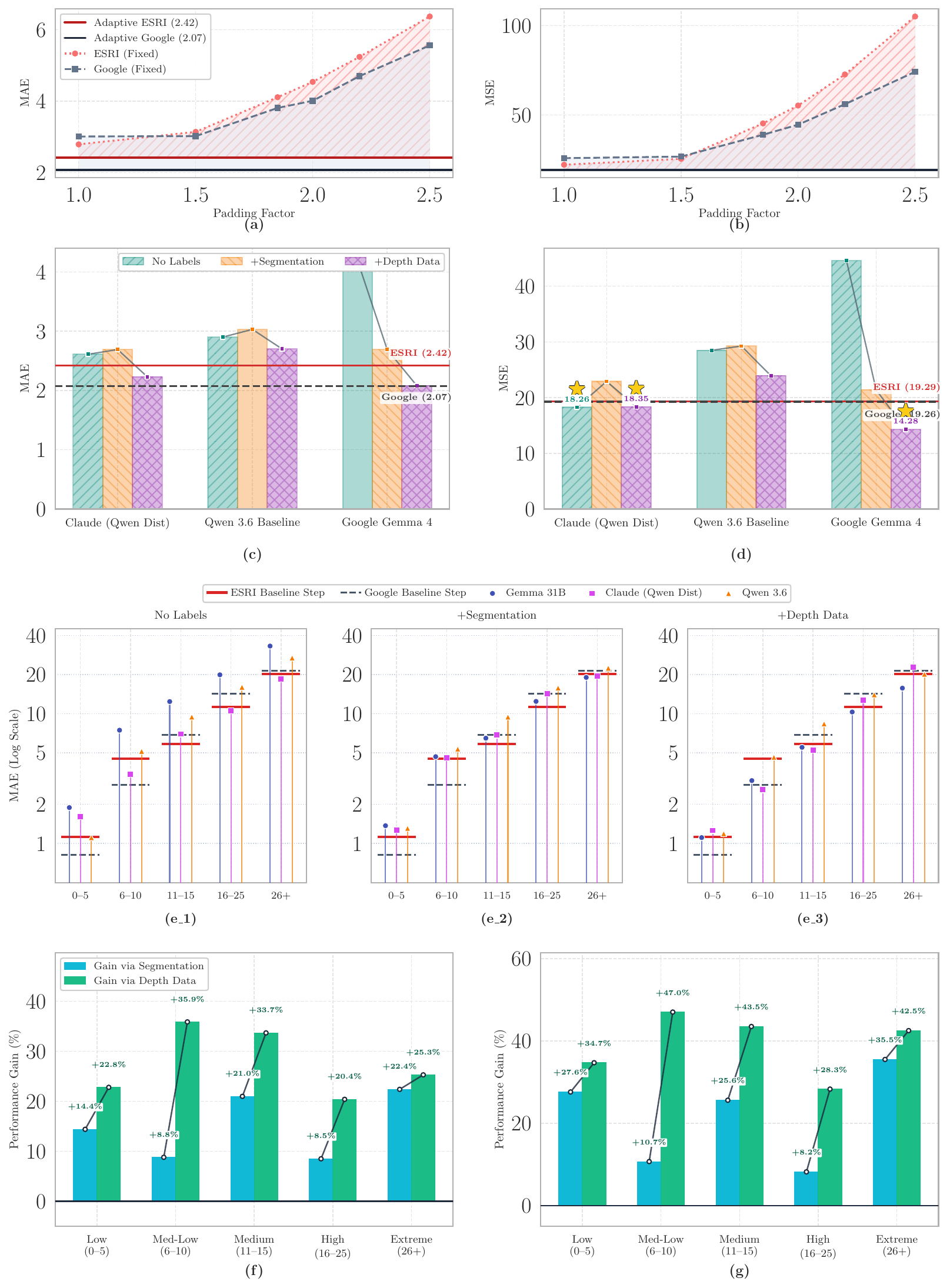}
    \caption{\textbf{Performance Evaluation and Ablation Analysis of the LVLM Architecture (ArcGIS StoryMap).} 
    \textbf{(a)~\&~(b)} Error metrics (MAE and MSE) across spatial padding factors, contrasting the adaptive strategy against fixed baselines on ArcGIS imagery. 
    \textbf{(c)~\&~(d)} Global macro-level benchmarking of three vision-language model configurations across three prompt modalities against deterministic references. 
    \textbf{(e1--e3)} Localized building-level MAE breakdown across five distinct building density tiers under progressive feature configurations. 
    \textbf{(f)~\&~(g)} Relative spatial performance gains (\%) demonstrating the marginal structural impact of downstream segmentation masks and relative depth cues across regional density cohorts.}
    \label{fig:arcgis_results_llm}
\end{figure}

\begin{figure}[t]
    \centering
    \includegraphics[width=\linewidth]{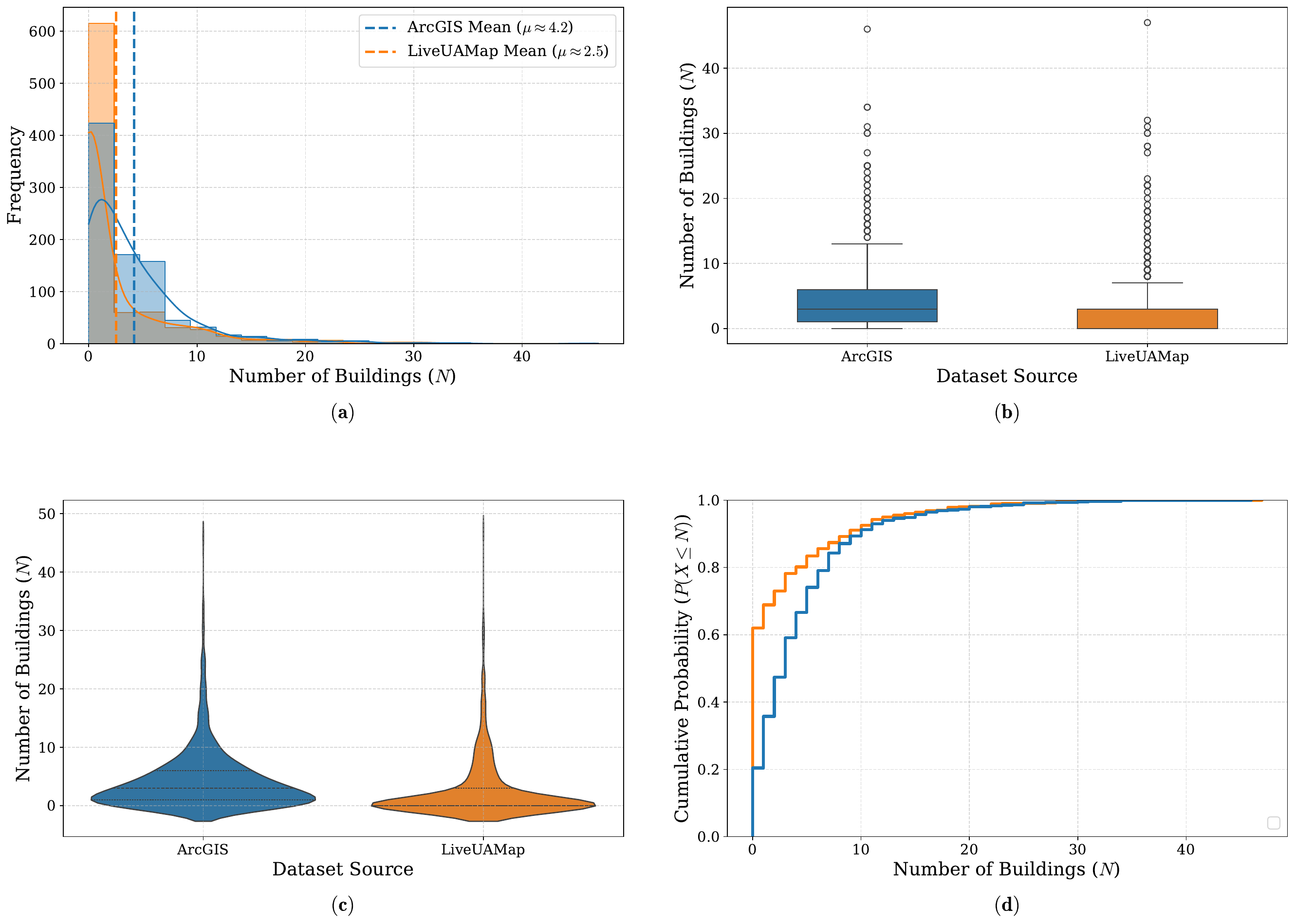}
    \caption{\textbf{Empirical distribution and structural density analysis of ground-truth datasets.} Comparison of building count distributions ($N$) derived from ArcGIS StoryMaps and LiveUAMap profiles: 
    \textbf{(a)} Empirical histogram overlaid with Kernel Density Estimation (KDE) curves, where dashed lines represent mean values ($\mu_{\text{ArcGIS}} \approx 20.3$, $\mu_{\text{LiveUAMap}} \approx 13.9$); 
    \textbf{(b)} Interquartile range (IQR) and outlier profiles mapped via boxplot distribution; 
    \textbf{(c)} Probability density shape mapping using a quartile-resolved violin plot; and 
    \textbf{(d)} Empirical Cumulative Distribution Function (ECDF) tracking cumulative probability boundaries ($P(X \leq N)$).}
    \label{fig:groundtruth_distribution}
\end{figure}
\begin{figure}
    \centering
    \includegraphics[width=\linewidth]{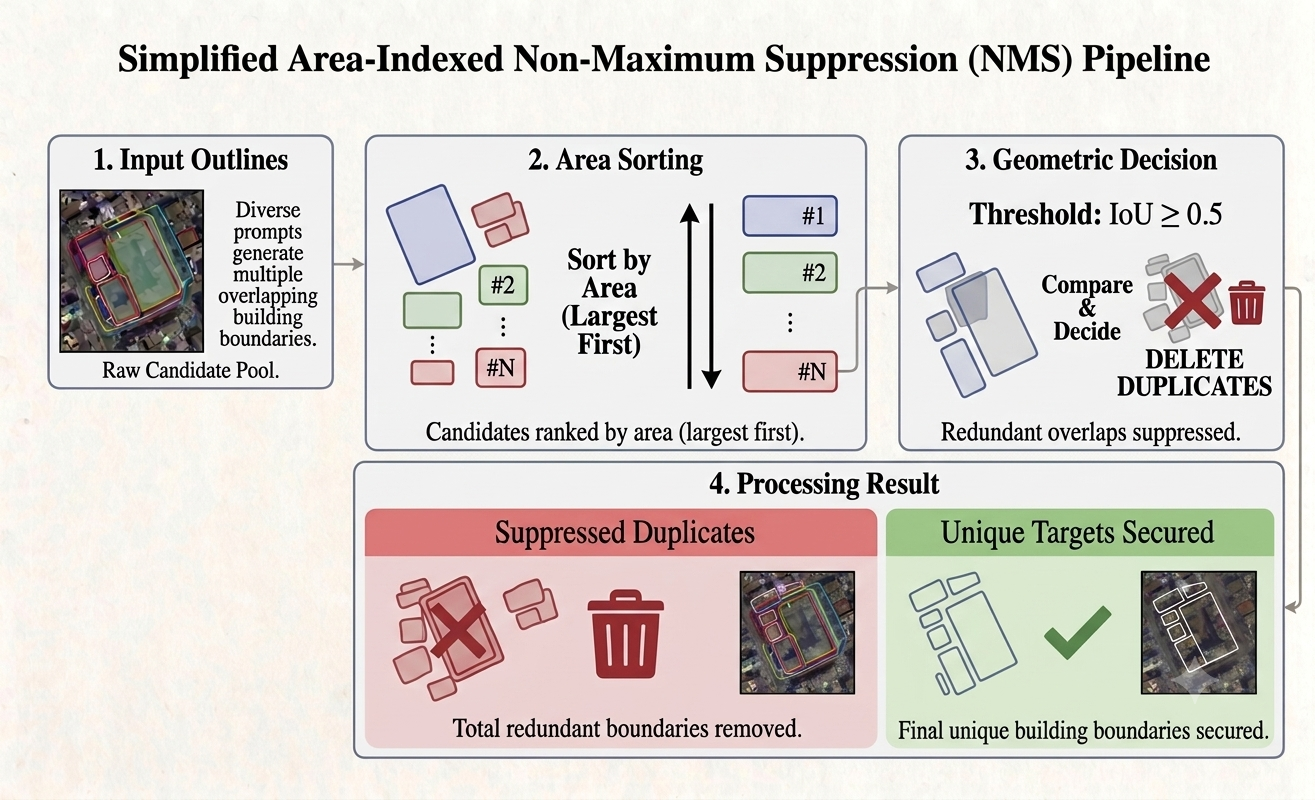}
    \caption{\textbf{Methodological pipeline of the Simplified Area-Indexed Non-Maximum Suppression (NMS) framework.} The process operates sequentially across four key phases: 
    (1) \textbf{Input Outlines}, where diverse prompts generate a raw pool of heavily overlapping building candidate boundaries; 
    (2) \textbf{Area Sorting}, which ranks candidate polygons by area in descending order from largest (\#1) to smallest (\#N); 
    (3) \textbf{Geometric Decision}, utilizing an Intersection over Union ($\text{IoU} \ge 0.5$) threshold to compare proposals and isolate duplicates; and 
    (4) \textbf{Processing Result}, yielding the final suppression of redundant overlaps and securing the definitive, unique building target boundaries.}
    \label{fig:area_indexed_nms}
\end{figure}

\begin{figure*}[!htbp]
    \centering
    \includegraphics[width=\linewidth]{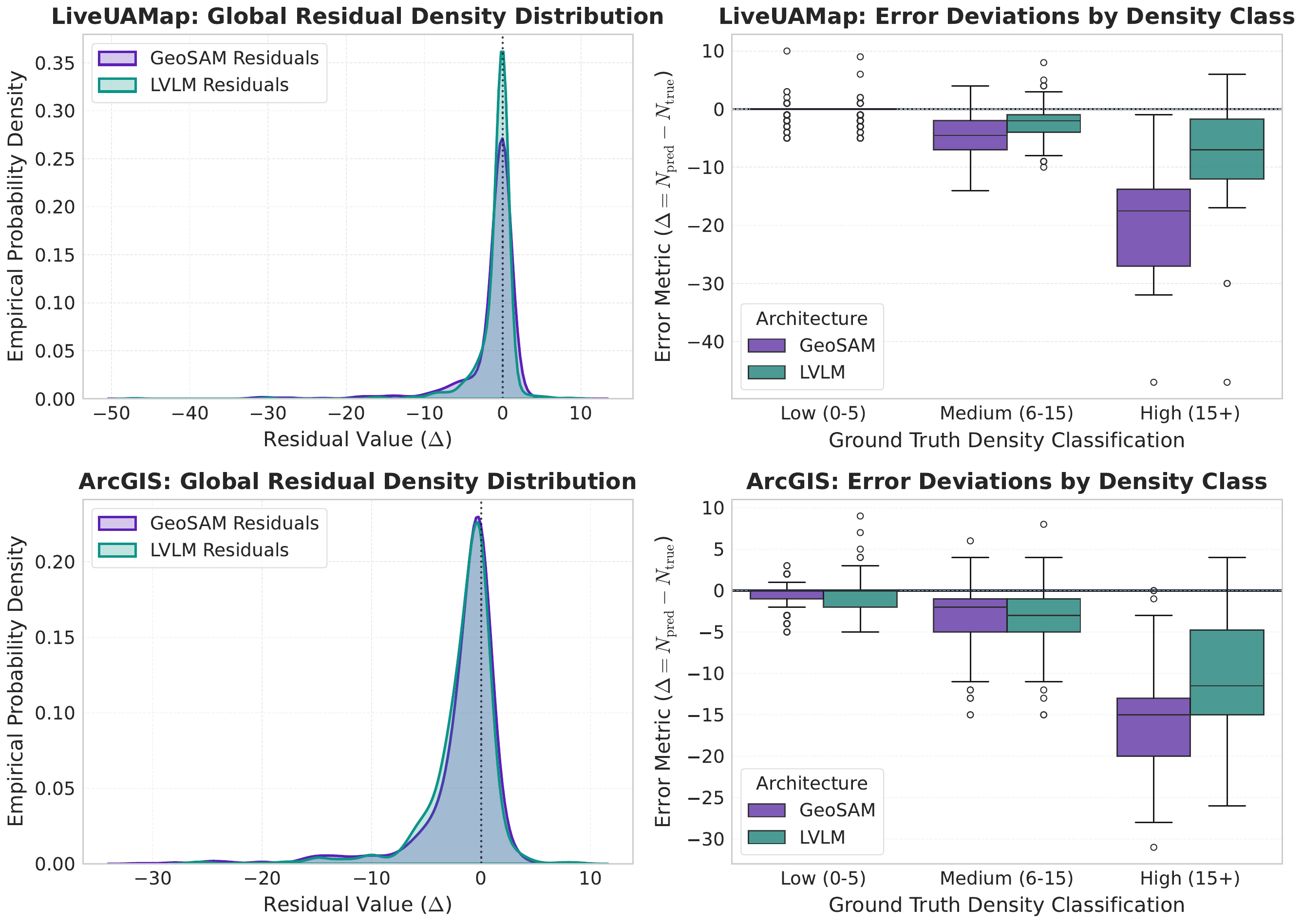}
    \caption{\textbf{Supplementary Residual Diagnostics and Density-Stratified Error Profiles (LVLM vs. GeoSAM).} 
    Global empirical probability density distributions (left) and density-stratified error boxplots (right) of prediction residuals ($\Delta = N_{\text{pred}} - N_{\text{true}}$) across the LiveUAMap and ArcGIS benchmarks. 
    \textbf{Top Row (LiveUAMap):} The LVLM framework exhibits a sharp, zero-centered residual peak, suppressing the severe negative bias (under-counting) and high error variance that plague the GeoSAM baseline as spatial congestion increases. 
    \textbf{Bottom Row (ArcGIS):} Both architectures display comparable residual dispersion across density tiers, with GeoSAM maintaining a slightly sharper zero-centered density peak in low-to-moderate density vector environments.}
    \label{fig:supplementary_diagnostics}
\end{figure*}

\section{Algorithmic Details: Size-Prioritized Greedy Non-Maximum Suppression}
\label{supp:greedy_nms}

As shown in Fig.~\ref{fig:area_indexed_nms}, to resolve spatial redundancies and handle overlapping polygon boundaries generated by correlated open-vocabulary prompts (e.g., \textit{"building"} vs. \textit{"structure"}), we implement an area-indexed Non-Maximum Suppression (NMS) routine. Unlike standard bounding-box NMS which relies on coarse axis-aligned rectangles, our approach acts directly on the boolean pixel matrices of the segmentations. 

The candidate pool $\mathcal{M}_{\text{pool}}$ is sorted in a strict descending linear sequence based on absolute pixel area:
\begin{equation}
    \text{Area}(M_i) = \sum_{p \in \mathbf{I}} M_i(p).
\end{equation}
This sorting ensures that wide-area architectural backbones are evaluated prior to localized structural subsets. During downward sequential traversal, the intersection metric between an unverified candidate mask $M_{\text{cand}}$ and the accepted ground-truth compilation set $\mathcal{M}_{\text{final}}$ is continuously tracked. The candidate mask is pruned if its maximum overlapping Intersection-over-Union ($\text{IoU}$) satisfies:
\begin{equation}
    \max_{M_{\text{acc}} \in \mathcal{M}_{\text{final}}} \left( \frac{\sum (M_{\text{cand}} \cap M_{\text{acc}})}{\sum (M_{\text{cand}} \cup M_{\text{acc}}) + \epsilon} \right) \ge 0.5,
\end{equation}
where $\epsilon = 1\times10^{-7}$ prevents division-by-zero anomalies over high-resolution rasters. Pruned matrices are discarded, while surviving independent building instances are appended to $\mathcal{M}_{\text{final}}$, finalizing the target profile.
\section{Justification of Operational Radii}
\label{sec:radius_justification}

The baseline radius is computed using the Hopkinson--Cranz cube-root scaling law,

\[
R_{\text{base}} = ZW^{1/3},
\]

where \(W\) is the TNT-equivalent explosive mass (kg) and \(Z\) is the scaled distance \((\mathrm{m/kg^{1/3}})\). The scaled distance normalizes the standoff distance with respect to the explosive mass, enabling consistent comparison of blast effects across different charge sizes.

Blast loading is commonly divided into contact, near-field, mid-field, and far-field regimes based on the scaled distance~\cite{hilding2016methods}. In the far-field, the influence of the explosive charge geometry on the blast wave becomes much smaller, and the blast loading is primarily characterized by the scaled propagation distance, making the Hopkinson--Cranz scaling law suitable for defining a consistent reference distance across different explosive masses. Hilding~\cite{hilding2016methods} identifies far-field conditions as \(Z > 4\,\mathrm{m/kg^{1/3}}\), a classification that is also adopted in reinforced concrete blast analyses by Zhou et al.~\cite{zhou2026damage}.

Accordingly, this work adopts a representative scaled distance of \(Z = 5\,\mathrm{m/kg^{1/3}}\), which lies comfortably within the far-field regime. Selecting a value slightly above the far-field threshold provides a simple and consistent engineering reference while avoiding dependence on the exact transition between blast regimes. This value is not intended to represent a universal damage radius. Instead, it provides a physically motivated reference distance for scaling the spatial extent around the reported impact location where strike-related damage is expected to be observable in remote sensing imagery. \Cref{tab:munition_payload_scaling} summarizes the resulting baseline operational radii across the representative threat classes.

\begin{table}[h]
\centering
\small
\caption{\textbf{Weapon Class Bounding Radius Scaling Parameters.} Baseline operational radii ($R_{\mathrm{base}}$) computed using the Hopkinson--Cranz cube-root scaling relationship with representative TNT-equivalent explosive masses ($W$) and a calibrated scaled-distance coefficient of $Z = 5.0\,\mathrm{m/kg}^{1/3}$.}
\label{tab:munition_payload_scaling}
\begin{tabular}{lcc}
\toprule
\textbf{Weapon Class} &
\textbf{Representative TNT-Equivalent Explosive Mass $W$ (kg)} &
\textbf{Baseline Radius $R_{\mathrm{base}}$ (m)}\\
\midrule
Small Kamikaze     & 1     & 5  \\
Tactical Rocket    & 14    & 12 \\
Loitering Munition & 27    & 15 \\
Heavy Artillery    & 64    & 20 \\
Cruise Missile     & 343   & 35 \\
Air-to-Surface     & 512   & 40 \\
Ballistic Missile  & 1,728 & 60 \\
\bottomrule
\end{tabular}
\end{table}

The representative munition classes, corresponding TNT-equivalent explosive-mass estimates, and supporting literature used to calibrate the operational spatial priors are summarized below. The selected explosive masses are intended as representative class-level values for spatial-prior calibration rather than exact specifications of individual munitions.

\begin{itemize}

\item \textbf{Small Kamikaze:}
Based on First-Person View (FPV) attack drones, which typically carry approximately $1\,\mathrm{kg}$ of explosive payload~\cite{khs7_payload}:
\begin{equation*}
R \approx 5.0 \cdot (1)^{1/3}=5\,\mathrm{m}.
\end{equation*}

\item \textbf{Tactical Rocket:}
Based on tactical battlefield rockets such as the BM-21 Grad 122\,mm 9M22U high-explosive fragmentation rocket, which contains approximately $6.4\,\mathrm{kg}$ of explosive filler within an $18.4\,\mathrm{kg}$ warhead~\cite{grad_9m22u}. To account for the diversity of explosive fills reported across operational 122\,mm rocket systems rather than the BM-21 9M22U alone, a representative TNT-equivalent explosive mass of approximately $14\,\mathrm{kg}$ is adopted:
\begin{equation*}
R \approx 5.0 \cdot (14)^{1/3}\approx12\,\mathrm{m}.
\end{equation*}

\item \textbf{Loitering Munition:}
Based on the Shahed-131 and Shahed-136 loitering munition family. Reported warhead explosive masses are approximately $10$--$20\,\mathrm{kg}$ for the Shahed-131 and $20$--$50\,\mathrm{kg}$ for the larger Shahed-136~\cite{shahed136_csis,shahed136_armyrec}. A representative intermediate TNT-equivalent explosive mass of approximately $27\,\mathrm{kg}$ is adopted:
\begin{equation*}
R \approx 5.0 \cdot (27)^{1/3}=15\,\mathrm{m}.
\end{equation*}

\item \textbf{Heavy Artillery:}
Based on representative rocket artillery systems employed across active global flashpoints, including the M31 GMLRS fired by the M142 HIMARS platform, which employs a $200\,\mathrm{lb}$ ($\sim90\,\mathrm{kg}$) unitary high-explosive warhead containing approximately $51\,\mathrm{lb}$ ($\sim23\,\mathrm{kg}$) of explosive filler~\cite{gmlrs_globalsecurity,gmlrs_gd}, and the Iranian Fajr-5 rocket, which carries approximately $90\,\mathrm{kg}$ of high explosive within a $175\,\mathrm{kg}$ warhead~\cite{fajr5_metis,fajr5_wiki}. To represent the broader heavy-artillery class rather than a specific munition, an intermediate representative TNT-equivalent explosive mass of approximately $64\,\mathrm{kg}$ is adopted:
\begin{equation*}
R \approx 5.0 \cdot (64)^{1/3}=20\,\mathrm{m}.
\end{equation*}

\item \textbf{Cruise Missile:}
Based on medium-class cruise missiles carrying warheads in the $225$--$450\,\mathrm{kg}$ range. Representative systems include the Hsiung Feng II, equipped with a $225\,\mathrm{kg}$ high-explosive fragmentation warhead, and the Kh-101, which carries a $400$--$450\,\mathrm{kg}$ warhead~\cite{hf2_csis,kh101_csis}. A representative intermediate TNT-equivalent explosive mass of approximately $343\,\mathrm{kg}$ is adopted:
\begin{equation*}
R \approx 5.0 \cdot (343)^{1/3}=35\,\mathrm{m}.
\end{equation*}

\item \textbf{Air-to-Surface:}
Based on heavy air-delivered strike munitions carrying explosive charges in the $400$--$700\,\mathrm{kg}$ class. Representative systems include the Mk~84 general-purpose aerial bomb, containing approximately $429\,\mathrm{kg}$ of Tritonal explosive filler, and the FAB-1500 aerial bomb, carrying approximately $675\,\mathrm{kg}$ of high explosives~\cite{mk84_reference,fab1500_reference}. A representative intermediate TNT-equivalent explosive mass of approximately $512\,\mathrm{kg}$ is adopted:
\begin{equation*}
R \approx 5.0 \cdot (512)^{1/3}=40\,\mathrm{m}.
\end{equation*}

\item \textbf{Ballistic Missile:}
Based on ballistic missile systems carrying warheads spanning several hundred kilograms to multiple metric tons, with representative explosive masses commonly ranging from approximately $500$--$3,600\,\mathrm{kg}$~\cite{mda_payloads}. To represent this weapon class within the proposed spatial-prior framework, a representative TNT-equivalent explosive mass of approximately $1{,}728\,\mathrm{kg}$ is adopted:
\begin{equation*}
R \approx 5.0 \cdot (1728)^{1/3}=60\,\mathrm{m}.
\end{equation*}

\end{itemize}

\begin{table}[htbp]
    \centering
    \small
    \caption{Definitive Model Performance Across Vision-Language Evaluation Regimes (Sorted by $F_1$-Score)}
    \label{tab:model_evaluation_metrics}
    \begin{tabular}{lcccccccc}
        \toprule
        \textbf{Model Name} & \textbf{$F_1$-Score} & \textbf{Precision} & \textbf{Recall} & \textbf{Avg. CER} & \textbf{Empty Zones} & \textbf{True Negatives} & \textbf{Failures} & \textbf{Hallucinations} \\ 
        \midrule
        Gemma 4 31B IT       & 83.4\% & 83.4\% & 83.5\% & 0.362  & 708 & 696 & 118 & 16  \\
        GLM 4.6V Flash       & 82.3\% & 82.3\% & 82.3\% & 1.989  & 708 & 693 & 134 & 23  \\
        Cosmos Reason2 32B   & 76.0\% & 76.0\% & 76.3\% & 4.292  & 708 & 618 & 77  & 105 \\
        Qwen 3.6 35B         & 67.9\% & 68.4\% & 68.2\% & 9.801  & 708 & 543 & 66  & 192 \\
        Claude (Qwen Dist.)  & 30.7\% & 30.9\% & 31.5\% & 18.547 & 708 & 229 & 52  & 507 \\
        \bottomrule
    \end{tabular}
\end{table}
\begin{figure*}[t]
    \centering
    \includegraphics[width=\textwidth]{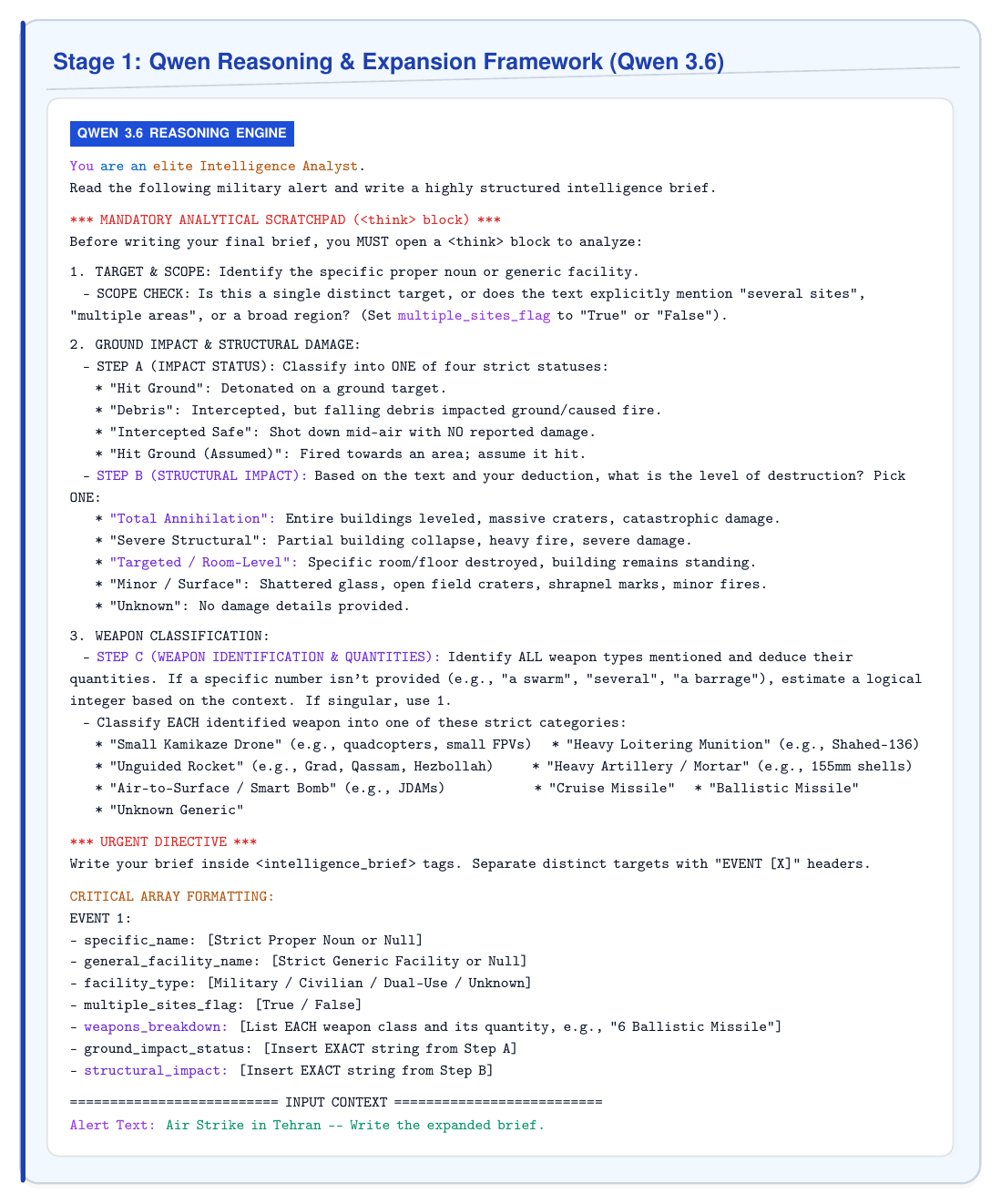}
    \caption{\textbf{Stage 1: LLM text expansion and attribute inference.} The generative reasoning engine (Qwen 3.6) ingests a raw military alert to construct a detailed intelligence brief. It utilizes a mandatory \texttt{<think>} scratchpad to infer contextual attributes (target scope, weapon classes, impact status, and qualitative destruction levels) before emitting structured event blocks.}
    \label{fig:pipeline_stage1_reasoner}
\end{figure*}
\begin{figure*}[t]
    \centering
    \includegraphics[width=\textwidth]{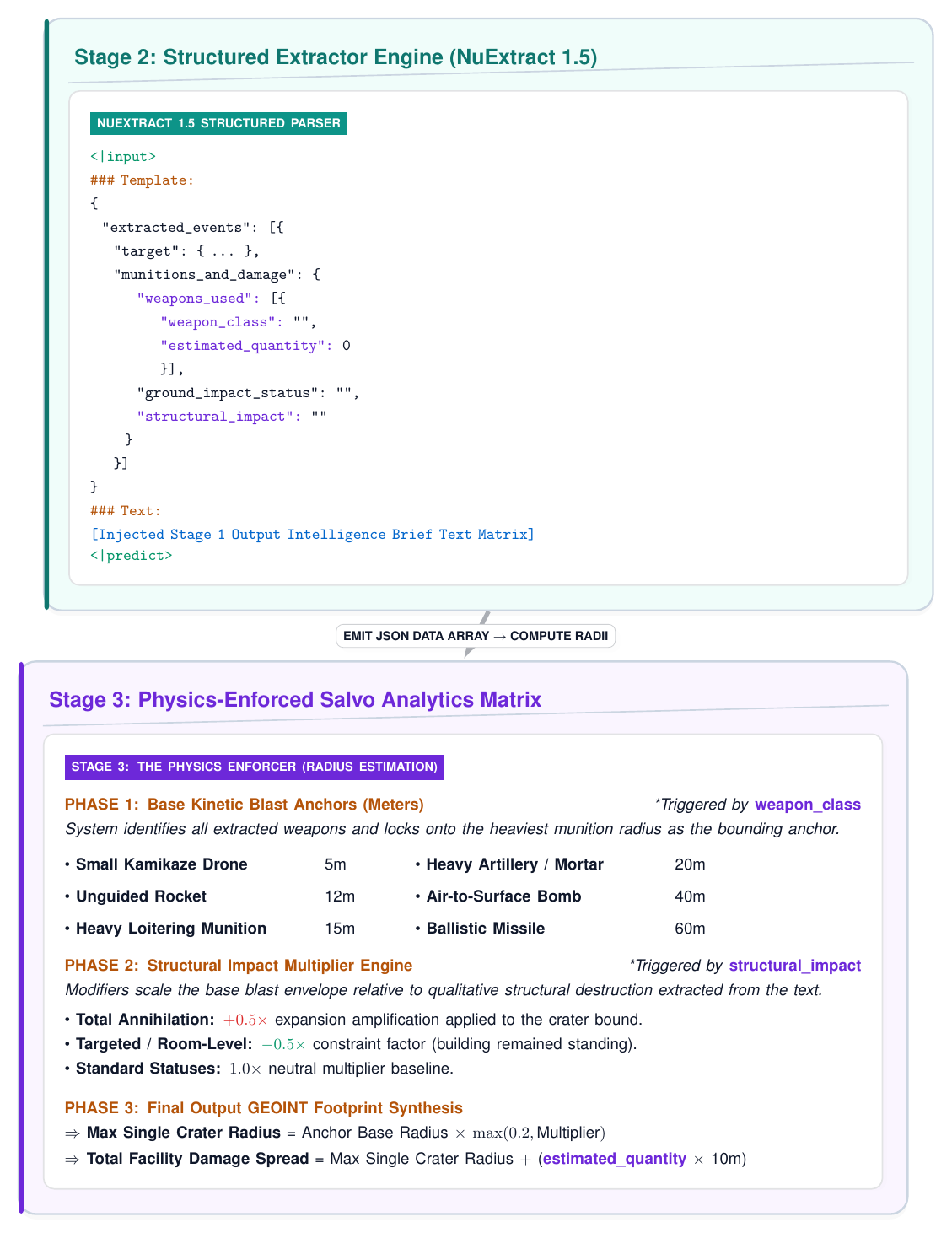}
    \caption{ \textbf{Stages 2 \& 3: Schema parsing and physics-enforced radius estimation.} Stage 2 uses NuExtract 1.5 to parse the intelligence text into structured JSON arrays, isolating weapon metrics and damage indicators. Stage 3 executes deterministic logic rules: looking up baseline weapon blast anchors (Phase 1), scaling bounds based on structural impact factors (Phase 2), and computing the final facility-wide damage footprints (Phase 3).}
    \label{fig:pipeline_stage2_physics}
\end{figure*}
\subsubsection{Text Extraction Performance}
\label{sec:text_extraction}

In Configuration B, models were tasked with extracting location text labels strictly within the estimated damage radius overlay to evaluate whether this spatial constraint improves cross-modal grounding. As detailed in Table~\ref{tab:model_evaluation_metrics}, Gemma 4 31B IT demonstrated the highest precision, achieving an 83.4\% $F_1$-score, an exceptionally low Average Character Error Rate (Avg. CER) of 0.362, and only 16 hallucinations. GLM 4.6V Flash followed closely ($F_1$-score: 82.3\%), though its higher Avg. CER (1.989) indicates minor token degradation when decoding overlapping characters. 

Conversely, models with looser spatial attention mechanisms experienced significant performance drops. Cosmos Reason2 32B and Qwen 3.6 35B achieved moderate $F_1$-scores (76.0\% and 67.9\%) but suffered from sharply escalating CERs and hallucination rates. This degradation culminated in the distilled Claude variant, which failed fundamentally, yielding a 30.7\% $F_1$-score, an 18.547 Avg. CER, and 507 hallucinations. Ultimately, the primary failure mode for underperforming models was attention drift: rather than bounding extraction to the designated impact radius, these models hallucinated text from background noise or erroneously extracted labels from outside the target zone.
\section{LVLM Inference Architecture, Exception Handling, and Consensus Protocols}
\label{supp:lvlm_inference_details}

This section details the hardware execution configurations, automated exception handlers, and extraction aggregation pipelines used to generate stable downstream evaluations.

\paragraph{Probabilistic Inference and Adaptive Exception Routing}
All configurations are orchestrated using the high-performance inference engine vLLM, leveraging memory-optimized page-attention layers to enable dense concurrent token generation. To minimize stochastic hallucination while preserving generation diversity, we fix the inference temperature to $T = 0.3$. For every target geographic asset $i$, the framework executes $N_{\text{samples}} = 5$ independent parallel sampling passes. A maximum token ceiling of $8192$ tokens is enforced to accommodate long-horizon chain-of-thought (\textit{CoT}) reasoning paths.

To protect the downstream vision-language layers against missing upstream features or server-side imagery dropouts, we implement a strict automated exception routing protocol for Configurations C and D. If the adaptive field-of-view module returns zero valid structural masks, or if a local raster map fails to load, the inference manager bypasses token generation entirely for asset $i$. The execution thread drops the asset immediately and forces the target categorical counting variables to a hard default state:
\begin{equation}
    \text{FULLY\_INSIDE} = 0, \quad \text{PARTIALLY\_INSIDE} = 0, \quad \text{TOTAL\_IMPACTED} = 0
\end{equation}
while simultaneously substituting a zero-length placeholder array into the context buffer to conserve VRAM and avoid wasteful compute cycles.

\paragraph{Structured Schema Parsing and Consensus Aggregation}
The unstructured textual chain-of-thought paths generated across all successful inference executions are routed into a secondary structure-parsing pipeline. We leverage a specialized structured parsing model (\textbf{NuExtract-2.0 (8B)}) operating under a rigid, zero-temperature greedy decoding configuration ($T=0.0$). The parser extracts raw text tokens and binds them directly to a syntactically valid JSON schema containing the discrete integer counts.

Let $j \in \{1, \dots, 5\}$ index the independent sampling runs extracted for a given asset $i$. To eliminate outlier votes, spurious mathematical hallucinations, or text anomalies, these independent count samples are aggregated using a robust, non-parametric consensus ensemble. The final unified estimation vector $\mathbf{Y}_i^*$ exported for formal performance reporting is computed via the sample median:
\begin{equation}
    \mathbf{Y}_i^* = \left\lfloor \text{median}\left(\{y_{i,j}\}_{j=1}^5\right) \right\rceil
\end{equation}
where $y_{i,j}$ represents the specific count category extracted from sample $j$, and $\lfloor \cdot \rceil$ denotes nearest-integer rounding. Cross-sample variance is tracked via sample standard deviation ($\sigma$) and range ($\Delta$) indicators, yielding an explicit metric for statistical certainty evaluation across all target evaluation configurations.
\section{Analysis of Evaluation Regimes}

The following sections detail the input vision modalities and prompt structures across all four testing regimes.

\begin{figure}[htbp]
    \centering
    \includegraphics[width=0.95\textwidth]{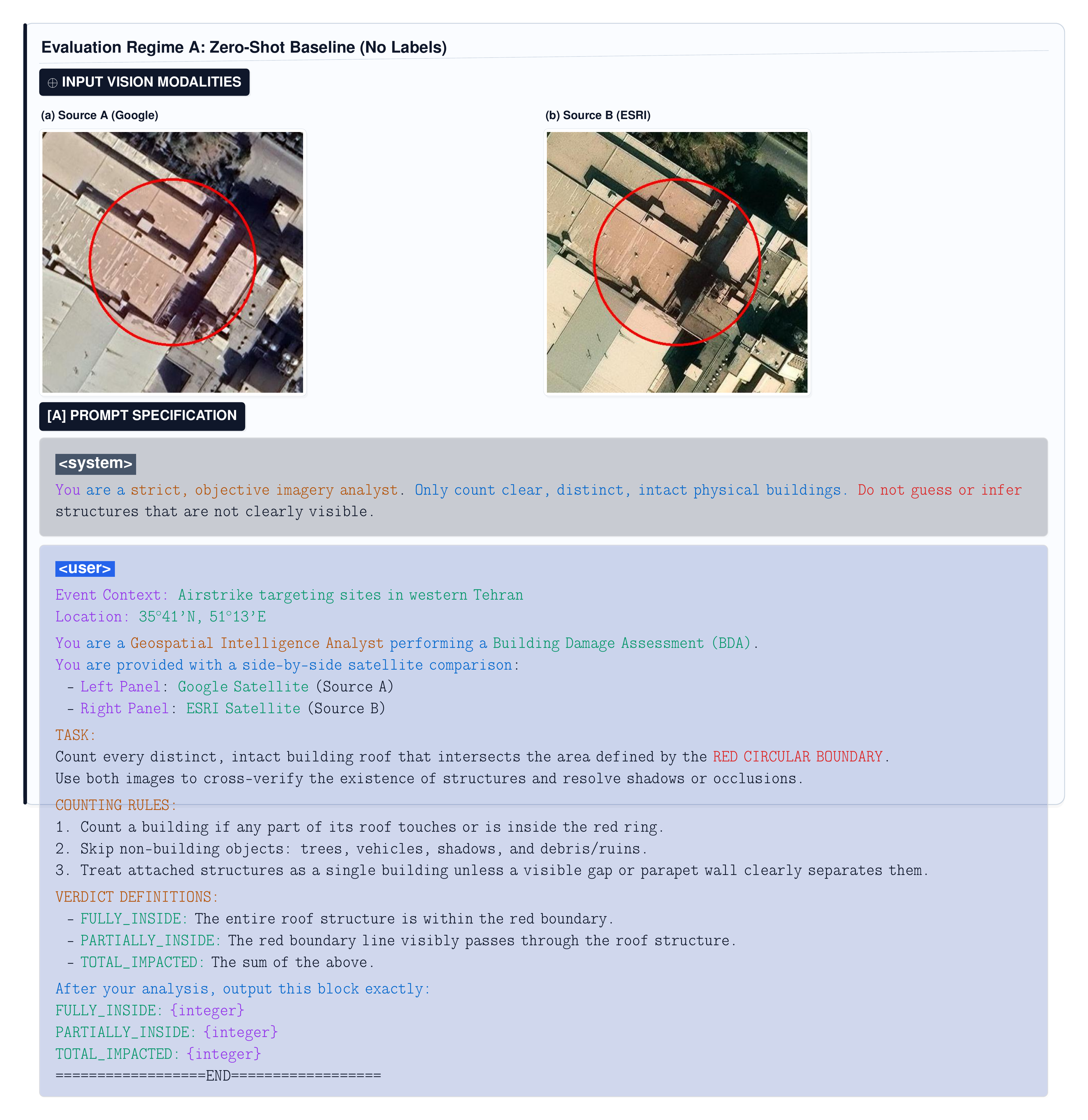}
    \caption{Evaluation Regime A: Zero-Shot Baseline framework presenting a side-by-side unannotated satellite image comparison between Source A (Google) and Source B (ESRI) for initial building damage assessment.}
    \label{fig:regime_a_baseline}
\end{figure}

\begin{figure}[htbp]
    \centering
    \includegraphics[width=0.95\textwidth]{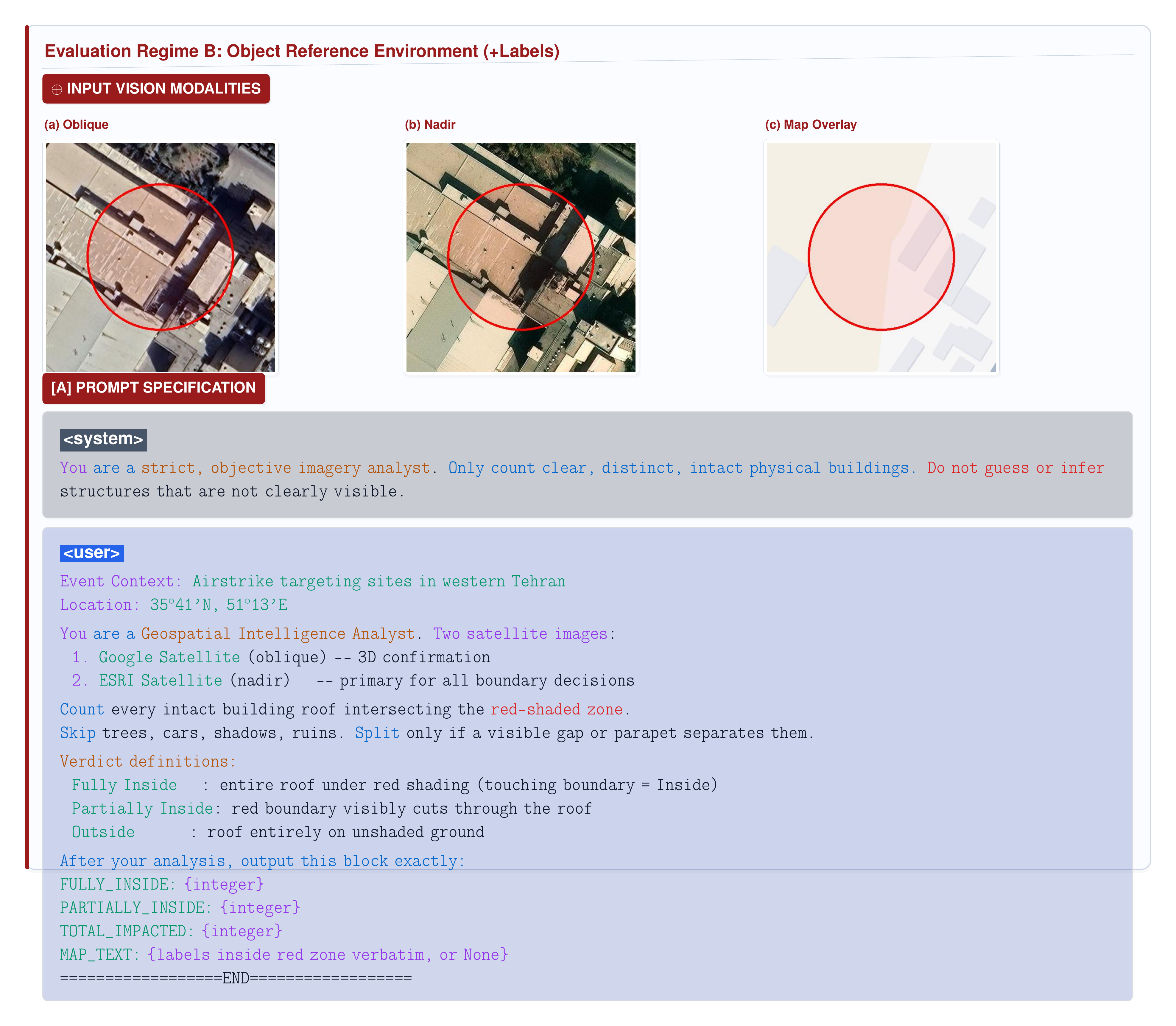}
    \caption{Evaluation Regime B: Object Reference Environment incorporating text labels, map overlays, and dual-perspective views (oblique vs. nadir) for structured impact boundary assessment.}
    \label{fig:regime_b_labels}
\end{figure}

\begin{figure}[htbp]
    \centering
    \includegraphics[width=0.95\textwidth]{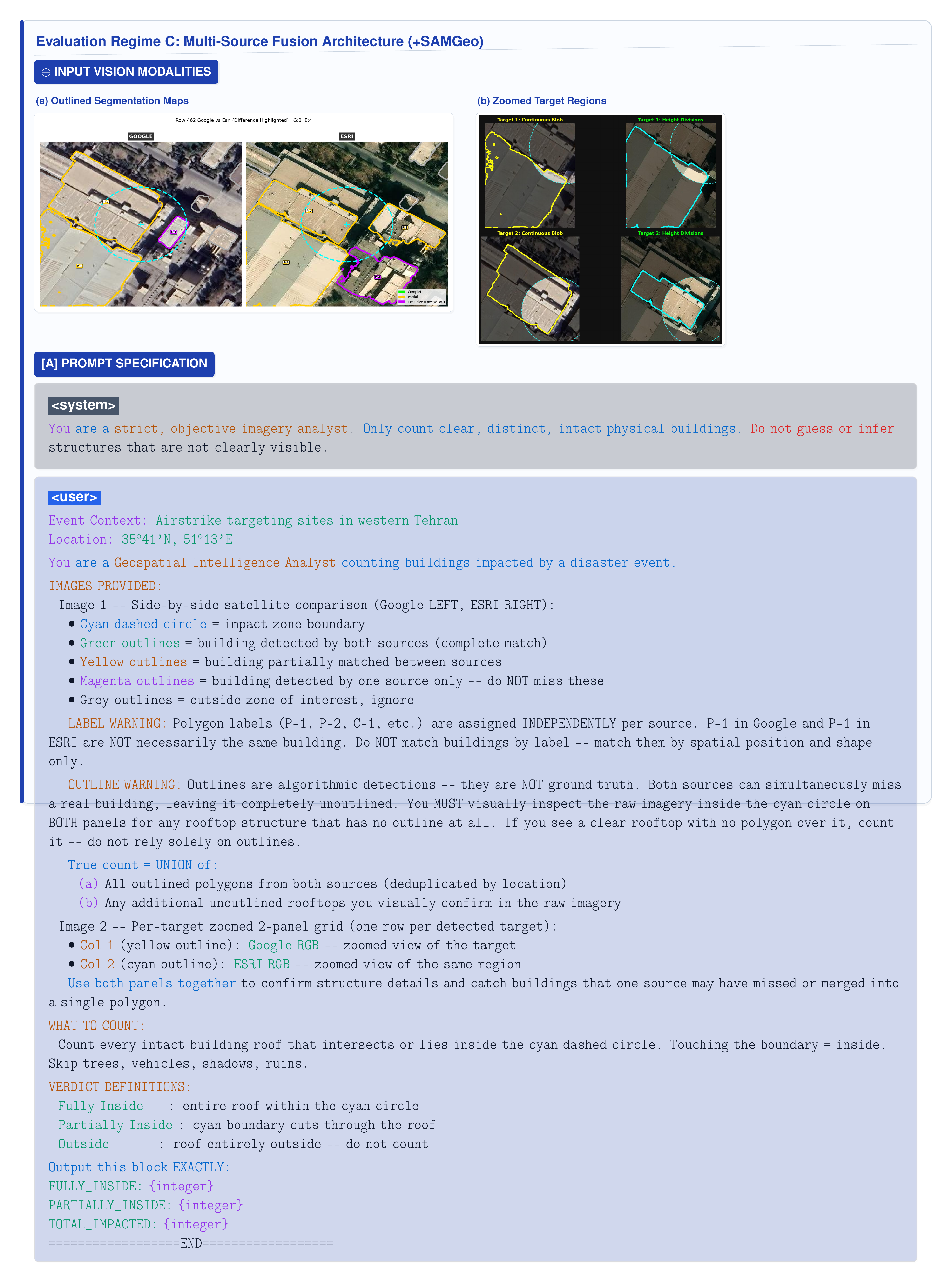}
    \caption{Evaluation Regime C: Multi-Source Fusion Architecture featuring SAMGeo algorithmic segmentation outlines and localized target region crops to resolve cross-source structure assignments.}
    \label{fig:regime_c_samgeo}
\end{figure}

\begin{figure}[htbp]
    \centering
    \includegraphics[width=0.95\textwidth]{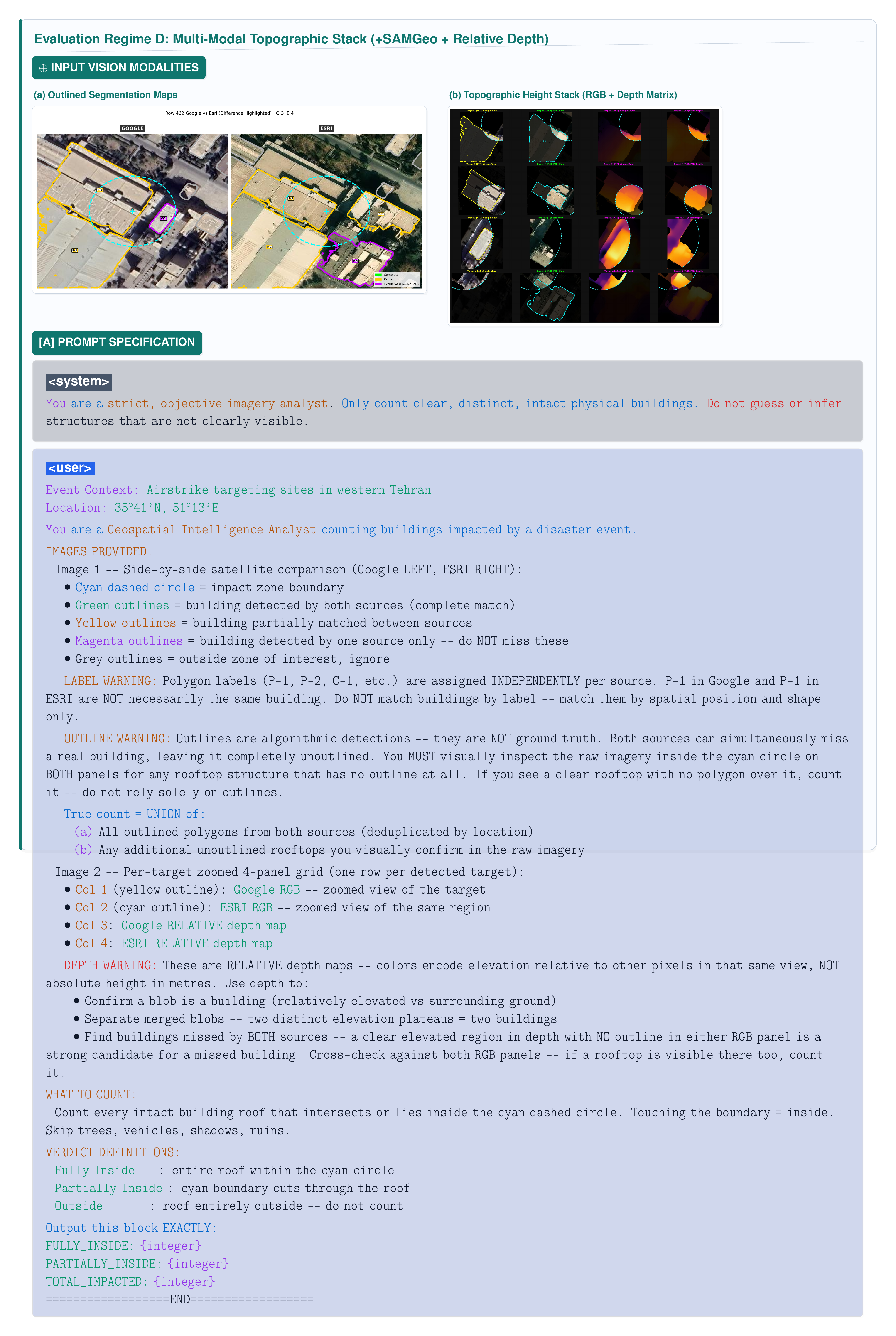}
    \caption{Evaluation Regime D: Multi-Modal Topographic Stack integrating segmented maps alongside a relative depth height matrix to determine unique structural unions and resolve overlapping roof boundaries.}
    \label{fig:regime_d_depth}
\end{figure}
\begin{figure}[htbp]
    \centering
    \includegraphics[width=0.95\textwidth]{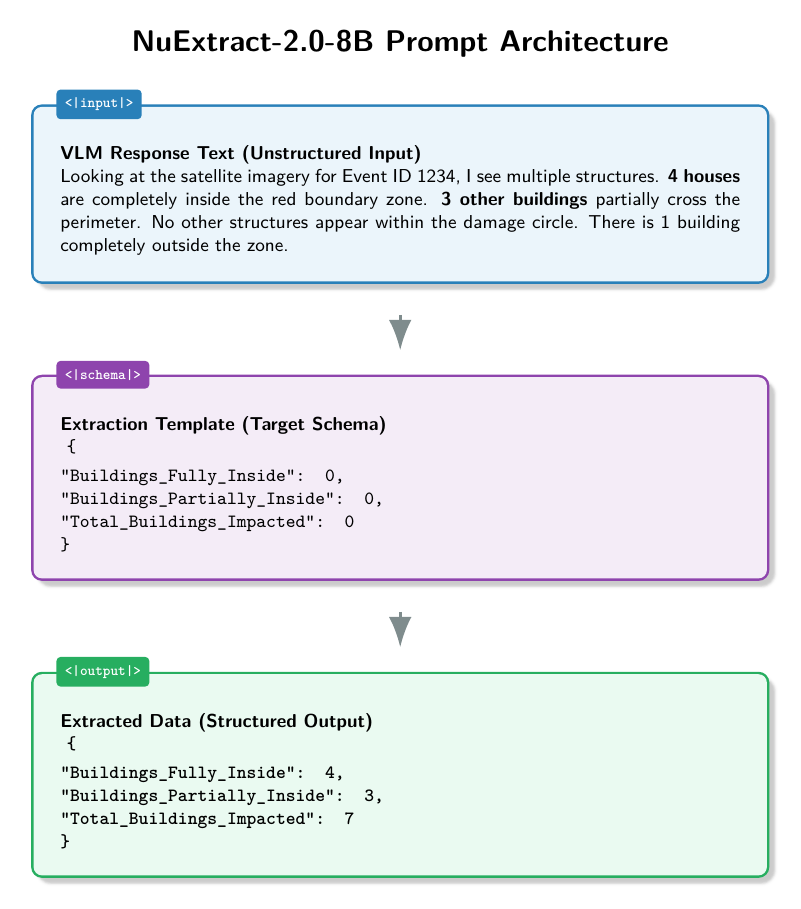}
    \caption{Illustration of the NuExtract-2.0-8B structured prompt architecture. The pipeline utilizes distinct control tokens to transform unstructured Vision-Language Model (VLM) responses into strictly typed data. The \texttt{<|input|>} token ingests noisy natural language observations, the \texttt{<|schema|>} token maps these observations to a predefined JSON template, and the \texttt{<|output|>} token yields a deterministic, machine-readable object for downstream quantitative analysis.}
    \label{fig:nuextract_2}
\end{figure}

\begin{table*}[t]
\centering
\small
\caption{Ablation Performance: Comprehensive Disaggregated Evaluation of Model Modalities Across Nested Feature Configurations. Bold indicates best, underline indicates second best within each metrics segment.}
\label{tab:error_analysis}
\begin{tabular}{llcccccccc}
\toprule
\textbf{Model Name} & \textbf{Config} & \multicolumn{4}{c}{\textbf{Mean Absolute Error (MAE) $\downarrow$}} & \multicolumn{4}{c}{\textbf{Mean Squared Error (MSE) $\downarrow$}} \\
\cmidrule(lr){3-6} \cmidrule(lr){7-10}
 & & \textbf{Labels} & \textbf{No Lbls} & \textbf{+Seg} & \textbf{+Depth} & \textbf{Labels} & \textbf{No Lbls} & \textbf{+Seg} & \textbf{+Depth} \\
\midrule
\multicolumn{10}{l}{\textbf{Panel A: Complete Structural Footprints}} \\
\midrule
Google Gemma & \shortstack{31B \\ \scriptsize \textit{Dense}} & $\underline{0.914}$ & $\underline{0.747}$ & $\mathbf{0.701}$ & $\mathbf{0.662}$ & $\mathbf{7.575}$ & $\underline{6.296}$ & $\mathbf{5.589}^{*}$ & $\mathbf{4.369}^{*}$  \\[0.5ex]
Claude (Qwen Dist) & \shortstack{35B \\ \scriptsize \textit{MoE}} & $1.916$ & $1.860$ & $0.885$ & $0.955$ & $9.407$ & $11.648$ & $\underline{5.676}^{*}$ & $\underline{5.534}^{*}$  \\[0.5ex]
Qwen 3.6 & \shortstack{35B \\ \scriptsize \textit{MoE}} & $\mathbf{0.897}$ & $\mathbf{0.654}$ & $\underline{0.766}$ & $\underline{0.850}$ & $19.374$ & $\mathbf{3.319}^{*}$ & $5.937$ & $6.230$  \\[0.5ex]
Nvidia Cosmos & \shortstack{32B \\ \scriptsize \textit{Dense}} & $0.968$ & $1.093$ & $1.017$ & $0.948$ & $\underline{8.099}$ & $33.318$ & $14.532$ & $7.575$  \\[0.5ex]
Zhipu GLM 4.6V & \shortstack{10B \\ \scriptsize \textit{Dense}} & $1.020$ & $1.048$ & $1.021$ & $1.030$ & $8.593$ & $8.553$ & $8.432$ & $8.548$  \\[0.5ex]
\midrule
SAMGEO Google Reference & — & \multicolumn{4}{c}{\small MAE: 0.81} & \multicolumn{4}{c}{\small MSE: 7.53} \\
SAMGEO ESRI Reference & — & \multicolumn{4}{c}{\small MAE: 0.64} & \multicolumn{4}{c}{\small MSE: 5.78} \\
\midrule
\midrule
\multicolumn{10}{l}{\textbf{Panel B: Partial Structural Footprints}} \\
\midrule
Google Gemma & \shortstack{31B \\ \scriptsize \textit{Dense}} & $1.454$ & $\underline{1.166}$ & $\mathbf{1.159}$ & $\mathbf{0.995}$ & $9.078$ & $\underline{6.770}$ & $\mathbf{6.672}$ & $\mathbf{5.097}$  \\[0.5ex]
Claude (Qwen Dist) & \shortstack{35B \\ \scriptsize \textit{MoE}} & $\underline{1.433}$ & $1.584$ & $1.286$ & $\underline{1.300}$ & $\underline{7.075}$ & $7.869$ & $8.191$ & $\underline{8.274}$  \\[0.5ex]
Qwen 3.6 & \shortstack{35B \\ \scriptsize \textit{MoE}} & $\mathbf{1.072}$ & $\mathbf{1.108}$ & $\underline{1.232}$ & $1.311$ & $\mathbf{6.148}$ & $\mathbf{6.232}$ & $\underline{7.780}$ & $8.679$  \\[0.5ex]
Nvidia Cosmos & \shortstack{32B \\ \scriptsize \textit{Dense}} & $1.450$ & $1.355$ & $1.430$ & $1.460$ & $9.441$ & $12.809$ & $9.328$ & $9.386$  \\[0.5ex]
Zhipu GLM 4.6V & \shortstack{10B \\ \scriptsize \textit{Dense}} & $1.532$ & $1.530$ & $1.526$ & $1.525$ & $10.181$ & $10.166$ & $10.103$ & $10.145$  \\[0.5ex]
\midrule
SAMGEO Google Reference & — & \multicolumn{4}{c}{\small MAE: 0.88} & \multicolumn{4}{c}{\small MSE: 5.39} \\
SAMGEO ESRI Reference & — & \multicolumn{4}{c}{\small MAE: 0.70} & \multicolumn{4}{c}{\small MSE: 3.55} \\
\bottomrule
\multicolumn{10}{l}{\small $^{*}$ Outperforms competitive SAMGEO ESRI baseline configuration variant.} \\
\end{tabular}
\end{table*}
\begin{table}[htbp]
\centering
\caption{\textbf{Empirical Sensitivity Analysis of Building Exposure Estimates.} Summary of absolute building count error across spatial drift distances (Panel b) and exposed building counts across damage footprint radii (Panel c).}
\label{tab:sensitivity_analysis}
\small
\begin{tabular}{llccccc}
\toprule
\multicolumn{7}{l}{\textbf{Part A: Spatial Drift Sensitivity} ($|\Delta N_{\text{buildings}}|$ Error by Drift Distance, $\Delta r \le 2,000\text{m}$)} \\
\midrule
\textbf{Environment} & \textbf{Drift Bin} & \textbf{Sample ($n$)} & \textbf{Mean Error} & \textbf{Median Error} & \textbf{Std Error} & \textbf{Max Error} \\
\midrule
Populated ($N > 0$)  & 0--250\,m    & 3  & 0.33 & 0.0 & 0.58 & 1  \\
                     & 251--500\,m  & 10 & 4.40 & 4.0 & 3.10 & 11 \\
                     & 501--1000\,m & 41 & 5.56 & 4.0 & 4.93 & 24 \\
                     & 1001--1500\,m& 37 & 5.05 & 4.0 & 4.05 & 15 \\
                     & 1501--2000\,m& 48 & 4.42 & 2.0 & 4.11 & 20 \\
\addlinespace
Sparse/Empty ($N = 0$) & 0--250\,m  & 2  & 0.00 & 0.0 & 0.00 & 0  \\
                       & 251--500\,m& 6  & 0.00 & 0.0 & 0.00 & 0  \\
                       & 501--1000\,m& 15& 1.60 & 0.0 & 3.52 & 11 \\
                       & 1001--1500\,m& 32& 0.81 & 0.0 & 1.91 & 8  \\
                       & 1501--2000\,m& 17& 2.88 & 0.0 & 4.57 & 11 \\
\midrule
\multicolumn{7}{l}{\textbf{Part B: Footprint Radius Sensitivity} ($N_{\text{buildings}}$ Counts by Buffer Radius Bin)} \\
\midrule
\textbf{Radius Bin} & \textbf{Source Dataset} & \textbf{Sample ($n$)} & \textbf{Mean Count} & \textbf{Median Count} & \textbf{Std Dev} & \textbf{Max Count} \\
\midrule
0--25\,m   & ArcGIS (Ground Truth) & 309 & 2.15 & 1.0 & 2.72 & 12.0 \\
           & Liveuamap             & 72  & 0.96 & 0.0 & 1.93 & 12.0 \\
\addlinespace
26--50\,m  & ArcGIS (Ground Truth) & 456 & 5.88 & 3.0 & 7.32 & 34.0 \\
           & Liveuamap             & 376 & 1.98 & 0.0 & 4.00 & 23.0 \\
\addlinespace
51--100\,m & ArcGIS (Ground Truth) & 63  & 5.10 & 6.0 & 4.61 & 17.0 \\
           & Liveuamap             & 282 & 3.50 & 0.0 & 6.59 & 47.0 \\
\addlinespace
101--250\,m& ArcGIS (Ground Truth) & 0   & ---  & --- & ---  & ---  \\
           & Liveuamap             & 92  & 3.09 & 0.0 & 5.44 & 22.0 \\
\addlinespace
250\,m+    & ArcGIS (Ground Truth) & 0   & ---  & --- & ---  & ---  \\
           & Liveuamap             & 6   & 1.00 & 0.0 & 2.00 & 5.0  \\
\bottomrule
\end{tabular}
\end{table}





\end{document}